\documentclass[preprint, nopreprintline, 10pt]{elsarticle}
\usepackage{graphicx}
\usepackage{epsfig}
\usepackage{amssymb}
\usepackage{amsmath}
\usepackage{multirow}
\usepackage{color}
\usepackage{algorithm}
\usepackage{algpseudocode}
\usepackage{wrapfig}
\usepackage{caption}
\usepackage{hyperref}
\usepackage{float}
\usepackage{algorithm}
\usepackage{algpseudocode}
\usepackage{longtable}

\usepackage[percent]{overpic} 
\usepackage{placeins}

\usepackage{setspace}

\usepackage{bm}
\usepackage{dsfont}
\usepackage{array}
\newcolumntype{M}[1]{>{\centering\arraybackslash}m{#1}}

\usepackage[dvipsnames]{xcolor}

\graphicspath{ {./images/} }

\usepackage{nfontdef}
\input{math_commands}

\journal{}

\begin{document}
\begin{frontmatter}

\title{Segmenting Low-Contrast XCTs of Concrete: An Unsupervised Approach}


\author[k132]{Kaustav Das}
\author[list]{Gaston Rauchs}
\author[k132]{Jan~S\'ykora}
\author[k132]{Anna~Ku\v{c}erov\'a}
\ead{anna.kucerova@cvut.cz}

\cortext[auth]{Corresponding author. Tel.:~+420-2-2435-4375; fax~+420-2-2431-0775}

\address[k132]{Department of Mechanics, Faculty of Civil Engineering,
  Czech Technical University in Prague, Th\'{a}kurova 7, 166 29 Prague
  6, Czech Republic}
\address[list]{{Structural Composites Unit}, Luxembourg Institute of Science and Technology (LIST), 5 avenue des Hauts-Fourneaux
L-4362 Esch-sur-Alzette, Luxembourg}


\begin{abstract}
    X-Ray Computed Tomography (XCT) is a compelling tool in experimental mechanics, capable of non-destructively extracting information pertaining to the internal morphology of materials. For materials with random heterogeneous morphology such as concrete, such information is of particular relevance since it allows for studies of morphology-related behaviour and for predictive modelling. Nevertheless, XCT images require semantic segmentation for practical usage. Here, concrete poses a unique challenge due to the similar X-ray attenuation coefficients of aggregates and mortar, which result in low contrast between the two phases in the ensuing XCT images. As such, purely intensity-dependent semantic segmentation tools remain unfeasible. While vision transformers (ViTs) and convolutional neural networks (CNNs) are proven techniques for semantic segmentation in such challenging cases, they typically require labelled training data, which is often unavailable  for concrete or resource-intensive to obtain, thereby limiting their relevance. To address this challenge, a self-annotation technique is presented here that leverages superpixel algorithms to identify perceptually similar local regions in an image and relates them to the global context by utilizing the receptive field of a CNN-based model. This enables the model to learn a global-local relationship in the images and facilitates the identification of semantically similar structures. When evaluated against manually annotated ground truth on out-of-distribution data, the proposed methodology consistently outperformed direct greyscale thresholding across all pertinent metrics, demonstrating improved discernibility between aggregates and mortar, and providing the most favourable balance of sensitivity and precision for aggregate-phase identification.
\end{abstract}

\begin{keyword}
X-ray Computed Tomography (XCT), Cementitious Composites, Self-Annotation, Phase Identification, Unsupervised Semantic Segmentation
\end{keyword}

\end{frontmatter}

\section{Introduction}
\label{sec:intro}

X-ray Computed Tomography (XCT) has garnered substantial interest in experimental mechanics due to its ability to provide information about the internal structure of materials in a non-contact and non-invasive manner. The possibility of extracting detailed morphological information has resulted in increasing adoption of XCT for the inspection of multi-phase materials such as composites to study both their internal morphological heterogeneity and defects \cite{vasarhelyi2020microcomputed, naresh2020use}. Furthermore, subsequent developments in \emph{in-situ} XCT devices now allow the observation of materials under loading, enabling the application of full-field measurement techniques such as digital volume correlation (DVC) \cite{hufenbach2012test, bohm2015quantitative, yang2017situ}. 

In a wider context, having information about a material's internal morphology and the related property metrics can allow for the application of \emph{microstructure-sensitive} design methodologies \cite{fullwood2010microstructure} to facilitate the design of materials with internal structures tailored for achieving target performance. This is further enabled by the emergence of powerful computational tools both in terms of hardware and software algorithms that can handle large amounts of data, thereby motivating the use of high-throughput computational methodologies for the development of new materials by enabling targeted experiments \cite{curtarolo2013high}. Demonstrations of such methodologies are seen in the recent emergence of concepts such as digital twins of materials \cite{kalidindi2022digital}, which are conceptualized as highly detailed digital representations of materials, where specialized morphological characterization tools \cite{bostanabad2018computational} are among key enablers. 

In the present study, we limit ourselves to concrete, a material that is second only to water in usage worldwide \cite{gagg2014cement}. Hence, even modest improvements in its design, production, or utilization can translate into significant environmental and economic benefits at a global scale. It is recognized that morphological heterogeneities play an important role in the overall behaviour of concrete under various loading conditions. Morphology dependence is observed in phenomena such as drying rate and associated shrinkage and cracks \cite{bisschop2008effect, fujiwara2008effect}, compressive strength \cite{meddah2010effect}, interface properties \cite{myong1992fracture, rao2004influence}, porosity-dependent strength \cite{kumar2003porosity}, among others. In most use cases involving analysis of concrete in a multi-scale setting, it is  generally separated into three different phases, namely:\emph{ aggregates, mortar} and \emph{voids} in the form of cracks and porosity \cite{sleiman2023tomographic}. Additionally \emph{interfacial transition zones} (ITZs) \cite{scrivener2004interfacial} are often considered as an additional phase for analysis, but these depend on the morphology of aggregates. Numerical modelling of concrete using methods such as the finite element analysis (FEA) \cite{wriggers2006mesoscale, huang2023ct_image, hai2024dynamic}, the discrete element method \cite{beckman2012dem, nitka2018three_dimensional, wang2020dem_analysis}, or discrete lattice systems \cite{bolander2021discrete, grassl2010mesoscale, elias2015stochastic}, plays a crucial role in understanding its behaviour, especially when taking into consideration the morphological heterogeneity of its constitutive phases for multi-scale modelling to determine homogenized macroscopic material properties. Despite the variety in numerical approaches, the common pre-requisite for such studies is the availability of the computational domain i.e., the meso-structure. 

With regard to concrete, it is a common practice to create the computational domain for multi-scale numerical analysis using a placement algorithm which positions aggregate geometries inside the domain by avoiding overlaps and also taking into account factors such as granulometry, wall-effect or aggregate settlement \cite{ren2023methods}. Aggregate geometries in such approaches, when generated using parametrisation modelling, are frequently in the form of spheres, ellipsoids or polyhedroids \cite{Thilakarathna2020mesoscale}, with spheres and ellipsoids being the more frequent and convenient choice \cite{idiart2012numerical, havlasek2016multiscale, wriggers2006mesoscale, miao2024numerical}, although it is also noted that polyhedroids lead to more realistic accounting of local stress concentration and crack initiation as seen in the case of crushed aggregates with rough surfaces \cite{Thilakarathna2020mesoscale}. A more complex approach to generating parametrized aggregate shapes is in the use of spherical harmonic functions \cite{garboczi2002three, liu2011spherical}, or random fields coupled with discrete Fourier transformations \cite{mollon20143d}, which result in aggregates with more realistic geometries. Alternatively, aggregate shapes are also obtained from large databases of reference aggregate geometries obtained through laser-scanning of actual aggregates \cite{zhou2015concrete} or from XCT scans of bulk concrete \cite{wang2022establishing, huang2023ct_image}. An alternative approach is to use greyscale XCT scans of concrete castings, which not only result in realistic geometric and dispersion characteristics of the computational domain, but also include defects such as porosity and cracks that are common in concretes \cite{yang2017situ, nitka2018three_dimensional}. Furthermore, greyscale XCT images are also a valuable source of morphological information for microstructure characterisation and reconstruction (MCR) approaches \cite{bostanabad2018computational} that are extensively used for analysis and generation of synthetic material micro- and meso-structures for statistical analysis. In the case of concrete, examples of such application can be observed in works involving generation of synthetic aggregates that are statistically similar to observed aggregate shapes \cite{guo2023spherical} or in applications such as generation of an entire concrete domain comprising various phases using diffusion-based deep generative models \cite{liang2025conditional}.

However, it is rarely feasible to directly use the greyscale XCT images in most cases. In particular, the images must be semantically segmented in order to partition them into distinct, non-overlapping contiguous regions, with each region representing an instance of one of the material phases. In case of concrete, semantic segmentation of XCT images is particularly challenging because aggregates and mortar have similar X-ray attenuation coefficients, which result in similar and often overlapping greyscale intensities for these two phases, making distinction between them challenging \cite{stamati2018phase, thakur2023phase}. Attempts have been made to improve the discernibility of these phases, for example, through the use of contrast enhancers as additives during concrete mixing \cite{carrara2018improved} or through the use of X-ray tomography \cite{brisard2020multiscale} and neutron tomography \cite{zhang2018application} in conjunction to improve the contrast between the phases \cite{kim2021reconstruction, sleiman2023tomographic}. While the former approach for contrast enhancement requires preparation of special test samples exclusively for the purpose of tomography which may not be representative of the actual concrete mix being studied, and brings additional overheads, the later approach requires access to two distinct tomographic devices, which may not be feasible in most practical cases. 

Semantic segmentation of images is a much-studied topic in the field of computer vision and has seen substantial progress since the emergence of deep convolutional neural networks (CNNs) \cite{lecun1989backpropagation, lecun1998gradient}, facilitated by gradient-based optimization using efficient back-propagation algorithms \cite{rumelhart1986learning}. The specialized case involving computed tomography (CT) images is a major topic of research in biomedical sciences, where imaging is used in a wide range of applications. A few examples include CT imaging-guided modelling of the human femur and its internal structure \cite{falcinelli2020image, musy2017notonly}, stability study of dental implants considering bone damage caused by cyclic loading \cite{khorshidparast2023measurement}, segmentation of kidney images for diagnosis \cite{torres2018kidney}, and detection of lung cancer \cite{chen2021lung, primakov2022automated}. Notably, developments such as the widely used U-Net architecture for semantic segmentation of CT images \cite{ronneberger2015u} have emerged out of this field. Considering the prevalence of imaging-based analysis in biomedical applications, there exists a large diversity of openly accessible datasets, for example, The Cancer Imaging Archive \cite{clark2013thecancer} comprising millions of images, the MIMIC-CXR database of chest X-rays comprising over 300 thousand images \cite{PhysioNet-mimic-cxr-jpg-2.1.0}, and the LIDC/IDRI database of lung imaging \cite{armato2011thelung}. Consequently, such datasets enable works that employ thousands of images for training CNN-based supervised image segmentation models \cite{chen2021lung,primakov2022automated}.

In the case of concrete, however, semantic segmentation of XCT images using deep learning (DL)-based methods has not reached the same degree of maturity when compared to biomedical applications due to the lack of access to large, well-curated datasets. Therefore, a notable section of works continues to rely on heuristic-based morphology operations \cite{zhou2015concrete, kim2024recursive}, pixel intensity thresholding-based approaches \cite{loeffler2018detection, stamati2018phase, sleiman2023tomographic}, classical machine learning algorithms \cite{saha2020use, thakur2023phase, wen2024simplified} or K-means clustering and level set-based methods \cite{pecha2017advanced, chen2021identification} to name a few. All of these methods primarily consider local image information, such as pixel intensity values, and are ill-suited for noisy images or images with uneven illumination, which are frequently observed in concrete XCT images \cite{carrara2018improved, kim2021reconstruction}. Furthermore, such methods struggle to generalise beyond the specimen under consideration. Nevertheless, DL-based tools are increasingly being adopted for analysing concrete images across a variety of applications. CNN-based models have seen use in the identification of damage in concretes resulting from alkali-silica reaction using scanning electron microscopy images  \cite{bajcsy2020approaches}; in the identification of cracks and distinguishing them from porosity defects  \cite{dong2020microstructural, tian2021meso, jung2022towards}; for semantic segmentation of full-colour micrographs of concrete to identify aggregates using supervised \cite{wang2022automatic, hilloulin2022modular, bangaru2022scanning} and  semi-supervised learning\cite{shi2024weakly, coenen2022consinstancy}; for semantic segmentation of XCT images of cement-based fibre reinforced composites to identify air-voids, quartz sand and fibres \cite{lorenzoni2020semantic}; for XCTs of concrete castings to partition the domain into its major constitutive phases \cite{yang2024fine, zhang2026meso, jin2026attention} and XCT of 3D-printed concrete specimens \cite{zhao2026integrated} among others. However, all of these works depend on the availability of labelled or annotated data, including the semi-supervised methods, where a substantial proportion of the images were manually annotated. Focusing specifically on XCT scans, training the model in \cite{tian2021meso} required 2400 image slices for identifying cracks, which is a relatively easier problem considering the difference in greyscale values between the solid and porous phases. In contrast, while the studies involving semantic segmentation for identifying aggregates in XCT scans reported a modest number of slices ranging between 150 and 400 image slices before data augmentation \cite{yang2024fine, zhang2026meso, zhao2026integrated}, the training data in these cases were obtained by selecting slices from each concrete specimen at fixed intervals, and the tests were reported on other slices of the same specimens. This makes it difficult to ascertain the generalizability of the method to other concrete specimens, since slices in close proximity in XCTs are often structurally very similar. Furthermore, in  \cite{jin2026attention}, the test sample comprised full-colour images, which usually possess better contrast between aggregates and concrete, thereby leaving the assessment of the model's actual performance on XCT images unclear.

Consequently, the scarcity of annotated training data remains a primary bottleneck preventing the wider adoption of DL-based tools for concrete XCT image segmentation to distinguish between aggregates and mortar. To address this, approaches such as transfer learning can be used to adapt pre-trained models \cite{weinberger2024unsupervised} with minimal retraining. For instance, models such as Segment Anything (SA) \cite{kirillov2023segment, ravi2024sam}, which are large, general-purpose architectures built specifically for image segmentation, can be adapted for this purpose. Nevertheless, SA models are designed to operate on a wide variety of generic images and are trained on over 11 million colour images with more than 1 billion annotations. For specialized applications, these foundational models may not be computationally optimal and satisfactory performance may still be achieved using smaller, purpose-built segmentation models.

An alternative approach to address this challenge is the adoption of unsupervised learning techniques. Unlike supervised segmentation, where each pixel is paired with a fixed target label throughout training, unsupervised image segmentation pairs a pixel with a set of multiple feasible target labels, and the model is expected to learn the correct label for any given pixel during the training process. However, because a given pixel often gets assigned different target labels from this set during training, the learning process becomes inherently difficult. Such unsupervised semantic segmentation frameworks have been explored within the biomedical domain for CT images of organs \cite{quyanf2022self}, in the agricultural domain for images of vegetation on fields \cite{gueldenring2021self_supervised, roggiolani2025unsupervised} and in the broader computer vision domain focusing on large datasets comprising everyday objects \cite{xia2017w, kanezaki2018unsupervised, kim2020unsupervised, niu2024unsupervised}, meeting with varying degrees of success. However, in case of concrete unsupervised methods have primarily been applied to detection and segmentation of cracks \cite{jung2022towards, beyene2023unsupervised, chun2024self_training,zhao2025cross_dataset, aung2026cracknet}, with \cite{ziabari2025constrained} reporting the unfeasibility of the unsupervised segmentation method in \cite{quyanf2022self} on unlabelled XCT images of concrete due to their lower contrast compared to the greater variability observed among human organs in CT images, putting emphasis on supervised training of segmentation models using augmented datasets comprising synthetically generated XCT-like image-label pairs. 

A common characteristic of existing unsupervised learning frameworks dealing with semantic segmentation of everyday objects is that the models are trained by defining a set of feasible labels that is often larger than the number of possible classes in a given dataset, which can result in similar semantic objects being assigned different class labels. Moreover, they place limited emphasis on precisely capturing complex object boundaries. In contrast, materials science applications require a higher level of accuracy, as the segmented components must strictly adhere to physical constraints, such as phase volume fractions or precise limits on size and shape, making the preservation of these boundaries absolutely critical. This is especially important if the downstream applications involve mesh generation for multi-scale numerical modelling. These observations therefore indicate an existing gap in research concerning high-throughput, spatially-aware domain partitioning techniques for low-contrast concrete XCT images that achieve acceptable accuracy while circumventing the training data bottlenecks typical of deep learning models. Furthermore, the technique must rely on only a single tomography imaging process while eliminating the need to create concrete samples with additives exclusively for obtaining better contrast in XCT images.

The present study therefore demonstrates the feasibility of segmenting low-contrast concrete XCT images to distinguish between aggregates and mortar using a CNN-based model trained in an unsupervised manner. To this end, the method originally proposed in \cite{kanezaki2018unsupervised} for unsupervised segmentation, which has also seen application in training a semantic segmentation model for steel micrographs in \cite{Kim2020metallurgy}, is adapted. A salient feature of this technique is the use of superpixels \cite{barcelos2024superpixel}, which allow deterministic control over the size of the smallest perceptually similar regions of interest in the images and the identification of their boundaries. In addition to detailing the underlying mechanism of the technique, this work highlights a critical limitation stemming from a necessary normalization step. Furthermore, in a departure from the common practice of splitting a single XCT volume into training and test sets, the model is evaluated on out-of-distribution data from an entirely separate XCT volume, demonstrating the inherent robustness of the proposed methodology prior to dataset-specific fine-tuning.

The remainder of this paper is organized as follows. Section~\ref{sec:dataset} details the XCT data acquisition procedure and the subsequent preprocessing steps for standardization of the model inputs. Section~\ref{sec:unsupervised_segmentation} provides a comprehensive overview of the unsupervised segmentation framework, while Section~\ref{sec:Implementation} describes its algorithmic implementation. The experimental setup, post-processing techniques, and evaluation metrics are covered in Sections~\ref{sec:experiments}, \ref{sec:postprocessisng}, and \ref{sec:results_and_eval}, respectively. Finally, Section~\ref{sec:conclusion} concludes the paper with a summary of the core findings and an outlook on future work.
\section{X-Ray Computed Tomography Data of Concrete Specimens} 
\label{sec:dataset}

\subsection{X-Ray Tomography Procedures}
\label{subsec:x-ray_tomo_methodology}

\begin{figure}[h!]
\centering
\includegraphics[trim=0 200 0 0, width=1.0\linewidth]{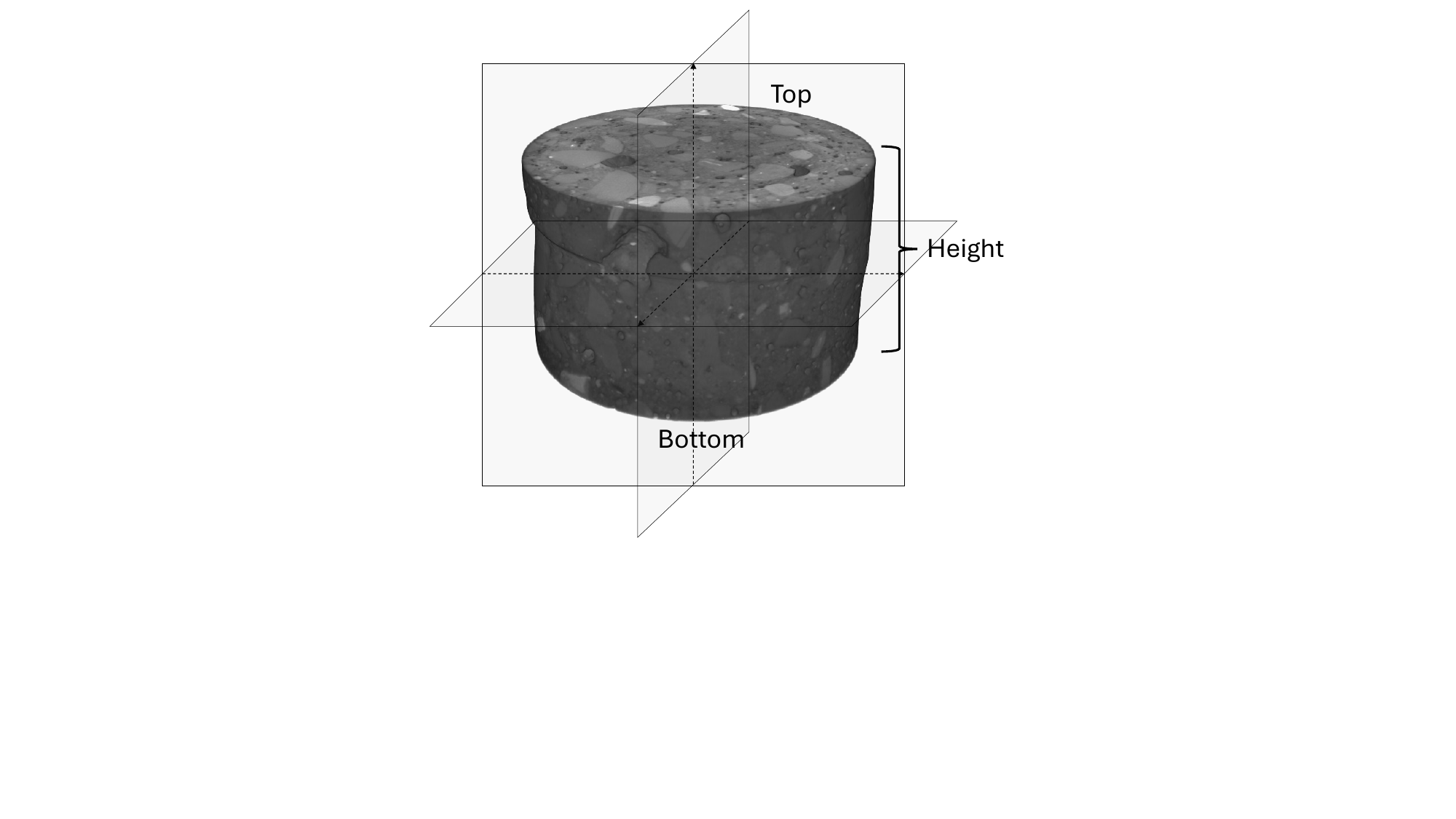}
\caption{Example of the concrete XCT specimen in 3D.}
\label{fig:concrete_cylinder_3D_view}
\end{figure}

X-ray computed tomography (XCT) was conducted with the objective to identify three distinct phases in concrete: pores, aggregate, and mortar. To this avail, cylindrical concrete castings, each with a diameter of approximately $94\,\text{mm}$ and varying heights were used. The composition of the concrete is given in \cite{havlasek2021shrinkage}. XCT was performed using an X-Ray MicroCT EasyTom 160 tomograph and Xact reconstruction software (RX Solutions, Chavanod, France). All samples were scanned with a tube voltage of $160\, \text{keV}$ and a maximum tube current of $60\,\mu\text{A}$.  For each specimen, 1440 X-ray projections were recorded during one full rotation of the specimen, using a flat panel detector. The frame rate of the detector ranged between 5 and 6 frames per second, depending on specimen height, with 20 frames averaged per projection. Subsequently, the X-ray projections were processed using Xact reconstruction software to obtain a 3D reconstruction of the specimen, an example of which is presented in \autoref{fig:concrete_cylinder_3D_view}. 

Although a uniform tube voltage and maximum tube current was used while scanning all the cylindrical castings, some issues typical of X-ray computed tomography had to be addressed:
\begin{itemize}
    \item {beam hardening leads to an increase of brightness close to the specimen surface. This effect is reduced by a numerical correction during reconstruction. The correction parameters have to be set manually on an image representative of the specimen, generally in the middle of the specimen. For this reason beam hardening correction is less efficient close to the top and bottom surfaces of the specimen.}
    \item{beam diffusion at specimen edges, leading to increased brightness at specimen edges and a halo-effect outside of the surface of the specimen. This effect was strongly reduced by tilting the specimen at an angle of $10^{\,\circ}$ with respect to the rotation axis of the rotation table.}
    \item{ring artifacts arise during reconstruction, yielding concentric rings centred with the rotation axis. These rings are attenuated numerically by the reconstruction software via an operator-applied threshold, compromising contrast, and experimentally by the automatic application of small random transversal shifts to the specimen with respect to the rotation axis, different for each image.}
\end{itemize}

In addition, a contrast-enhancing filter was used in the Xact software, resulting in a user-adjusted compromise between sharpness and contrast of the reconstruction. As a result, manual specimen-specific operator interventions introduced noticeable inter-sample variations in the resulting dataset.

The reconstruction  was obtained as single channel 16-bit unsigned integer 3D-image stack with voxel values ranging between $0$ and $65535$ with low values corresponding to the rarer phase such as air or voids and high values corresponding to the denser phases such as mortar and aggregates. Each sample consists of a 3D volumetric representation, comprising a stack of 2D cross-sectional images of the casting, oriented perpendicular to the longitudinal axis of the cylinder. The reconstructions were subsequently re-scaled using bilinear interpolation to achieve a resolution of 1024x1024 per slice, resulting in an approximate voxel size of $93\, \mu\text{m}$. For clarity, the term \emph{sample} refers to the XCT reconstruction corresponding to a specific cylindrical casting. The 2D images are referred to as \emph{slices} in this work. Any individual slice or part of it, when considered independently of the sample it originates from, is referred to as an \emph{image}. 

It is also to be noted that the neural network architecture employed in the present work processes data as 2D images rather than as a 3D volume, therefore, in the subsequent sections, we refer to each component within a slice as a \emph{pixel}, although it originally corresponds to a \emph{voxel} in the volumetric data.


\subsection{Preprocessing}
\label{subsec:pre-processing_for_segmentation}

\begin{figure}[h!]
    \centering
    \includegraphics[width=0.60\linewidth]{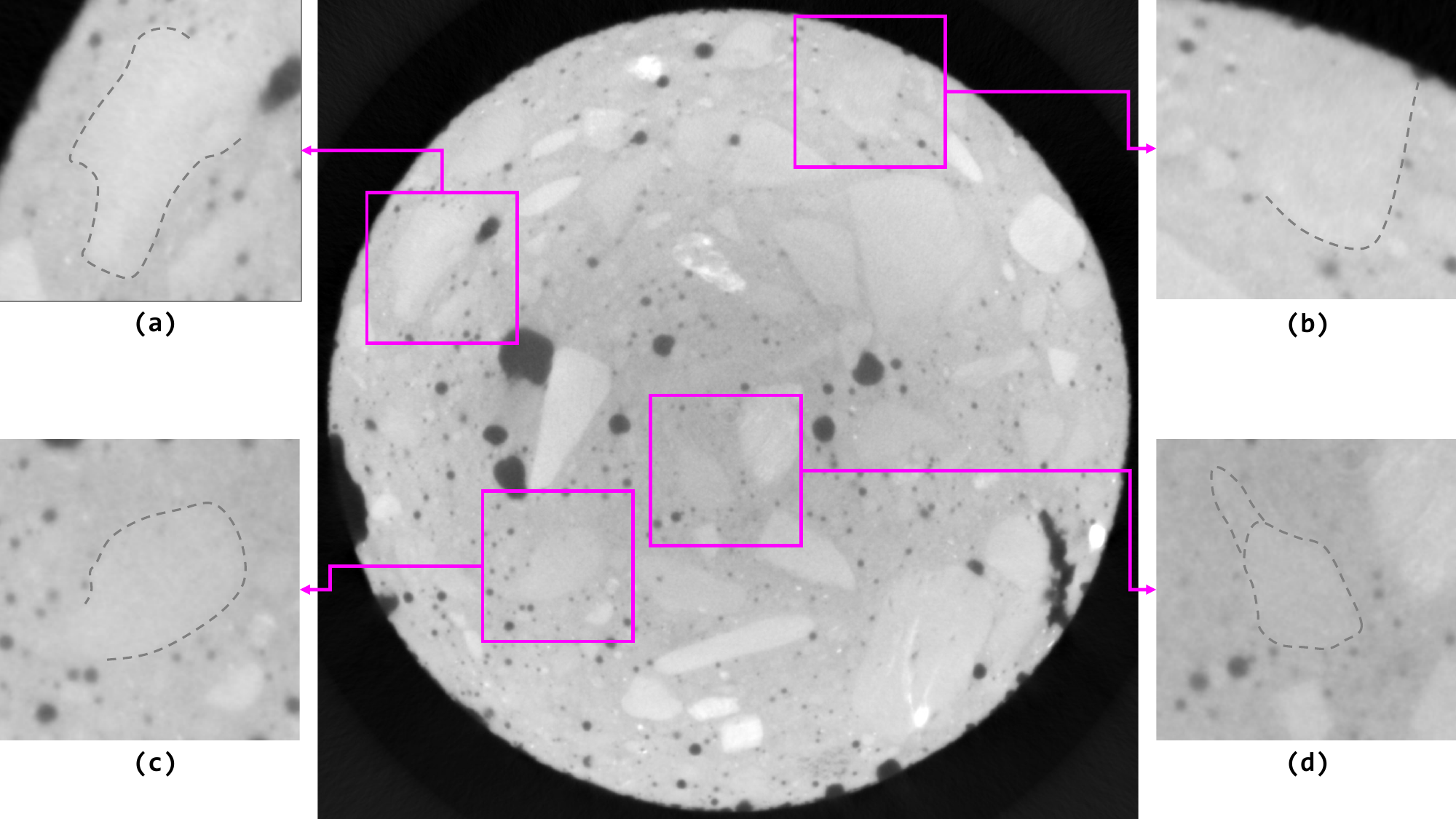}
    \caption{Example of low-contrast regions in an XCT image of concrete. (a)–(c) illustrate aggregates with only part of the           aggregate-mortar interface being clearly discernible (indicated by \textcolor{gray}{\textbf{-~-}}). (d) highlights              ambiguity in determining the aggregate-mortar interface on a part of the interface.}
    \label{fig:low_contrast_regions}
\end{figure}

\begin{figure}[ht!]
\centering
    \includegraphics[width=0.8\linewidth]{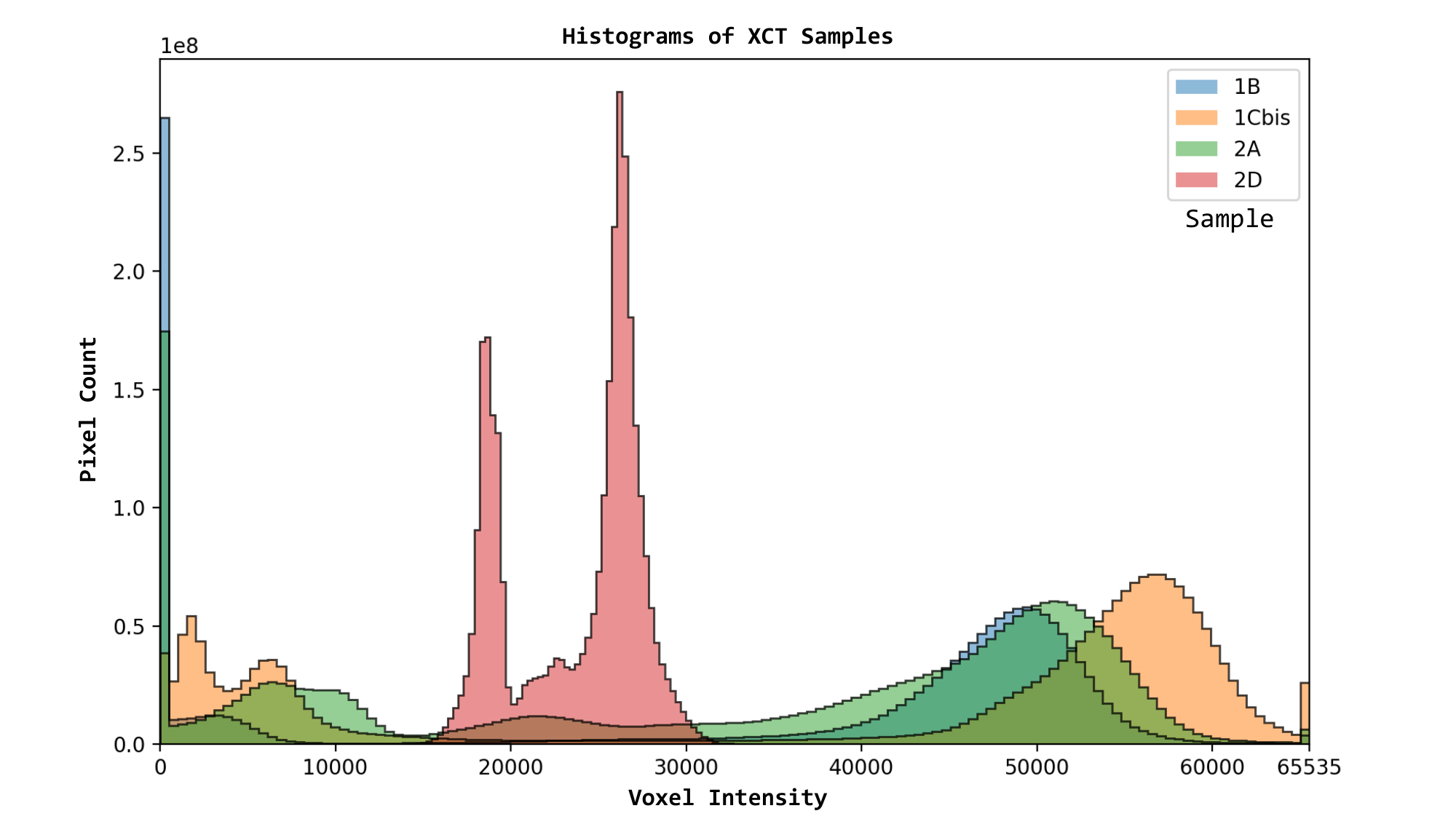}
    \caption{Inter-sample variation of voxel intensity depicted using histograms. For each sample, the peak to the left corresponds     to the porosity/air phase, while the peak to the right represents the aggregate and mortar phases in conjunction.}
    \label{fig:inter-sample_intensity_profile}

\centering
\includegraphics[trim=0 0 0 0, width=0.8\linewidth]{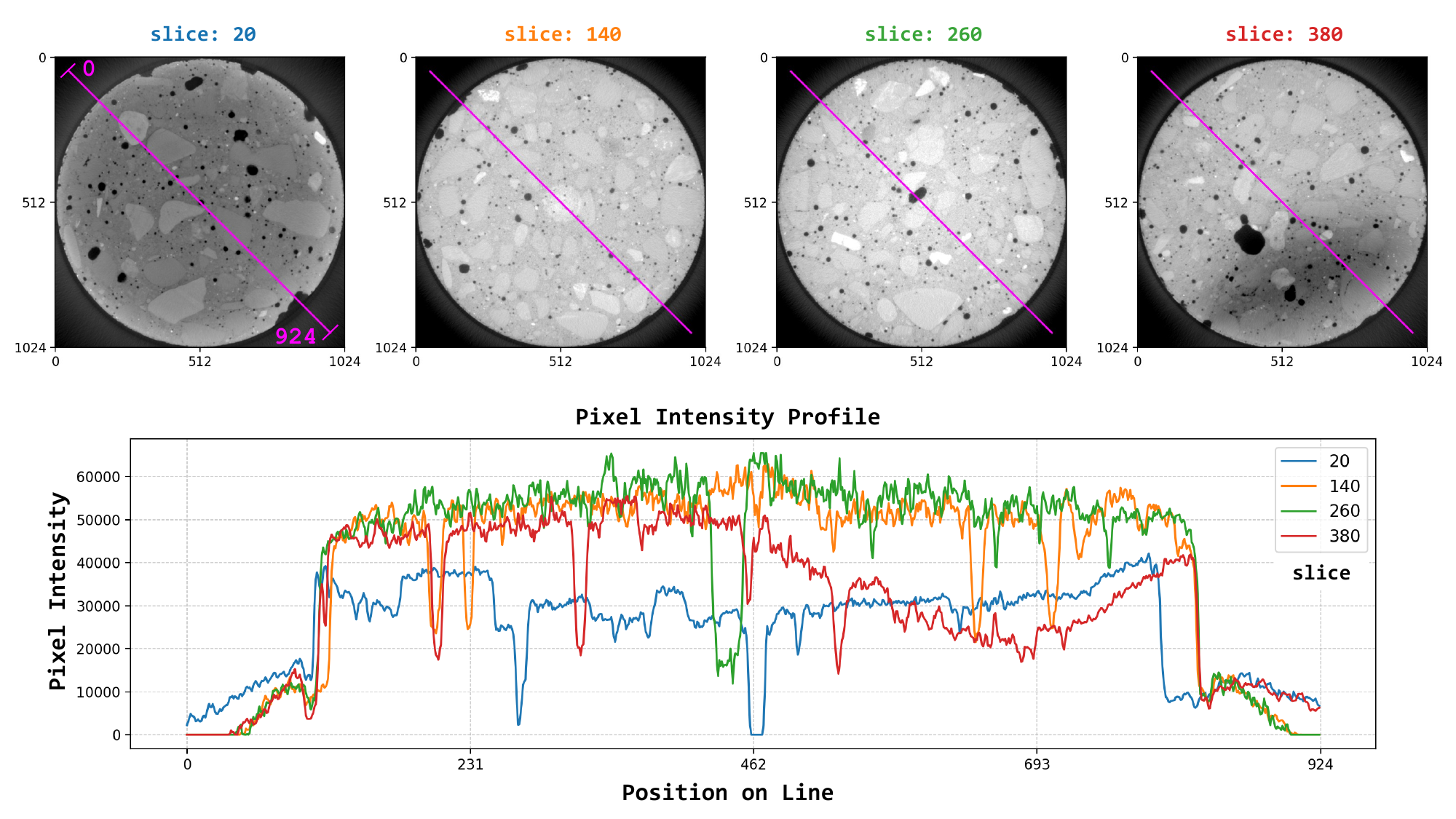}

\caption{Illustration of variations in pixel values typically observed in all samples shown using selected slices from one sample and their intensity profile across the slice produced using line-intensity profiling. The pixel value along the magenta line in each slice (top row) are plotted below. Slice numbers corresponding to their position along the sample's height. }
\label{fig:intra-sample_intensity_profile}

\end{figure}

The similar X-ray attenuation coefficients of aggregates and mortar resulted in low-contrast XCT samples, making it difficult to distinguish between the two materials, as shown in \autoref{fig:low_contrast_regions}. This issue is highlighted in the histograms of voxel intensity of samples, presented in \autoref{fig:inter-sample_intensity_profile}, where each sample exhibits two distinct peaks: the left peak corresponds to porosity or air, while the right peak represents the combined aggregate and mortar phase. This overlap complicates the separation of mortar and aggregate phases where ideally, for multiphase materials, one would expect a multimodal distribution with a distinct mode for each phase.

Owing to the inherent heterogeneity of concretes, differences in tomography devices, reconstruction software and operator-specific preferences, it is challenging to eliminate all artifacts and achieve consistent scans across all specimens. Inconsistencies in the concrete XCT dataset include gradual variation of brightness both within slices and between slices of any given sample, a phenomenon observed across all samples. As illustrated in \autoref{fig:intra-sample_intensity_profile} for one sample, the slices towards the top of the cylinder, indicated with a lower number (e.g. slice 20), exhibit lower overall intensity but have relatively high pixel intensities at the periphery of the cylindrical cross-section. As one progresses further along the sample (e.g. slice 140 and 260), overall pixel intensity gradually increases. In the slices towards the bottom of the cylinder (e.g. slice 380), brightness becomes uneven across the cross-section.

\begin{figure}[h!]
\centering
\includegraphics[trim=0 0 0 0, width=0.6\linewidth]{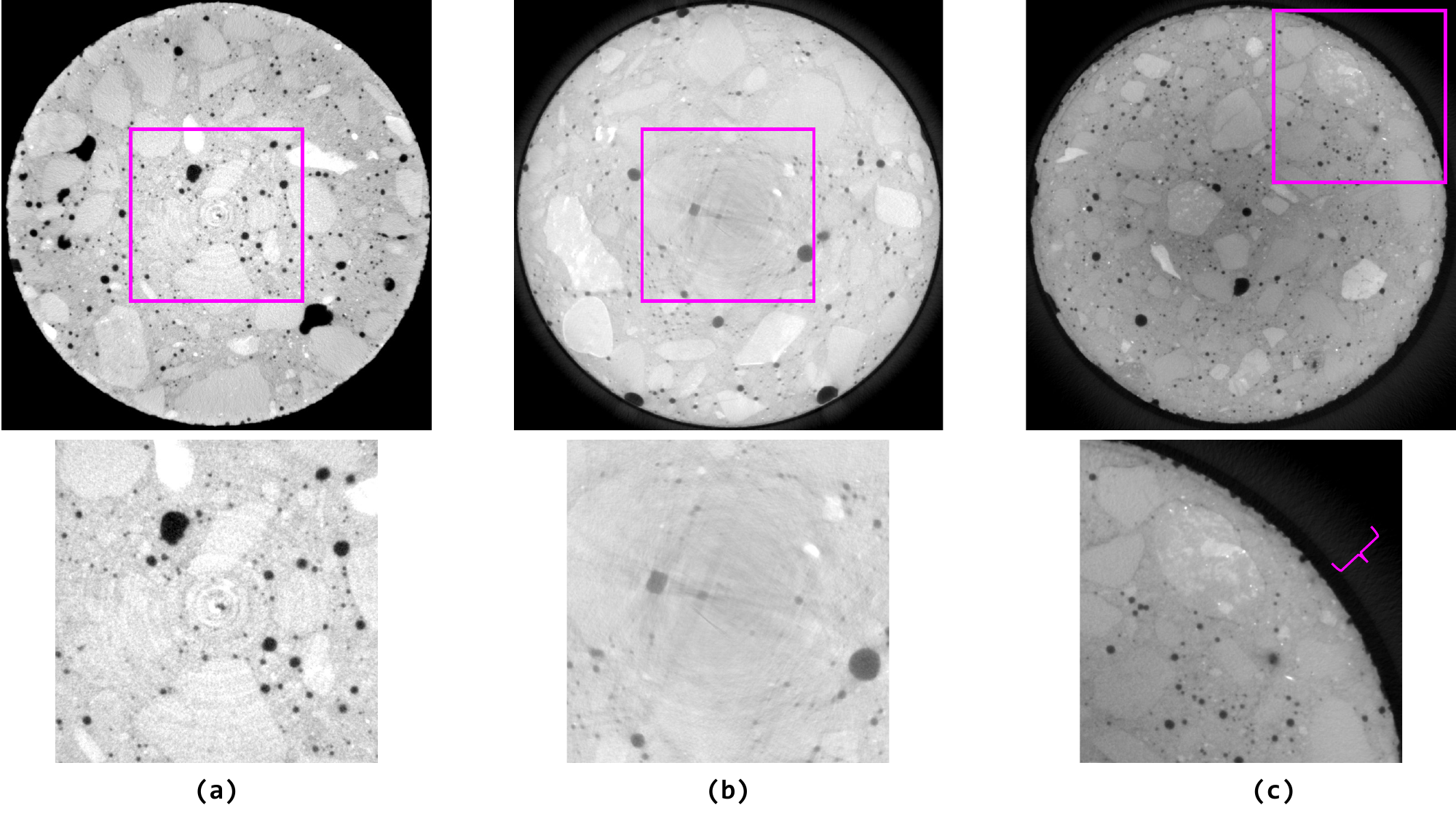}
\caption{Examples of concentric ring artifact (a), and blurring (b) seen near the centre of the cylinder and halo-effect (c), where regions around the sample show a bright band similar to a halo where no solid material is present.}
\label{fig:artifacts}

\centering
\includegraphics[trim=0 180 0 0, width=0.8\linewidth]{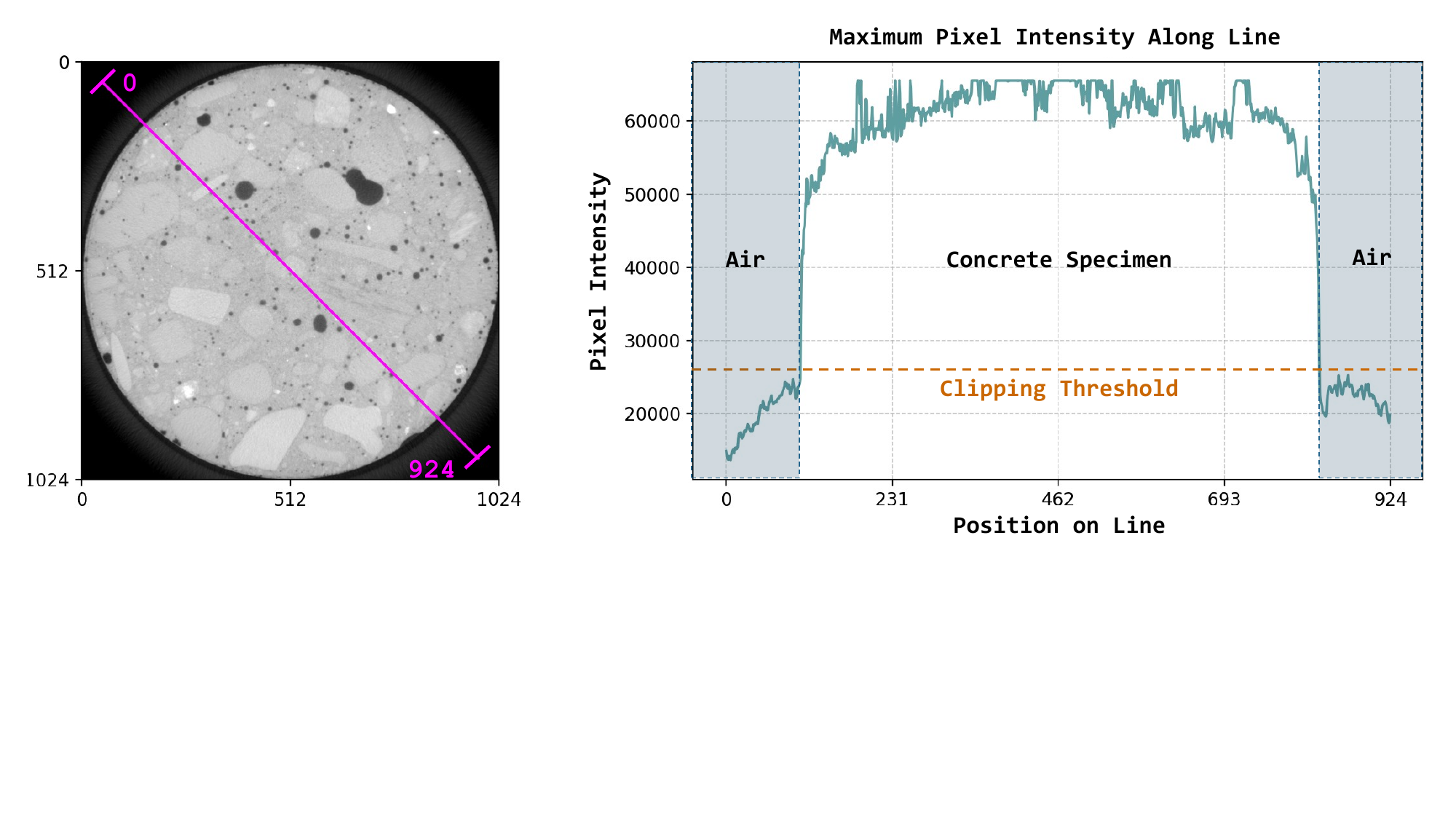}

\caption{Maximum pixel intensity measured along the magenta line (left image) for the entire sample (along the cylinder height) is plotted (right). Voxels with intensities below the clipping threshold (indicated by \textcolor{orange}{\textbf{-~-}}) are classified as part of the porous phase and are set to zero.}
\label{fig:low_intensity_pixels_line_intensity_profile}

\centering
\includegraphics[trim=0 0 0 0, width=0.75\linewidth]{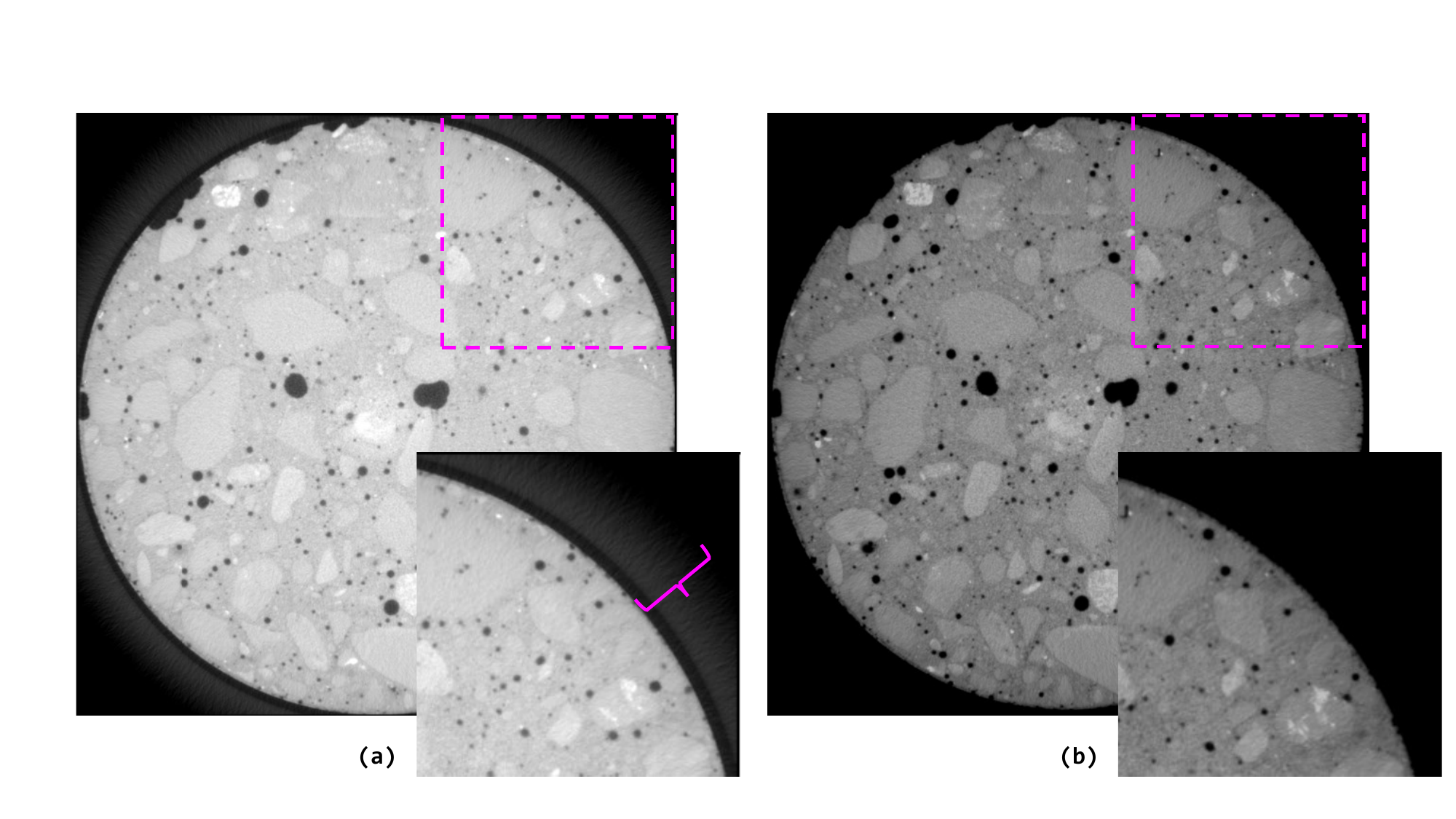}

\caption{Example of an original image (a) and its preprocessed version (b). Notably, the halo-like artifacts visible in the original are effectively removed after preprocessing. Image (a) is presented in greyscale by setting the greyscale thresholds to the range [0, 65536]. Image (b) is presented by setting the greyscale thresholds to the minimum and maximum values of the sample to which it belongs after standardization.}
\label{fig:standardized_image}
\end{figure}

Other artifacts include concentric bands of high- and low-intensity rings, loss of detail at the centre of the slices manifesting as blurred regions and a halo-effect around the sample in regions occupied by air as illustrated in \autoref{fig:artifacts}(a) - (c) respectively.

To attain uniformity in the XCT data, first, the XCT scans were visually inspected and slices exhibiting severe artifacts were discarded such as those exhibiting concentric circle artifacts and loss of detail as in \autoref{fig:artifacts}(a)-(b), or with uneven brightness across the slice such as slice 380 in \autoref{fig:intra-sample_intensity_profile}. The halo-effect was addressed by setting all pixels belonging to the porosity and air phase to zero for each sample to only retain data pertaining to the regions occupied by the solid phase, guided by line-intensity profiles, as illustrated in \autoref{fig:low_intensity_pixels_line_intensity_profile}. This was performed individually for each sample and the intensity threshold was manually chosen to be such that non-zero voxel values belonging only to non-solid phase is set to zero.

Finally, for all samples, each slice was individually standardized to have a mean pixel value of $0$ and a standard deviation of $1$, to mitigate the inter- and intra-sample intensity variations as demonstrated in \autoref{fig:inter-sample_intensity_profile} and \autoref{fig:intra-sample_intensity_profile}, respectively. Example of an image obtained after preprocessing is presented in \autoref{fig:standardized_image}. The images thus obtained formed the basis for the training of the segmentation model and subsequent prediction.

\section{Unsupervised Semantic Segmentation}
\label{sec:unsupervised_segmentation}

\subsection{Segmentation Model}
\begin{figure}[htb!]
\centering
\includegraphics[width = 0.8\textwidth]{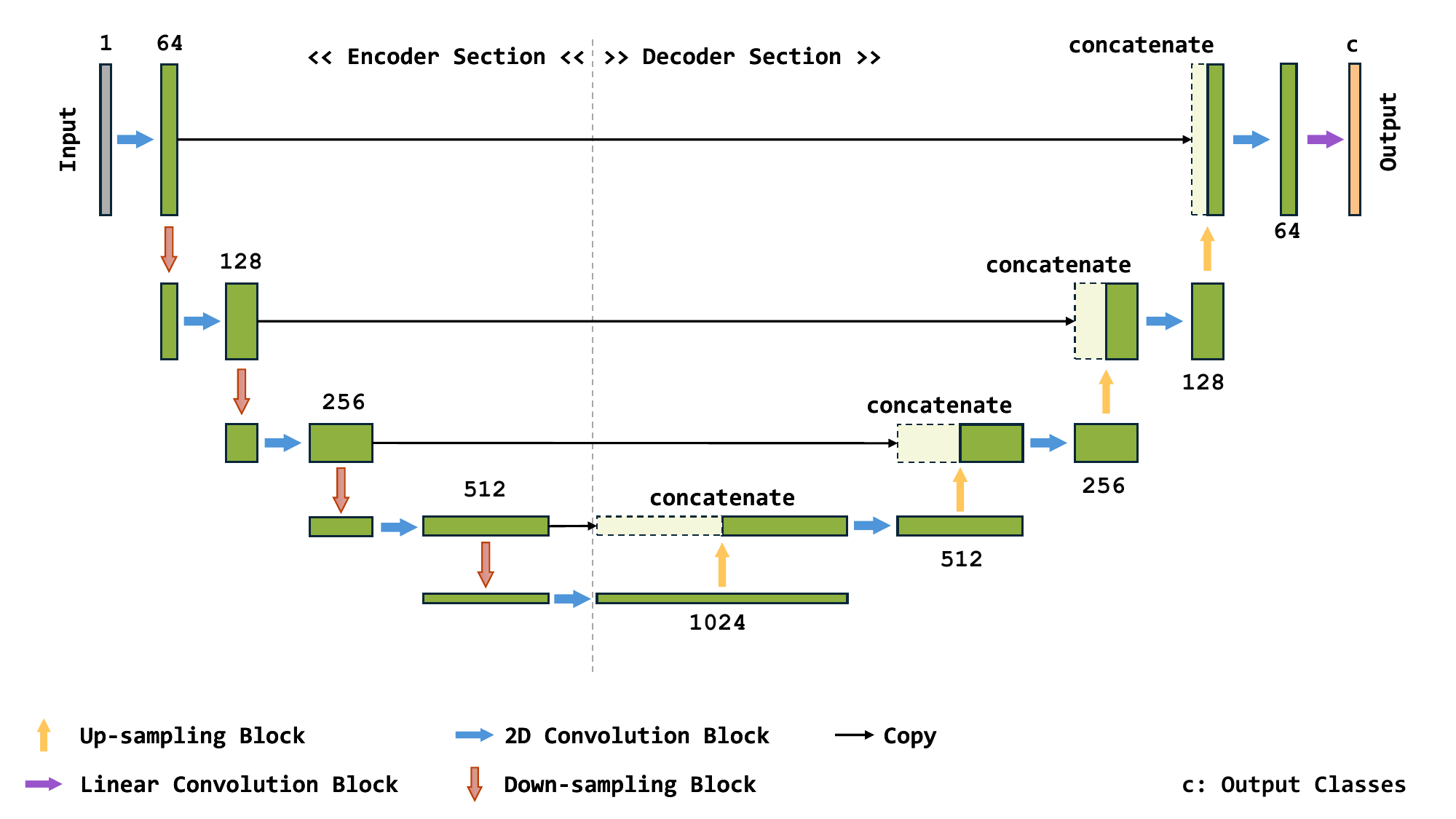}
\caption{The U-Net model architecture. Green rectangles correspond to feature tensors. Input (grey rectangle) is the greyscale input image. Numbers above or below each rectangle indicate the number of feature channels (or layers or depth) of each tensor, which is also schematically represent the width of the rectangle, whereas the height of each rectangle illustrates the change in spatial dimension (height, width) along the network. Output tensor (in yellow) in this case has $c=3$, i.e. $3$ feature channels, one for each output class. }
\label{fig:U-Net}
\end{figure}


\begin{figure}[h!]
\centering
\includegraphics[trim = 0 0 0 0, width=0.80\textwidth]{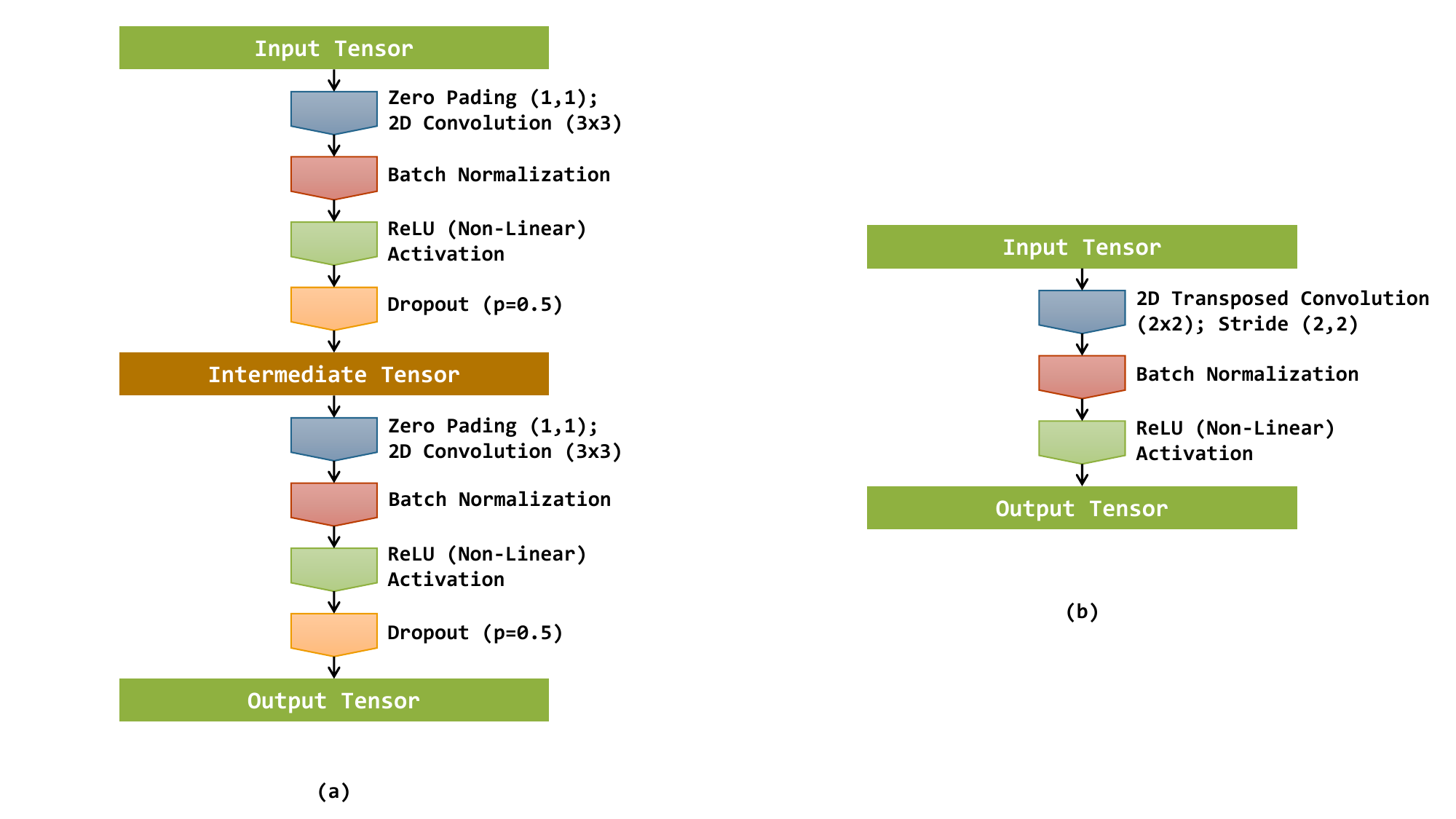}
\caption{Convolution block (a) and Upsampling block (b). Padding size, kernel size, stride and dropout probability are given in brackets.}
\label{fig:conv_block_upsampling_block}

\centering

\includegraphics[trim = 0 50 0 0, width=0.8\textwidth]{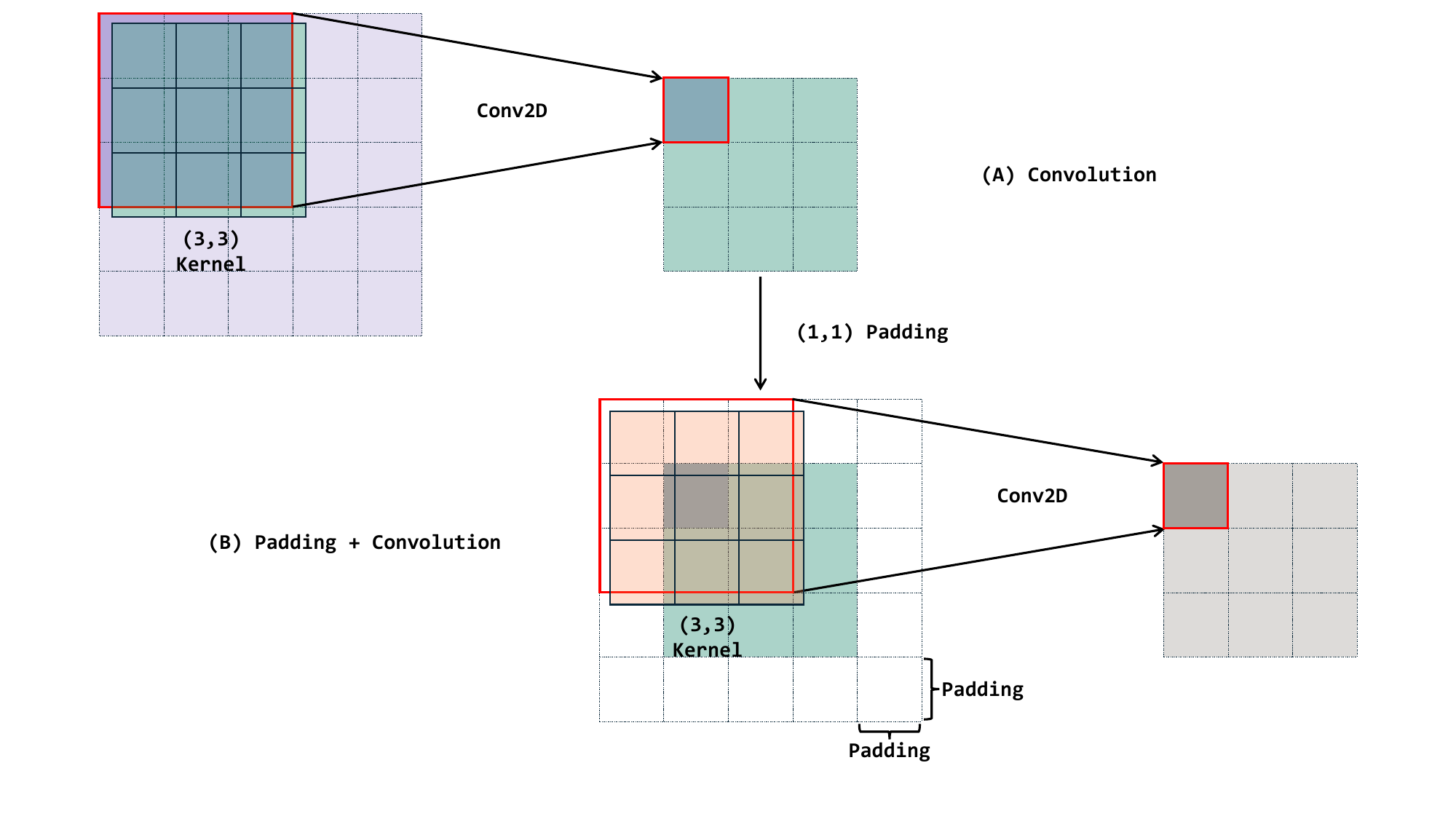}
\caption{A single-channel input tensor undergoes convolution using kernels of size (3,3) and a stride of (1,1) in (A). Applying the convolution without padding results in a smaller output tensor. By applying appropriate padding prior to convolution, as illustrated in (B), the output tensor maintains the original input size.}

\label{fig:convolution_operation}
\end{figure}

While segmentation models using newer architectures such as vision transformer (ViT) \cite{dosovitskiy_image_2020} outperform CNN-based models \cite{lecun1989backpropagation} in several aspects, it is also observed that ViT-based models require substantially more data to train compared to CNNs \cite{gai2024comparing}, offer slower convergence, and have a higher computational footprint and often require pre-training \cite{gai2024comparing, takahashi2024comparison,rosy2025replacing}. The data-frugality of CNNs is partly by design, which imparts an inductive bias allowing models using it to learn from spatial data with relative ease \cite{dAscoli2022convit}. Furthermore ViT-based models have quadratic complexity with respect to input size \cite{liu2022convnet} and often exhibit a deficiency in resolving fine-grained local textures and pixel-level boundaries due to their weak local constraints, while CNNs inherently preserve the spatial hierarchies essential for precise boundary delineation  \cite{li2023transforming}. Considering these factors and the present use-case, the use of CNN-based architecture in this work was retained.

The present work adopts a variant of the CNN-based U-Net architecture \cite{ronneberger2015u} (see \autoref{fig:U-Net}), expanding upon the original approach in \cite{kanezaki2018unsupervised} which utilized a simple feed-forward CNN. Although initially designed for biomedical image segmentation tasks, U-Net's established efficacy has led to the development of numerous architectural variants over time \cite{siddique2021u}. The architecture consists of alternating down-sampling and up-sampling segments, where localized convolutional operations within each block are central to extracting rich spatial features from the input images. Consequently, this section provides a concise overview of the core operations underpinning the U-Net framework.

The input to every operation of the U-Net is a tensor $\tensor{T} = ( \tensorComp{t}_{n,h,w,f}) \in \realNum^{N\times H \times W \times F}$, where $N$ denotes the number of training samples, so-called batch size (e.g. distinct images, which are the inputs to the very first convolution operation in this model),  $H$ and $W$ represent the spatial dimensions - height and width - of the input tensor (e.g. the spatial dimensions of the images in a 2D framework) and $F$ indicates the feature dimension (e.g., the number of channels, such as $F=1$ for greyscale images and $F=3$ for RGB images).

Each convolution operation used in this model is part of a five-step sequence occurring twice in each convolution block as illustrated in \autoref{fig:conv_block_upsampling_block}(a). As a part of this sequence, the tensors are first expanded along the spatial dimensions using the process of \emph{padding} and specifically, zero-padding in this case, where, $\scalar{s}_r$ rows and $\scalar{s}_c$ columns or $0$s are added around the tensor. This ensures that the output tensor of the convolution operation has the same height and width as the input tensor $\tensor{T}$, since convolution operation, as used in this work, would otherwise reduce the spatial dimensions. The padded tensor $\tensor{T}^\mathrm{0} = (\tensorComp{t}^\mathrm{0}_{n,h,w,f}) \in \realNum^{N\times (H+2\mathrm{s}_r) \times (W+2\mathrm{s}_c) \times F}$ is thus given as 
\begin{equation}
    \tensorComp{t}^\mathrm{0}_{n,h,w,f} = 
    \begin{cases}
    \tensorComp{t}_{n,h - \scalar{s}_r,w - \scalar{s}_c,f} & \text{ for } \scalar{s}_r < h \leq H + \scalar{s}_r \text{ and}\, \scalar{s}_c < w \leq W + \scalar{s}_c \text{ ;}\\
    0\, & \text{ otherwise}
    \end{cases}
\end{equation}
\noindent where $\scalar{s}_r \geq 0 \text{ and } \scalar{s}_c \geq 0$.

Zero-padding is followed by the convolution operation. Each convolution operation introduces a unique convolution kernel tensor $\tensor{K} = (\tensorComp{k}_{h,w,f,m}) \in \realNum^{H_k \times W_k \times F \times M}$ and a unique bias vector ${\tensor{b} = (\tensorComp{b}_m) \in \realNum^{M}}$ with $H_k$ and $W_k$ specifying its height and width and $M$ denoting the desired size of the feature dimension of the output. Given integer valued stride $\scalar{s}_H$ along height and $\scalar{s}_W$ along width, a convolution operation results in an output tensor $\tensor{T}^{\mathrm{c}} = (\tensorComp{t}^\mathrm{c}_{n,h,w,m}) \in \realNum^{N \times H_c \times W_c \times M}$ with all components defined as
\begin{equation}
    \tensorComp{t}^{\mathrm{c}}_{n,h,w,m} = \sum_{f = 1}^F \sum_{q=1}^{H_k}\sum_{r=1}^{W_k} \,\tensorComp{k}_{q,r,f,m} \tensorComp{t}^{\mathrm{0}}_{n, \scalar{s}_H(h-1)+q, \scalar{s}_W(w-1)+r,f} + \tensorComp{b}_m
    \label{eqn:convolution}
\end{equation}
\noindent and spatial dimensions given as $H_c = (\frac{H  + 2\scalar{s}_r - H_k}{\scalar{s}_H} + 1)$ and $W_c =  (\frac{W + 2\scalar{s}_c - W_k}{\scalar{s}_W}  + 1)$. The convolution operation is carried out under the constraints $(H  + 2\scalar{s}_r - H_k) = a \scalar{s}_H$ and $(W  + 2\scalar{s}_c - W_k) = b \scalar{s}_W$, where $a \text{ and } b$ are whole numbers. Strides $\scalar{s}_H \geq 1$ and $\scalar{s}_W  \geq 1$ are defined as the number of pixels the convolution kernel travels at a time along height and width respectively. And in the present case, $H_k = W_k  = 3$, $\scalar{s}_r = \scalar{s}_c = 1$ and $\scalar{s}_H = \scalar{s}_W = 1$ ensures $H_c = H$ and $W_c = W$. In the encoder section, $M=64$ in the first convolution block and is doubled in every subsequent convolution block, while in the decoder section, the first convolution block has $M=512$ and is halved in every subsequent convolution block.

It is to be noted that convolution kernels with either $H_k > 1$ or $W_k > 1$ enable extraction of spatial information from images. As a general rule of thumb, larger kernels allow for information extraction over a larger spatial region, albeit, at a higher computational cost. For more information on convolution operations and CNNs, one may refer to \cite{Goodfellow-et-al-2016, bishop2023deep}. A visual summary of the convolution process is presented in \autoref{fig:convolution_operation}.

In this work, unlike the original U-Net model, convolution operations are followed by \emph{batch-normalization} \cite{ioffe2015batch} prior to the application of a non-linear activation function in order to improve training stability and performance. Specifically, given the output of the convolution operation $\tensor{T}^\mathrm{c}$, batch-normalization is applied to scale all tensor components. This is followed by the application of a non-linear activation function, specifically the \emph{rectified linear unit} (ReLU), which sets all negative tensor components to zero. These two operations in combined form is given as
\begin{equation}
    \tilde{\tensorComp{t}}^\mathrm{c}_{n, h, w, m} = \mathrm{max}(0, \frac{\tensorComp{t}^\mathrm{c}_{n, h, w, m}  - \mu_{m}^\mathrm{c}}{\sigma_{m}^\mathrm{c}}\tensorComp{\gamma}_m + \tensorComp{\beta}_m)
    \label{eqn:Batch_norm_relu}
\end{equation}
\noindent where $\tensorComp{\gamma}_m \in \tensor{\gamma} \in \realNum^{M}$ and $\tensorComp{\beta}_m \in \tensor{\beta} \in \realNum^{M}$ are learnable parameters of the batch-normalization operation \cite{ioffe2015batch} and $\mu_m^\mathrm{c}$ and $(\sigma_m^\mathrm{c})^2$ denote the mean and variance, respectively, of all components across the batch for the $m$-th feature computed as 
\begin{align*}
    \mu_m^\mathrm{c} &=\frac{1}{N H_c W_c} \sum_{n=1}^N \sum_{h=1}^{H_c} \sum_{w=1}^{W_c}  \tensorComp{t}^\mathrm{c}_{n,h,w,m} \\
    (\sigma_m^\mathrm{c})^2 &=\frac{1}{N H_c W_c} \sum_{n=1}^N \sum_{h=1}^{H_c} \sum_{w=1}^{W_c} (\tensorComp{t}^\mathrm{c}_{n,h,w,m} - \mu_m^\mathrm{c})^2 
\end{align*}
\noindent thereby resulting in $\tilde{\tensor{T}}^{\mathrm{c}} = (\tilde{\tensorComp{t}}^\mathrm{c}_{n,h,w,m}) \in \realNum^{N \times H_c\times W_c\times M}$.

Furthermore, this work introduces an additional modification to the convolutional block of the original U-Net architecture through the incorporation of \emph{dropout} regularization, aimed at enhancing the model's generalization capabilities \cite{srivastava2014dropout, tompson2015efficient}. Dropout operates by randomly setting individual tensor components to zero with a probability $p_d$ following a Bernoulli distribution $\mathcal{B}(p_d)$. Hence, for the tensor $\tilde{\tensor{T}}^{\mathrm{c}}$ and a mask tensor $\tensor{M} = (\tensorComp{m}_{n, h, w, m}) \in \{0,1\}^{N\times H_c\times W_c \times M}$ with $\tensorComp{m}_{n, h, w, m} \sim \mathcal{B}(p_d)$, the result of dropout $\tilde{\tensor{T}}^{\mathrm{c,d}} = \tilde{\tensorComp{t}}^{\mathrm{c,d}}_{n,h,w,m} \in \realNum^{N, H_c, W_c, M}$ can be written as 
\begin{equation}
    \tilde{\tensorComp{t}}^{\mathrm{c, d}}_{n,h,w,m} = \tilde{\tensorComp{t}}^{\mathrm{c}}_{n,h,w,m}\tensorComp{m}_{n,h,w,m}\,.
\end{equation}
\noindent Here, $p_d$ is set to $0.5$, which is generally considered optimal \cite{srivastava2014dropout}. The resulting tensor $\tilde{\tensor{T}}^{\mathrm{c,d}}$ is the output of the sequence of five operations discussed above. 

In the encoder section of the model, each convolutional block is followed by a down-sampling operation (red arrows in \autoref{fig:U-Net}), to reduce the spatial dimensions of the tensors resulting from the preceding convolution block. Unlike the original U-Net architecture, which employs \emph{max-pooling}, this work utilizes \emph{average-pooling} $(\mathrm{AvgPool})$ for down-sampling. This modification showed improvement in segmentation performance during experiments. $\mathrm{AvgPool}$ is akin to computing moving average over the components of an input tensor using a convolution operation with a non-trainable moving kernel $\tensor{K}^{\mathrm{1}} = (\tensorComp{k}^{\mathrm{1}}_{h,w,f,m}) \in \{1\}^{H_1 \times W_1 \times F \times F}$ similar to (\ref{eqn:convolution}) but without an associated bias term. Given an input tensor $\tensor{T} = \tensorComp{t}_{n,h,w,f} \in \realNum^{N, H, W, F}$, this results in $\bar{\tensor{T}} =(\bar{\tensorComp{t}}_{n,h,w,f}) \in \realNum^{N, \bar{H}_A, \bar{W}_A, F}$ with components 
\begin{equation}
    \bar{\tensorComp{t}}_{n,h,w,f} = \frac{1}{H_1\,W_1}\sum_{q=1}^{H_1}\sum_{r=1}^{W_1} \tensorComp{t}_{n, \scalar{s}_H(h-1) + q,\scalar{s}_W(w-1) + r,f} 
\end{equation}
\noindent and spatial dimensions $\bar{H}_A =  (\frac{H - H_1}{\scalar{s}_H} + 1) \, \text{ and } \bar{W}_A = (\frac{W - W_1}{\scalar{s}_W} + 1)$. In this case, setting $H_1 = W_1 = 2$ and $\scalar{s}_H = \scalar{s}_W = 2$ leads to $\bar{H}_A = H/2$ and $\bar{W}_A = W/2$.

\begin{figure}[h!]
\centering
\includegraphics[trim = 0 100 0 0, width=0.8\textwidth]{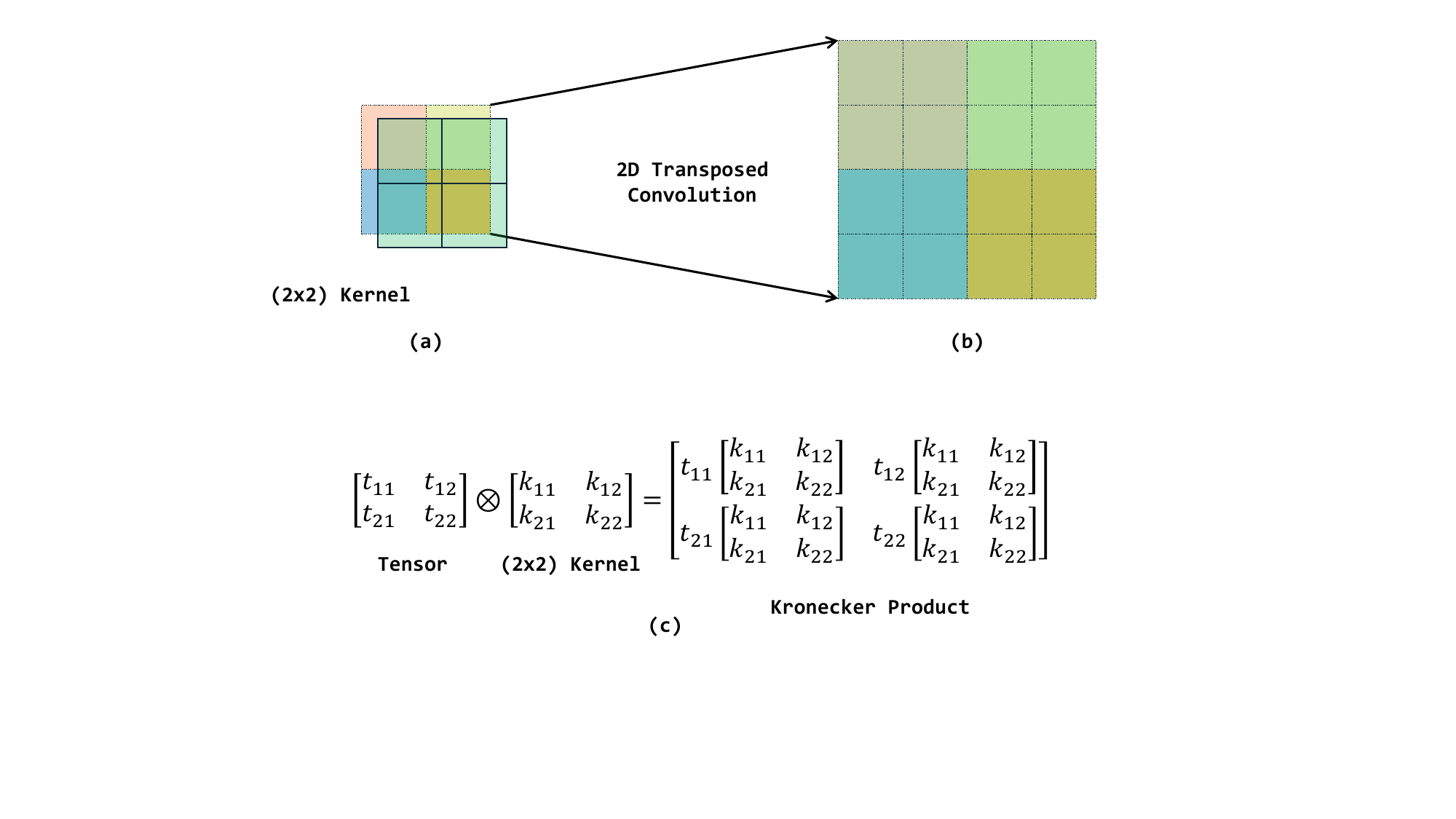}
\caption{2D transposed convolution on a single channel input, i.e. $F = 1$ with kernel size $H^T_k = W^T_k = 2$ and stride $\scalar{s}_H = \scalar{s}_W = 2$ resulting in tensor (b) that is twice the size of the input (a). The same can be represented using a Kronecker Product for a case when $\scalar{s}_H = H^T_k$ and $\scalar{s}_W = W^T_k$ individually for each channel and is presented in (c).}
\label{fig:transposed_convolution}
\end{figure}

In the decoder section, each convolution block is preceded by an up-sampling block and a concatenation operation indicated using yellow arrows and black arrows respectively in \autoref{fig:U-Net}. The up-sampling block, (see \autoref{fig:conv_block_upsampling_block}(b)), consists of a sequence of three operations and is used to increase the spatial dimensions of any given tensor starting with a transposed convolution  \cite{zeiler2010deconvolutional} (see \autoref{fig:transposed_convolution}(a)-(b)) to counteract the down-sampling. 

Formally, each transposed convolution operation introduces a unique transposed convolution kernel tensor $\tensor{K}^{\mathrm{T}} = (\tensorComp{k}^{\mathrm{T}}_{h,w,f,m}) \in \realNum^{H^T_k \times W^T_k \times F \times M}$ and a bias term $\tensor{b}^{\mathrm{T}} = (\tensorComp{b}^{\mathrm{T}}_m) \in \realNum^{M}$ with $H^T_k$ and $W^T_k$ specifying the height and width of the kernel and $M$ denoting the desired size of the feature dimension of the output. Given an input tensor $\tensor{T} = \tensorComp{t}_{n,h,w,f} \in \realNum^{N \times H \times W \times F}$, components of the transposed convolution output $\tensor{T}^{\mathrm{T}} = (\tensorComp{t}^{\mathrm{T}}_{n,h,w,m}) \in \realNum^{N\times H_T \times W_T \times M}$ are given as 
\begin{equation}
    \tensorComp{t}^{\mathrm{T}}_{n, h, w ,m} = \sum_{i=1}^{H} \sum_{j=1}^{W} \sum_{f=1}^F \sum_{q = 1}^{H^T_k} \sum_{r = 1}^{W^T_k} \tensorComp{t}_{n,i,j,f}\, \tensorComp{k}^{\mathrm{T}}_{q,r,f,m} \, \delta_{h, \scalar{s}_H(i-1)+q}\, \delta_{w, \scalar{s}_W(j-1)+q} + \tensorComp{b}^{\mathrm{T}}_m \,.
    \label{eqn:transposed_conv}
\end{equation}
\noindent Here $\delta_{i,j}$ is the Kronecker delta. In the present work, $\scalar{s}_H = \scalar{s}_W = 2$ and $H^T_k = W^T_k = 2$ are used in order to obtain $H_{T} = 2H$ and $W_T = 2W$. For every transposed convolution block the number of output feature dimensions is set at $M=F/2$. 

In this particular case, where the stride and kernel sizes are equal, transposed convolution operation reduces to a Kronecker product between the slices of the input tensor and corresponding slices of the convolution kernel. To be more specific, given tensor slice $\tensor{I}_{n,f}\in \realNum^{H \times W}$  of a given data sample $n$ and feature channel $f$ of $\tensor{T}$ such that $\tensor{T} = \{\tensor{I}_{n,f} \,|\, n \in [1,2, \cdots N], \, f \in [1, 2, \cdots F]\}$ and  kernel slices $\tensor{K}^T_{f,m}$, for any given input and output feature channel $f$ and $m$ of $\tensor{K}^T$ respectively, such that $\tensor{K}^T = \{\tensor{K}^T_{f,m} \,|\, f \in [1,2, \cdots F], m \in [1,2, \cdots M]\}$, the result of the transposed convolution becomes $\tensor{T}^T = \{ \tensor{I}^{T}_{n,m} \}$, where each tensor slice $\tensor{I}^{T}_{n,m}$ is equivalent to the Kronecker product $\tensor{I}^{T}_{n,m} = \tensor{I}_{n,f} \otimes \tensor{K}^{T}_{f,m}$ (see \autoref{fig:transposed_convolution}(c)). The transposed convolution operation is followed by batch normalization and a non-linear activation using ReLU, identical to (\ref{eqn:Batch_norm_relu}). It is worth noting that, instead of transposed convolution, tensors could alternatively be up-sampled using interpolation. However, this configuration was not explored in the present work.

Following the up-sampling block and prior to each convolution operation in the decoder section, at each hierarchical level of the model, the up-sampled tensor is concatenated to the convolution output of equal spatial dimension from the encoder section (black arrows in \autoref{fig:U-Net}). For any two tensors $\tensor{T}^\mathrm{A} = (\tensorComp{t}^\mathrm{A}_{n,h,w,m_A}) \in \realNum^{N \times H \times W \times M_A}\, \text{, and } \tensor{T}^\mathrm{B} = (\tensorComp{t}^\mathrm{B}_{n,h,w,m_B}) \in \realNum^{N \times H \times W \times M_B}$, concatenated tensor $\tensor{T}^\mathrm{C} = (\tensorComp{t}^\mathrm{C}_{n,h,w,m_C}) \in \realNum^{N \times H \times W \times M_A+M_B}$ is given as 
\begin{equation}
    \tensorComp{t}^\mathrm{C}_{n, h, w, m_C} = 
\begin{cases}
\tensorComp{t}^\mathrm{A}_{n, h, w, m_C} & \text{for } 1 \leq m_C \leq m_A\, \text{;} \\
\tensorComp{t}^\mathrm{B}_{n, h, w, m_C - m_A} & \text{for } m_A < m_C \leq m_A + m_B\, .
\end{cases}
\end{equation}

The final block of the model is the linear convolution block (purple arrow in \autoref{fig:U-Net}). This block is composed of a single convolution operation, equivalent to (\ref{eqn:convolution}), but with $H_k = W_k = 1$, $\mathrm{s}_H = \mathrm{s}_W = 1$ and $M=3$ or $4$ depending on the experiment. The choice of kernel size and strides preserve the spatial dimension and do not necessitate use of padding prior to convolution. Furthermore, no batch normalization, non-linear activation or dropout is performed in this block.

The U-Net model for semantic segmentation employed in this work is hereafter denoted by $\, \mathcal{U}$ and represents a mapping between tensors. $\,\mathcal{U}$ is parametrized by a set of variables collectively referred to as ${\set{\theta}}$, which comprises all the trainable convolution kernels and their corresponding bias terms associated with the convolution and up-sampling blocks of the network and also the trainable parameters associated all the batch-normalization operations. Assuming the entire model has $L$ convolution operations and $L^{\mathrm{T}}$ transposed convolution operations , each with a kernel-bias pair $(\tensor{K}, \tensor{b})_i$ and $(\tensor{K}^{\mathrm{T}}, \tensor{b}^{\mathrm{T}})_{\mathrm{j}}$, and $(L + L^{\mathrm{T}}-1)$ batch-normalization operation, each with a pair of parameter vectors $(\tensor{\gamma}\,, \tensor{\beta})_k$, the parameter set ${\set{\theta}}$ is formally defined as $\set{\theta} = \{(\tensor{K}, \tensor{b})_i,\, (\tensor{K}^{\mathrm{T}}, \tensor{b}^{\mathrm{T}})_j,\, (\tensor{\gamma}\,, \tensor{\beta})_k |\, i \in [1, 2, \cdots, L]\,,  j \in [1, 2, \cdots, L^{\mathrm{T}}]\, \text{and } k \in [1, 2, \cdots, L+L^{\mathrm{T}}-1]\}$.
The values of these parameters are initially unknown and are determined during training. Finally, the notation $\, \mathcal{U}_{\set{\theta}}$ is used to refer to the model parametrized with a specific ${\set{\theta}}$.

\begin{figure}[h!]
\centering
\includegraphics[trim = 0 275 0 0, width=0.8\textwidth]{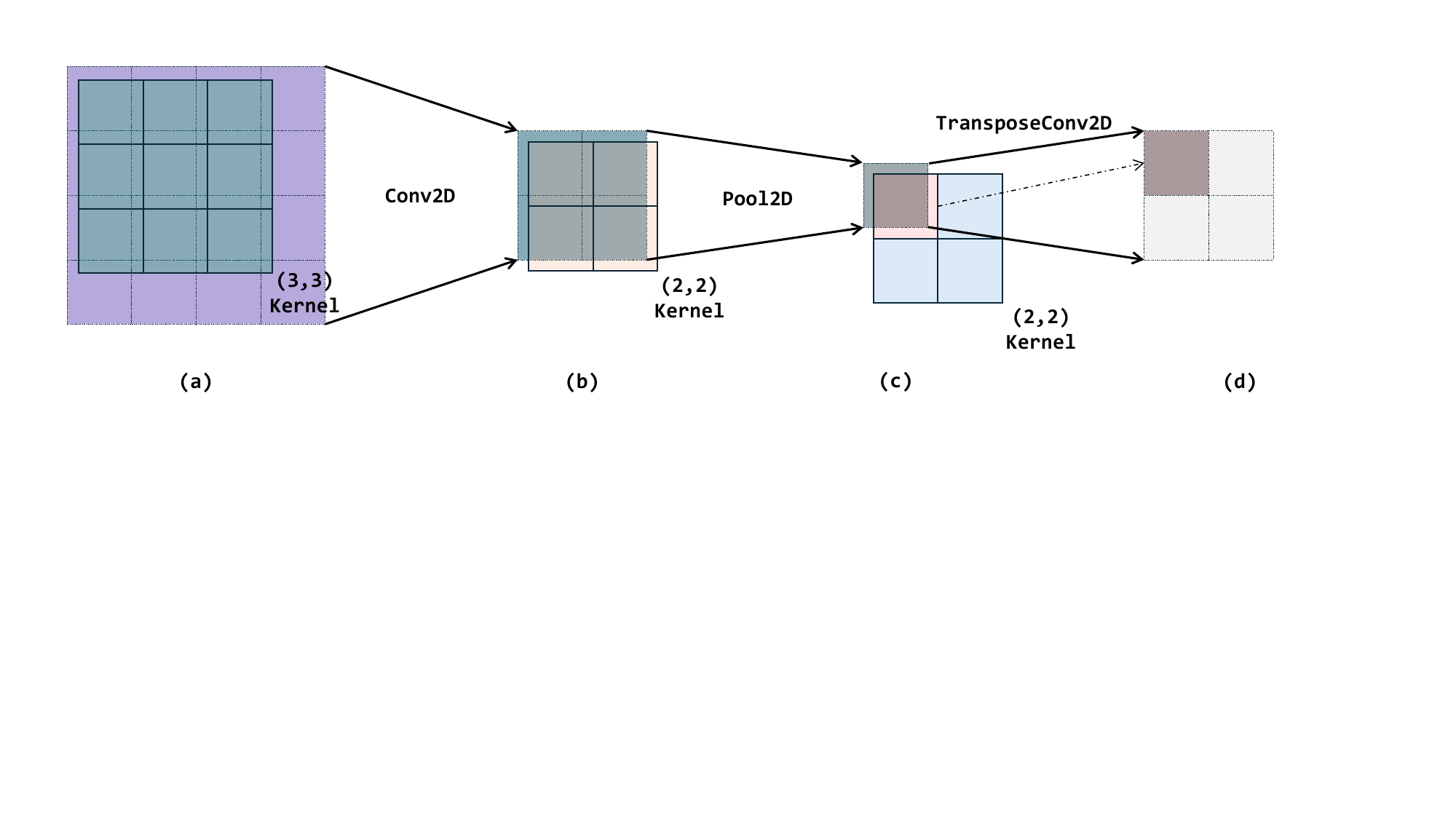}
\caption{Illustration of a receptive field. The highlighted pixel in (d) is the result of applying a convolution with a (3×3) kernel and stride (1,1) on the input in (a), followed by a pooling operation with a (2×2) kernel in (b), and a transposed convolution using a (2×2) kernel in (c). This pixel in (d) is influenced by a 4×4 region of the original input tensor in (a).}
\label{fig:receptive_field}
\end{figure}


\subsection{Model Input}


In this work, input to the first convolution operation of $\mathcal{U}$ is a tensor $\tensor{T} \in \realNum^{N \times H \times W \times 1}$. Here $N$ indicates the number of images being passed into the model at once and the last dimension of size $1$ indicates a single-channel/greyscale image. For convenience $\tensor{T}$ is here on defined as a set of images as $\tensor{T} = \{\tensor{I}_1, \tensor{I}_2, \cdots, \tensor{I}_N \}$ where $\tensor{I}_n = (\tensorComp{p}_{n,h,w}) \in \realNum^{H \times W}$ is a digital image composed of pixels $\tensorComp{p}_{n,h,w}$ \cite{gonzalez2018digital}. In the remainder of this section, the analysis is restricted to a single image; hence, the index $n$ has been omitted in favour of readability.

In practice, given the mapping $\tensor{I} \xmapsto{\mathcal{U}} \hat{\tensor{Y}} = (\hat{\tensorComp{y}}_{h,w,f}) \in \realNum^{H \times W \times F}$, each output component $\hat{\tensorComp{y}}_{h,w,f}$ at a given spatial location $(h, \,w)$ is influenced only by a unique subset of input image components  $\tensor{F}^{(h,w)} = \{\tensorComp{p}_{i,j}\}_{(h,w)} \subseteq \tensor{I}$, the so-called \emph{receptive-field} (RF) of a CNN, simplified example of which is illustrated in \autoref{fig:receptive_field}; for more information, one may refer to \cite{szegedy2016rethinking}. An RF is unique to each CNN architecture.

Therefore, to ensure compatibility of the input images with $\mathcal{U}$, it is required that for any image $\tensor{I}$ and output component $\hat{\tensorComp{y}}_{h,w,f}$, there exists a region $\tensor{F}^{(h,w)} \subseteq \tensor{I}$ that is compatible with the RF of $\,\mathcal{U}$. In the limiting case where the output tensor is spatially collapsed, i.e. $\hat{\tensor{Y}} \in \realNum^{1 \times 1 \times F}$, the input must correspond to the receptive field, i.e. $\tensor{I} = \tensor{F}^{(1,1)} = \{\tensorComp{p}_{i,j}\}_{(1,1)}$. Consequently, tensors with spatial dimensions smaller than the receptive field are not supported by $\,\mathcal{U}$.

\subsection{Supervised Semantic Segmentation}

Given a set of discrete variables, here referred to as a set of categorical classes $\set{C} = \{\mathrm{c}_1,\mathrm{c}_2, \mathrm{c}_3,\, \cdots\}$, representing the material phases of interest, the task of \emph{pixel classification} is defined as the problem of assigning to each pixel $\tensorComp{p}_{h,w}$ of an image $\tensor{I}$ a class $\mathrm{c}_{h,w} \in \set{C}$. \emph{Semantic segmentation} extends this task further to enforce spatial continuity, i.e., locations in close proximity and with similar $\tensor{F}^{(h,w)}$ are encouraged to have identical labels. As a result, contiguous regions of identically labelled pixels correspond to semantically meaningful structures within the image.  In other words, this means that nearby locations with similar $\tensor{F}^{(h,w)}$ are more likely to belong to the same real-life object.

Therefore, given an image $\tensor{I}$, and a model $\mathcal{U}_{{\set{\theta}}}$ with a set of parameters $\set{\theta}$, the mapping performed by the model is defined as: $\tensor{I} \xmapsto{\mathcal{U}_{\set{\theta}}} \hat{\tensor{Y}}_{\set{\theta}} = (\hat{\tensor{y}}^{\set{\theta}}_{h,w}) \in  \mathbb{R}^{H \times W \times |\set{C}|}$ where $|\set{C}|$ is the number of categorical classes and $\hat{\tensor{y}}^{\set{\theta}}_{h,w} \in \realNum^{|\set{C}|}$ represents the vector of prediction scores for each categorical class with each score indicating how likely it is for a given $\tensorComp{p}_{h,w}$ to belong to one of the $|\set{C}|$ categorical classes depending on the corresponding $\tensor{F}^{(h,w)}$. Generation of correct prediction scores requires an optimal set of parameters $\set{\theta}$, which is obtained by minimizing an objective function during training.

 Semantic segmentation models for images are usually trained using \emph{cross-entropy} loss $\scalFunc{L}_{\text{CE}}$ as the objective function. In the case of CNN-based architectures, for every region $\tensor{F}^{(h,w)}$, cross-entropy loss mandates the existence of a known label $\tensorComp{l}_{h,w} \in \set{C}$. These labels collectively form the tensor $\tensor{L} = (\tensorComp{l}_{h,w}) \in \set{C}^{H \times W}$, known as the labelled image (or annotated image) of $\tensor{I}$.  For practical reasons, $\tensor{L}$ is one hot-encoded as $\tensor{Y}^{01} = (\tensorComp{y}_{h,w,f}) \in \{0,1\}^{H \times W \times |\set{C}|}$ with components 
\begin{equation}
    \tensorComp{y}_{h,w,f} = 
    \begin{cases}
        1 & \text{if } \tensorComp{l}_{h,w} = \mathrm{c}_f \in \set{C}\\
        0 & \text{otherwise}
    \end{cases}
\end{equation}
\noindent  such that $\sum_{f=1}^{|\set{C}|} \mathrm{y}_{h,w,f} = 1$. $\tensor{Y}^{01}$ thus represents the target label for training the segmentation model in a supervised manner. Here, the labelled image imparts semantic supervision to the model outputs and enforces spatial contiguity.

Training is usually performed by iterating over the data multiple times. At any given iteration $t$, the segmentation model $\mathcal{U}$ is parametrized by the set $\set{\theta}_t$ specific to that iteration and the model prediction for an image $\tensor{I}$ is given as $\tensor{I} \xmapsto{\mathcal{U}_{\mathrm{\theta}_t}} \hat{\tensor{Y}}_{t} = (\hat{\tensorComp{y}}^{t}_{h,w,f}) \in \mathbb{R}^{H \times W \times |\set{C}|}$ at that iteration. Therefore, for any $\boldsymbol{\mathrm{I}}$, its model prediction $\hat{\tensor{Y}}_{t}$ and the corresponding one-hot encoded target label $\boldsymbol{\mathrm{Y}}^{01}$, the cross-entropy loss at iteration $t$ is defined as
\begin{equation}
    \scalFunc{L}^{t}_{\text{CE}} = - \sum_{h = 1}^{H}\sum_{w = 1}^{W}\sum_{f = 1}^{|\set{C}|} \log \left(\frac{\exp(\hat{\tensorComp{y}}_{h,w,f}^{t})}{\sum_{f = 1}^{|\set{C}|}\exp(\hat{\tensorComp{y}}_{h,w,f}^{t})}\right){\tensorComp{y}}_{h,w,f} \,.
    \label{eqn:crossentropy_loss}
\end{equation}
In this case, ${\tensorComp{y}}_{h,w,f}$ is fixed for all iterations throughout the training for a given image.

\subsection{Self-Annotation and Unsupervised Semantic Segmentation}
\label{subsec:self_annotation}

For any image $\tensor{I}$ in this case, $\tensor{L}$ does not exist and is expensive to obtain, necessitating an algorithm to generate the required labels during training. This algorithm must take into account the requirement that nearby pixels with similar global and local context are likely to share the same semantic class. As such, the generated labels must consist of contiguous regions with identically labelled pixels such that each such region forms a semantically meaningful structure. 
\begin{figure}[ht!]
\centering
\includegraphics[trim = 0 165 0 0, width=0.8\textwidth]{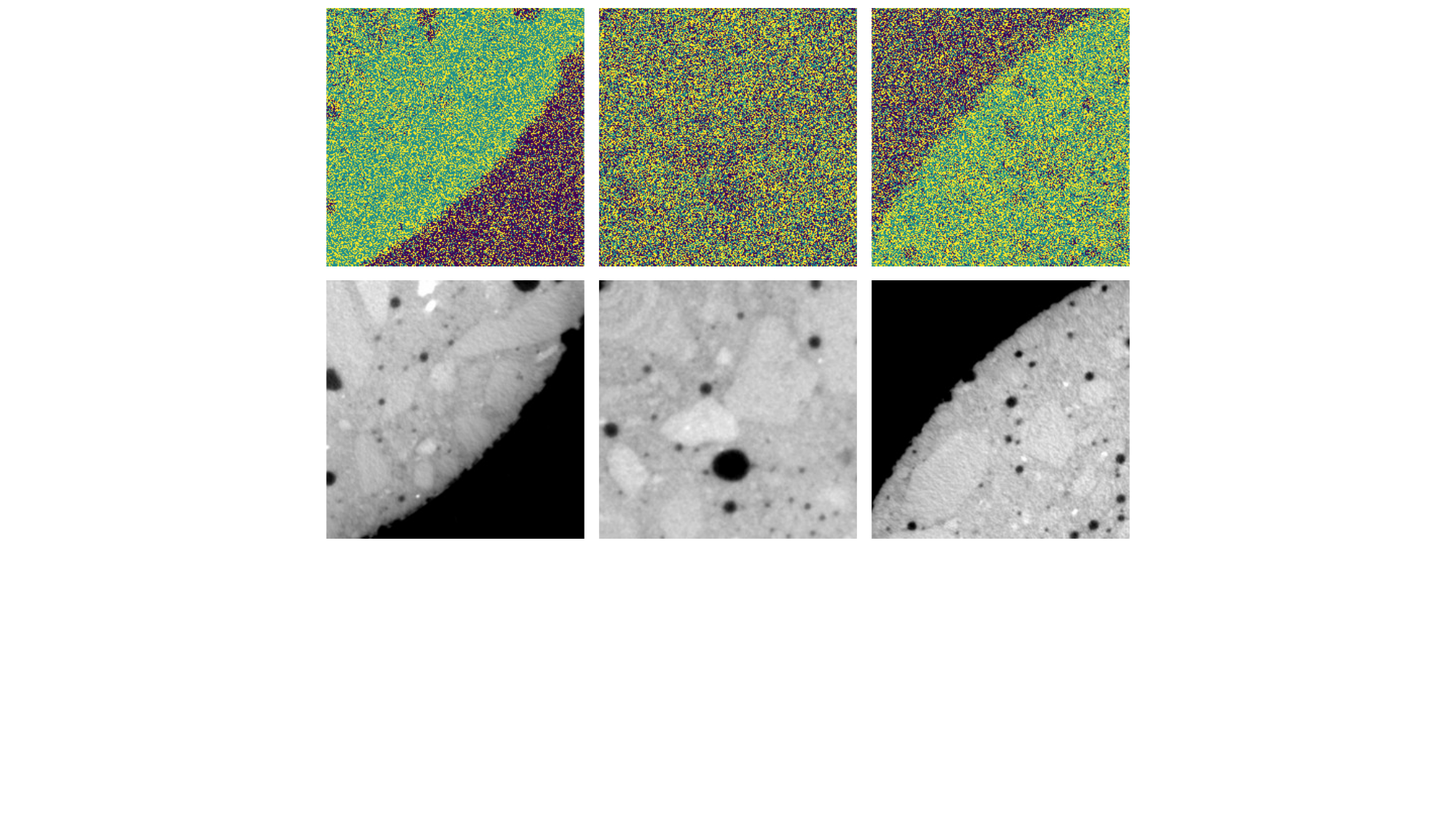}
\caption{Example of $\arg\max$ application on the prediction values of three randomly selected training samples in the first training epoch. Each colour is representative of the categorical class with the  maximum prediction value.}
\label{fig:similar_pix_dissimilar_label}
\end{figure}
A key property of CNN-based models is the fact that they are able to partition a given image based on complex information, such as spatial arrangement of objects contained in the receptive field corresponding to each pixel of the output. And therefore, it is possible for the model to identify individual pixels in close proximity, which are otherwise similar in intensity, as different semantic class due to difference in the information contained within the respective receptive fields and vice versa. This is particularly evident in the initial phases of training, as indicated in \autoref{fig:similar_pix_dissimilar_label}, where, despite the use of identical model parameters for a given image, identically coloured pixels in immediate proximity within each image do not generate correspondingly high score for the identical semantic class relative to the other semantic classes. As such, the parameter set ${\set{\theta}}$ of the model if simultaneously trained to associate pixels in close proximity and with similar intensities with identical semantic class, may develop a global-local relationship and learn to associate nearby pixels with similar intensity and global context with the identical semantic class.

This necessitates identification of localized regions in an image that, on the account of perceptual similarity, may belong to the same semantic class. To this end, there exists a class of heuristics-based algorithms that are able to identify perceptually similar regions in an image and generate clusters known as \emph{superpixels}. Pixels belonging to these clusters also additionally satisfy the requirement of spatial proximity. Therefore, in this context, we define a set of $I$ superpixels of an image $\tensor{I}$ as $\{\set{P}_i\}_{i=1}^I$ where each superpixel $\set{P}_i = \{\tensorComp{p}_{h,w}\}_i$ represents a spatially contiguous, non-empty and non-overlapping region of an image such that $\cup_{i = 1}^I \set{P}_i = \tensor{I}$, with any given superpixel occupying a spatial region $\Omega_{\set{P}_i} = \{(h,w)\}_i\,$ of an input image. Consequently, the set $\{\Omega_{\set{P}_i}\}_{i=1}^I$ of spatial regions that can be used to associate each pixel in an image to a unique superpixel is here on termed as \emph{superpixel-map} of an image.

\begin{figure}[h!]
\centering
\includegraphics[trim = 0 165 0 0, width=0.8\textwidth]{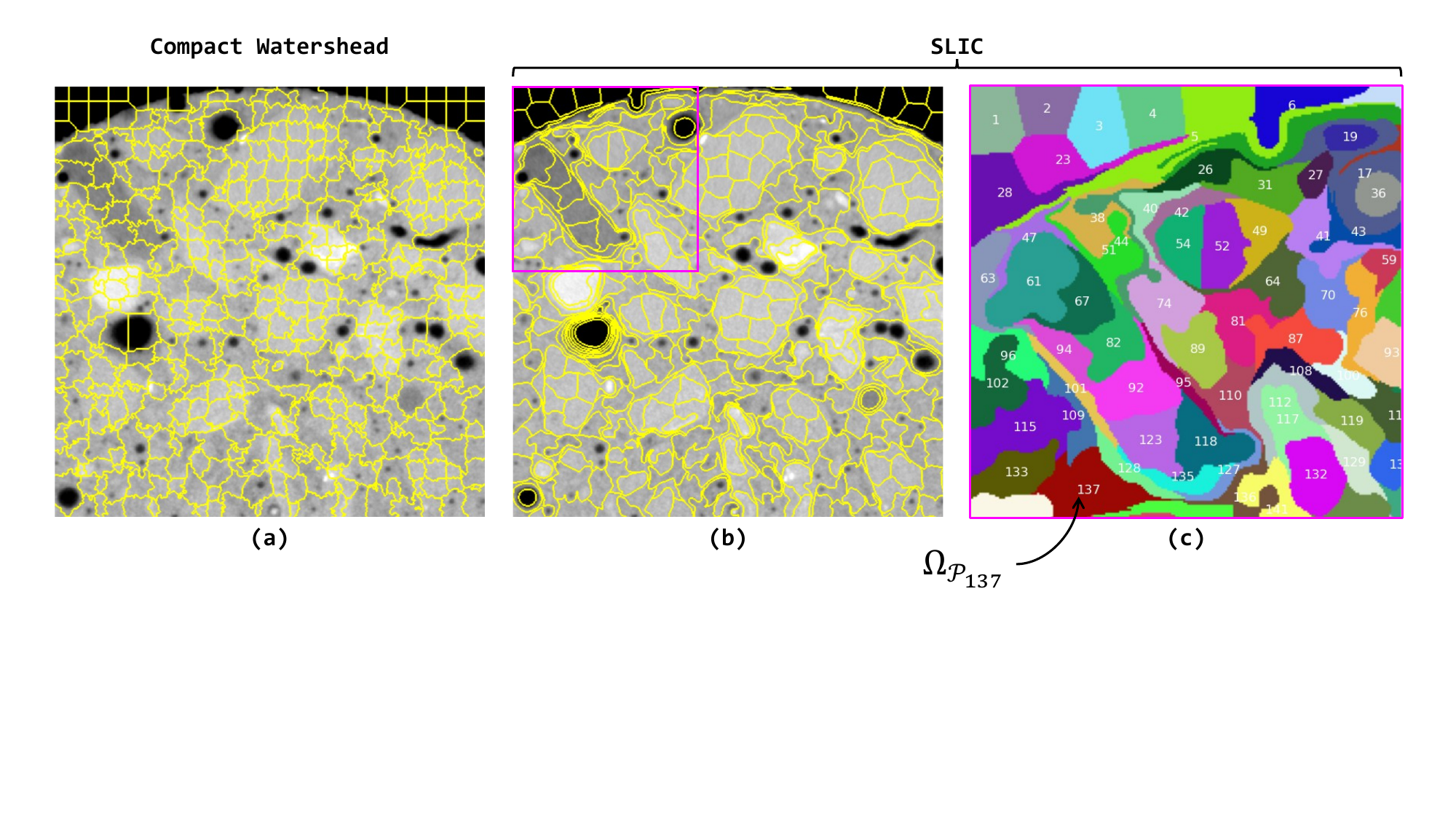}
\caption{Comparison of superpixels obtained using \emph{Scikit Image} \cite{van2014scikit} implementation of the compact watershed method \cite{neubert2014compact} (a) and the SLIC algorithm \cite{achanta2010slic} (b). These images are produced by overlaying the superpixel boundaries on the original greyscale image. Image (a) is composed of approximately 480 superpixels, while image (b) is composed of approximately 370 superpixels. In (c) the superpixel map is illustrated for the region inside the magenta box in (b). Each $\Omega_{\set{P}_i}$ indicates to the location of a given superpixel.}
\label{fig:superpixel_comparison}

\includegraphics[trim = 0 250 0 0, width=0.6\textwidth]{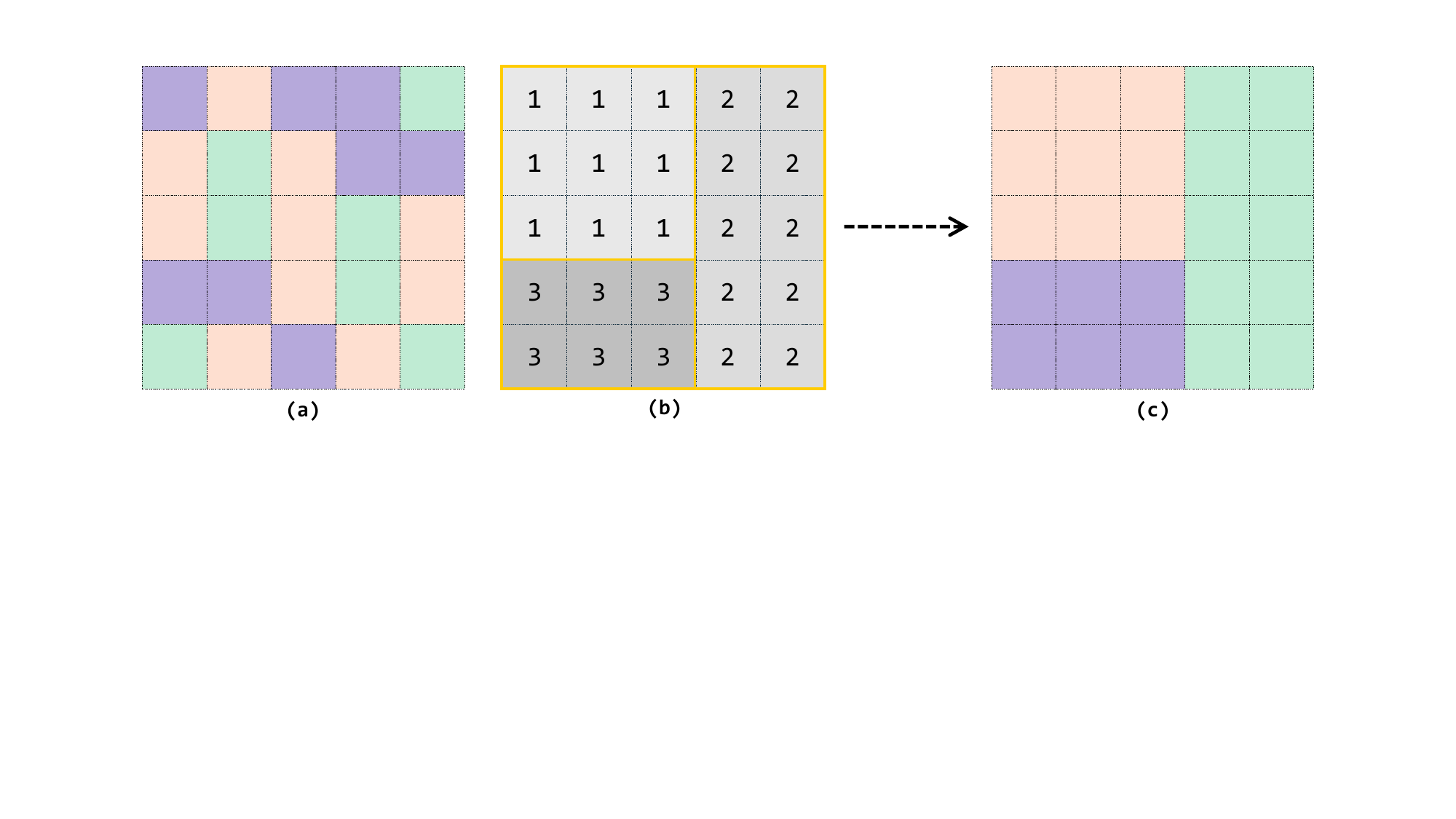}
\caption{A schematic depiction of the superpixel refinement operation. Given an image with categorical pixels as in (a), depicted using colours and given a superpixel map as depicted (b) with number indicating the superpixel index, superpixel refinement function assigns to the region occupied by each superpixel the most frequently occurring categorical class in (a), which results in the output depicted in (c).}
\label{fig:superpixel_refinement scheme}
\end{figure}

Since superpixel algorithms cluster together regions of perceptual similarity (see \autoref{fig:superpixel_comparison}), the generated superpixels are also able to demarcate boundaries between perceptually distinct regions and therefore, in this application, also enable identification of interface between the constituent phases. This is particularly evident when using the algorithm termed as SLIC \cite{achanta2010slic}, which is readily available as part of the image processing library \emph{Scikit-Image} \cite{van2014scikit}. Therefore, for generation of superpixels in this work, SLIC was chosen, as it outperformed other algorithms including the more recent watershed segmentation technique proposed in \cite{neubert2014compact}. Nevertheless, it must be noted that superpixels are limited to local perceptual similarity and do not identify objects or assign semantic meaning to the regions and therefore do not comprise any information beyond their boundaries.

Hence, in order to obtain automated labels, an assumption is made that given a superpixel $\set{P}_i$ at locations indicated by $\Omega_{\set{P}_i}$ the most frequently occurring class label for the pixels at these locations predicted using the model $\mathcal{U}_{\set{\theta}_t}$ is the most likely semantic category of all the pixels within that superpixel. And correspondingly, a label is generated by assigning to all locations in $\Omega_{\set{P}_i}$ this class label, see \autoref{fig:superpixel_refinement scheme}.


\begin{figure}[ht!]
\centering
\includegraphics[trim = 0 0 0 0, width=0.75\textwidth]{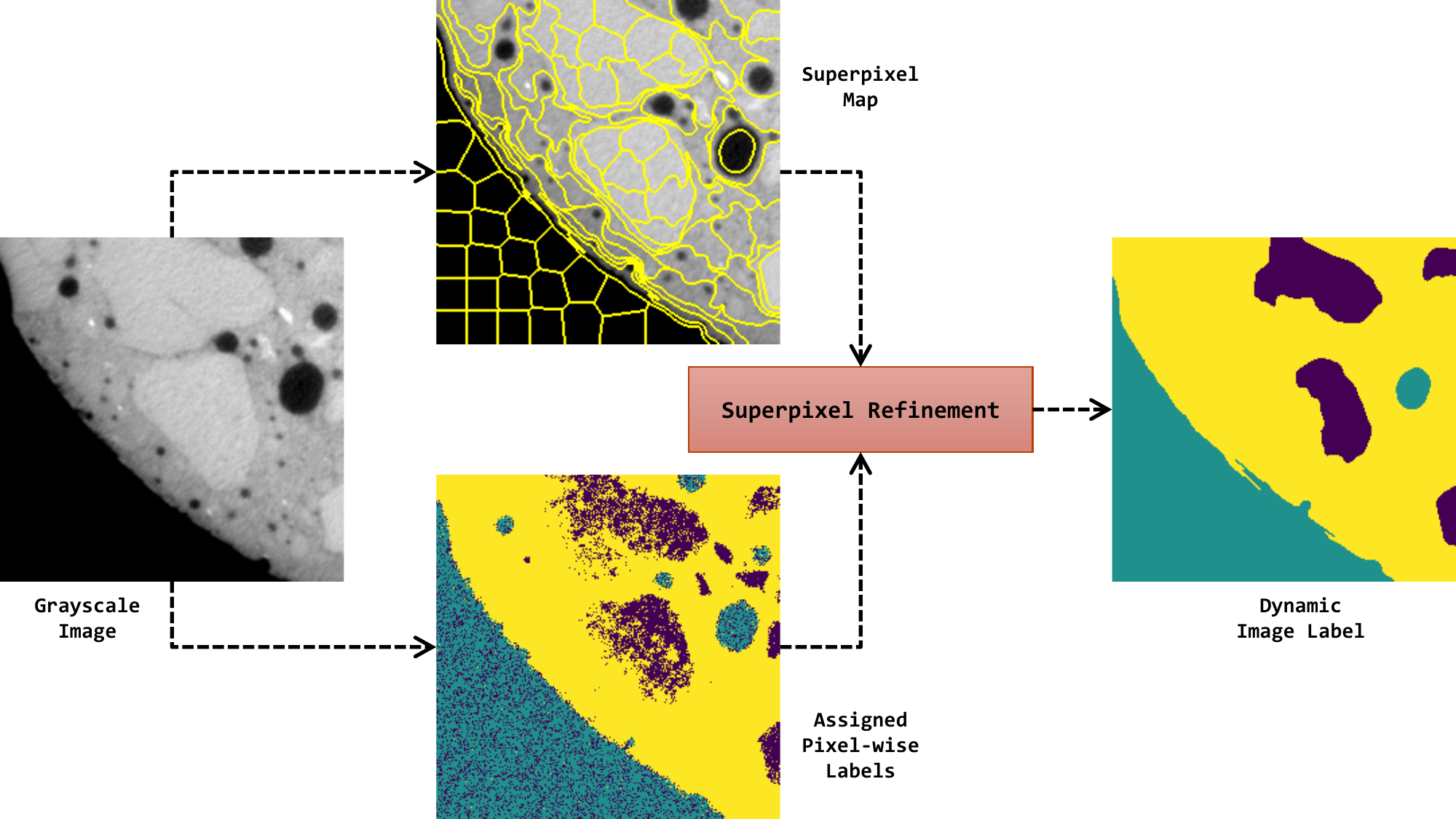}
\caption{Superpixel refinement as part of the overall scheme. Given an image, the superpixel map is computed and for each training iteration, the assigned pixel-wise labels are refined to produce dynamic image labels. }
\label{fig:self_annotation}

\includegraphics[trim = 0 150 0 0, width=0.75\textwidth]{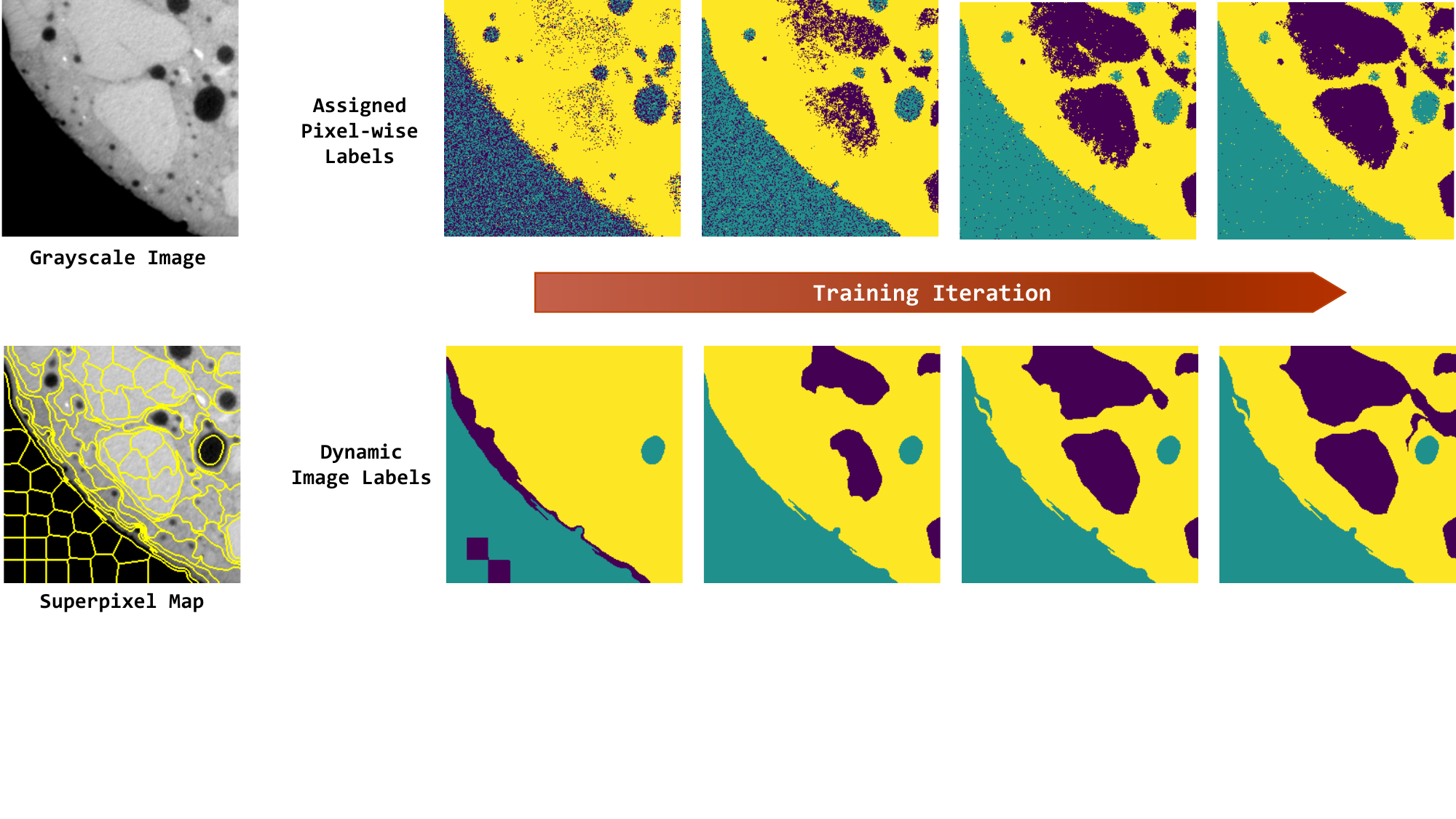}
\caption{Illustration of evolution of dynamic image labels with respect to training iterations (Note: Number of images do not equate to the number of training iterations).}
\label{fig:self_annotation_evolution}
\end{figure}


Nevertheless, components of $\boldsymbol{\mathrm{\hat{Y}}}_{t}$ have no bounds over their relative magnitudes across the feature dimensions, which is otherwise enforced using (\ref{eqn:crossentropy_loss}) in a supervised training scenario. As such, under the aforementioned assumptions, it is possible for the model to collapse to a trivial solution where one of the channels has systematically higher or lower values compared to the others. This can result in solutions where the segmentation consists of only one phase. To counter this, the components of $\hat{\tensor{Y}}_{t}$ are first normalized per feature dimension  to obtain $\bar{\tensor{Y}}_{t} = (\bar{\mathrm{y}}^{t}_{h,w,f}) \in \realNum^{H\times W \times |\set{C}|}$ with
\begin{equation}
    \bar{\mathrm{y}}^{t}_{h,w,f} = \frac{\hat{\mathrm{y}}^{t}_{h,w,f} - \hat{\mu}^{t}_{f}}{\hat{\sigma}^{t}_{f}} 
    \label{eqn:channel_standardization}
\end{equation}
where, $\hat{\mu}^{t}_{f}$ and $\hat{\sigma}^{t}_{f}$ denote the mean and variance respectively, of elements of $\boldsymbol{\mathrm{\hat{Y}}}_{t}$ for the $f$-th feature dimension, computed as 
\begin{align*}
    \hat{\mu}^{t}_{f} &= \frac{1}{HW}\sum_{h=1}^H \sum_{w=1}^W \hat{\mathrm{y}}^{t}_{h,w,f}\, , \\ 
    (\hat{\sigma}^{t}_{f})^2 &= \frac{1}{HW}\sum_{h=1}^H \sum_{w=1}^W (\hat{\mathrm{y}}^{t}_{h,w,f} - \hat{\mu}^{t}_{f})^2 \,.
\end{align*}
\noindent $\bar{\tensor{Y}}_{t}$ has components in each channel with zero mean and unit standard deviation, thereby, allowing each feature dimension to have similar chance of being selected using $\arg\max$ classification. 

However, the normalization in (\ref{eqn:channel_standardization}) also results in degradation of the output in certain circumstances. This is particularly evident in cases of class imbalance in images, i.e. instances where particular classes are absent or are not sufficiently represented. This adjustment then artificially translates and rescales the values in the channel corresponding to those classes in order to raise their chances of getting selected even though those classes are absent. Nevertheless, in the absence of this adjustment, the model collapses to produce less than the required number of phases. We note that this is an aspect that requires further studies.

Following this, for each location $(h,\,w)$ the task of assigning a unique class label to each pixel is undertaken using $\arg\max$ label assignment operation such that $\bar{\tensor{Y}}_{t} \xmapsto{\arg\max} \hat{\tensor{S}}_{t}$, where $\hat{\tensor{S}}_{t} = (\hat{\tensorComp{s}}^{t}_{h,w}) \in \set{C}^{H \times W}$ is the tensor of assigned pixel-wise labels such that
\begin{equation}
    \hat{\tensorComp{s}}^{t}_{h,w} = \argmax_{f\, \in \{ 1, \cdots, |\set{C}|\}} (\bar{\mathrm{y}}^{t}_{h,w,f})\,.
    \label{eqn:argmax}
\end{equation}

Finally, the constraint on spatial contiguity is enforced by what is referred to as \emph{superpixel refinement} that automates the annotation process by converting the tensor of assigned pixel-wise labels to regional annotations. For this purpose, the tensor $\tilde{\tensor{L}}_{t} = (\tilde{\tensorComp{l}}^{t}_{h,w}) \in \set{C}^{H\times W}$ is introduced, which is referred to as the \emph{dynamic image label} and evolves with every training iteration $t$. Given a spatial region $\Omega_{\set{P}_i}$ of an image occupied by superpixel $\set{P}_i$, the most frequently occurring label in the corresponding spatial region in $\hat{\tensor{S}}_{t}$ is given as 

$\tilde{\tensorComp{l}}^{t}_{h,w} = \mathrm{mode}(\{\hat{\tensorComp{s}}^{t}_{h,w}\})\, \forall \, (h,\,w) \in \Omega_{\set{P}_i}$. The dynamic image label $\boldsymbol{\mathrm{\tilde{L}}}_{t}$ is updated after every training iteration $t$. This process is schematically illustrated in \autoref{fig:self_annotation} as a part of the training workflow  and its evolution with training progression is presented in \autoref{fig:self_annotation_evolution}.

Therefore, at any iteration $t$, given the mapping $\tensor{I} \xmapsto{\mathcal{U}_{\set{\theta}_t}} \hat{\tensor{Y}}_{t}$,  its normalization $\bar{\tensor{Y}}_{t}  = (\bar{\tensorComp{y}}^{t}_{h,w,f}) \in \realNum^{H\times W\times |\set{C}|}$ and a dynamic label $\tilde{\tensor{L}}_{t}$ with its one-hot encoding $\tilde{\tensor{Y}}^{\mathrm{01}}_{t} = (\tilde{\tensorComp{y}}^{t}_{h,w,f}) \in \{0,\,1\}$, the cross-entropy in (\ref{eqn:crossentropy_loss}) is rewritten as 
\begin{equation}
    \tilde{\scalFunc{L}}^{t}_{\mathrm{CE}} = - \sum_{h = 1}^{H}\sum_{w = 1}^{W}\sum_{f = 1}^{|\set{C}|} \log \left(\frac{\exp(\bar{\tensorComp{y}}_{h,w,f}^{t})}{\sum_{f = 1}^{|\set{C}|}\exp(\bar{\tensorComp{y}}_{h,w,f}^{t})}\right){\tilde{\tensorComp{y}}}_{h,w,f}^{t}
    \label{eqn:crossentropy_loss_sp}
\end{equation}
\noindent where, ${\tilde{\tensorComp{y}}}_{h,w,f}^{t}$ is iteration dependent. In fact for any $\tensor{F}^{(h,\,w)}$, $\,{\tilde{\tensorComp{y}}}_{h,w,f}^{t}$ is not unique but may take any of the values in $\set{C}$ during training (see \autoref{fig:self_annotation_evolution}). Training a model in such a situation is termed as unsupervised segmentation and the model is expected to gradually converge to produce the correct label.

It is important to note that in practice, training over the entire dataset is performed by traversing over it in small non-overlapping subsets called \emph{mini-batches}, comprising multiple images, and the model weights are updated for every such subset, a process which is also termed as mini-batch learning. In this setting, the loss used for gradient computation is a cumulative metric, typically the sum or average over the individual losses computed for all images in a mini-batch. For more information on the reasoning behind the use of mini-batch for training, one may refer to \cite{bottou2012stochastic, bengio2012practical, keskar2016large, smith2020generalization, bishop2023deep}. Hence, in this work, the model parameters are updated using the gradient computed on the average loss over a set of images in a mini-batch $\mathrm{Avg}(\tilde{\scalFunc{L}}^{t}_{\mathrm{CE}})_{\mathrm{MB}}$ with respect to $\set{\theta}_t$
\begin{equation}
    \set{\theta}_{t+1} \leftarrow \set{\theta}_t - \eta \frac{\partial \mathrm{Avg}(\tilde{\scalFunc{L}}^{t}_{\mathrm{CE}})_{\mathrm{MB}} }{\partial{\set{\theta}_t}}\,.
    \label{eqn:parameter_update}
\end{equation} 


\section{Implementation}
\label{sec:Implementation}

\subsection{Training, Validation and Test Datasets}
\label{subsec:training_data}

\begin{figure}[h!]
\centering
\includegraphics[trim= 0 50 0 25, width = 0.80\textwidth]{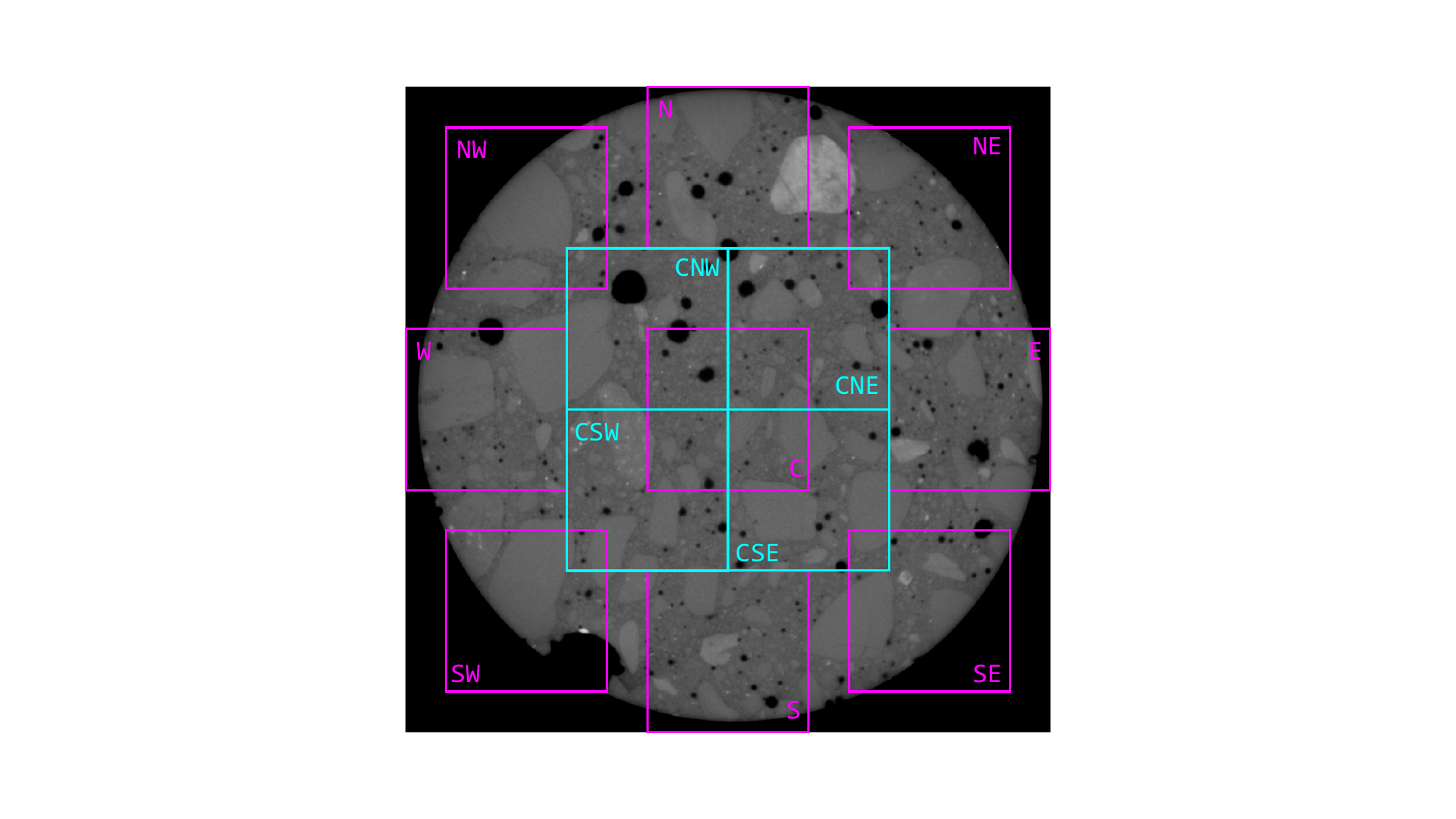}
\caption{Tiling strategy used for generating training data. Each image yielded $13$ tiles. The squares represent tile placement. Abbreviations - N: North, S: South, E: East, W: West, C: Center. Colours for representation purposes only. The squares at the four corners were placed $128$ pixels away from the edges in each direction.}
\label{fig:tiling_strategy}
\end{figure}

Due to limitations on GPU memory, training was performed using tiles measuring $256\,\times \,256$ pixels that were cropped out of images measuring $1024\,\times \,1024$ pixels using the tiling strategy presented in \autoref{fig:tiling_strategy}. The alternative approach of rescaling the images to a smaller dimension instead of tiling was not used since \emph{Interfacial Transition Zone} (ITZ) are of particular importance in concretes, and doing so would have resulted in loss of detail. 

For training, a subset comprising approximately $9500\,(= D_{train})$ randomly chosen tiles obtained from approximately $1470$ images belonging to $6$ distinct XCT samples were used, representing approximately $41\%$ of the available data in terms of pixels. A separate subset of approximately $4730\, (= D_{val})$ identically sized tiles was used for formulating a stopping criterion for the training process, and is hence termed as the validation dataset. These were also randomly chosen and it was ensured that there was no overlap between these two datasets.

Subsequently for each tile in both the datasets, a superpixel map was computed which remained fixed through all iterations. As already stated in the previous section, superpixels were generated using the algorithm termed as SLIC (Simple Linear Iterative Clustering) \cite{achanta2010slic}. This algorithm when used on greyscale images is parametrized by three degrees of freedom, i.e, two for spatial proximity and one for pixel intensity and  primarily requires two user-defined inputs, namely, the desired number of superpixels and the compactness parameter. The number of superpixels in an image also determines their size, while compactness controls, in simple terms, how square the ensuing superpixels are. In our case, the number of superpixels was chosen based on the fact that, for concretes, aggregates smaller than $4\text{ mm}$ are typically considered part of the mortar \cite{ren2023methods}. Therefore, this parameter was set to ensure that the superpixels were approximately $4\text{mm}$ equivalent in dimension. The compactness value was chosen through experimental trials to ensure that the superpixel boundaries adhered well to the phase boundaries in most cases and were not too sensitive to image noise, which otherwise resulted in jagged boundaries.

For evaluating the performance of the model on unseen data, a separate XCT-sample, distinct from the 6 XCT samples stated above, was used. This is the test dataset in this case and comprises approximately $590$ slices measuring $1024 \times 1024\, \mathrm{pixels}$. This sample was preprocessed identically to the rest of the samples and no part of it was previously used for training or validation. This sample is here on referred to as $\tensor{T}_{\mathrm{PR}} = \{ \tensor{I}_n^{\mathrm{PR}} \}$. In the subsequent text the index $n$, corresponding to the slice index, has been omitted for visual clarity.

\subsection{Training Procedure}
\label{subsec:training}


\begin{figure}[htb!]
\centering
\begin{overpic}[trim= 0 10 0 0, width = 1.0\textwidth]{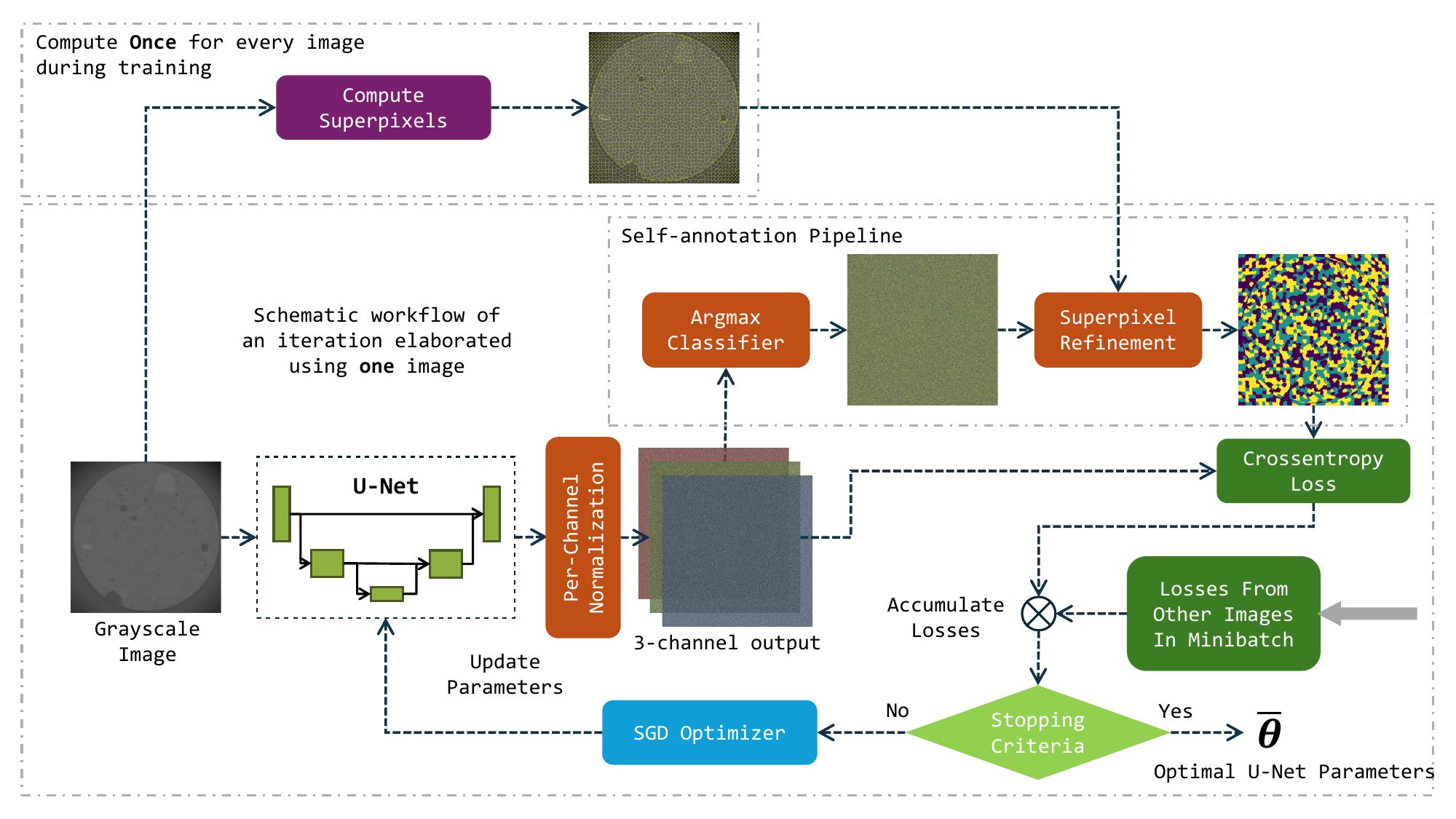}
\put(15.5,11){\footnotesize $\tensor{I}$}
\put(56.5,11){\footnotesize $\bar{\tensor{Y}}_t$}
\put(68.5,25.3){\footnotesize $\hat{\tensor{S}}_t$}
\put(51,42){\footnotesize $\{\set{P}_i\}$}
\put(93,25.3){\footnotesize $\tilde{\tensor{L}}_t / \tilde{\tensor{Y}}^{01}_t$}
\end{overpic}

\caption{Schematic flowchart for unsupervised training of the U-Net model aided by dynamic label generation assuming a 3-channel output for one image. Superpixels for each image is computed once. Training is performed by iterating over the images several times. The output produced by the model is used to generate dynamic image labels. Each output channel corresponds to each material phase of interest, i.e. aggregate, mortar and concrete.}
\label{fig:Training-flowchart}
\end{figure}
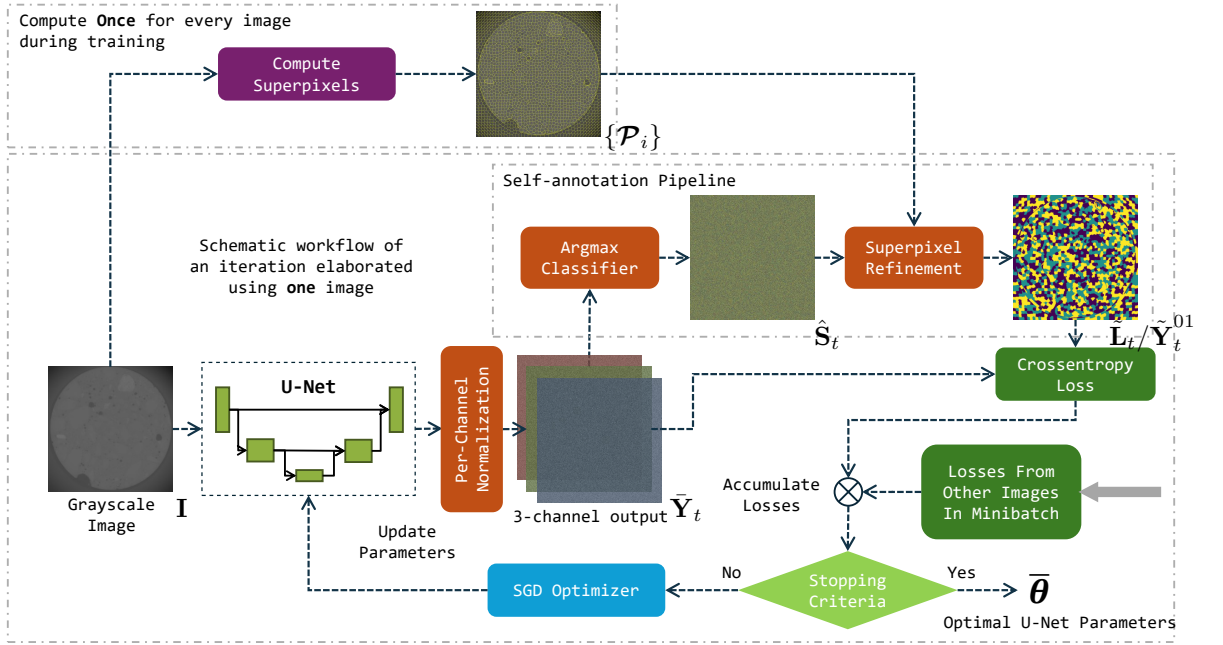


For training, the scheme presented in \autoref{fig:Training-flowchart} was followed. At each training iteration, for every input image (here tile), the model output was normalized per-channel and passed through the \emph{self-annotation pipeline} to produce a dynamic image label $\tilde{\tensor{L}}_{t}$. The cross-entropy loss was then computed between the normalized model output $\bar{\tensor{Y}}_{t}$ and one-hot encoded dynamic image label $\tilde{\tensor{Y}}^{\mathrm{01}}_{t}$ using (\ref{eqn:crossentropy_loss_sp}). As already mentioned, since training was performed using mini-batches, the model parameters were updated based on the average loss per mini-batch $\mathrm{Avg}(\tilde{\scalFunc{L}}^{t}_{\mathrm{CE}})_{\mathrm{MB}}$ using (\ref{eqn:parameter_update}). Here, we would also like to clarify that training involves multiple passes over the entire dataset, wherein, during each pass, the model parameters are updated multiple times, i.e. once per mini-batch. Therefore, a single pass over the dataset is referred to as an \emph{epoch} $(e)$, an update cycle per mini-batch is referred to as an \emph{iteration} $(t)$. Since, here, training dataset size was $D_{train} = 9500$ tiles and mini-batch size was $N=24$ tiles, each epoch comprised $t_e = 396$ iterations where, $t_e = \left\lfloor \frac{D_{train}}{N} \right\rfloor + \iota$ with $\iota = 1$ if $(D_{train} \bmod N) \neq 0$ else $\iota = 0$.

Although the choice of using smaller tiles instead of full slices was guided by limitations on GPU memory, it was also noted that using smaller mini-batches composed of entire images instead of using larger mini-batches composed of tiles resulted in severe under-segmentation. Here, we define \emph{under-segmentation} as a situation where the model fails to recognize all the phases present in an image, which results in segmentations with less than the required number of phases. In our case, it was primarily restricted to the aggregates and mortars being classified as the same phase. 

The set of model parameters at $t=0$ i.e. $\set{\theta}_0$ was initialized following the Xavier-normal (also known as Glorot-normal) initialization method \cite{glorot2010understanding} for the kernels while the bias terms were initialized to zero. This was optimized using stochastic gradient descent (SGD) with an additional momentum parameter \cite{sutskever2013importance}, where the gradient at iterations $t = 2,\, 3,\, \cdots$ are updated by adding a weighted contribution of the gradient from iteration $t-1$, except at iteration $t = 1$. This weighting factor is referred to as the momentum. An initial learning rate of $\eta = 0.0005 $ with optimizer momentum $0.5$ was used. The learning rate was adjusted during training through the use of \emph{cosine annealing} with restarts \cite{loshchilov2016sgdr}. The first restart interval was set to $10$ epochs and doubled for each subsequent restart. The minimum learning rate was set to $\eta_{min} = 0.0001$. 

Training performance was tracked at every epoch $e$ by observing the evolution of mean $\tilde{\scalFunc{L}}^{e}_{\mathrm{CE}}$ over the entire validation dataset using the model parameters $\set{\theta}_{e}$ obtained at the end of the stated epoch. During validation, the model is run in prediction mode and therefore the dropout layers are deactivated and consequently, the mean validation losses are consistently lower than training losses. Trainings were run for $e_{max}=100$ epochs, and following the usual practice, the set of model parameters for final prediction was chosen at the end of the epoch with minimum mean loss (at $e < e_{max}$ in all cases) on this dataset, which was the \emph{stopping criterion}, yielding  $\bar{\set{\theta}} = \set{\theta}_{e}$. This was done to ensure that the model did not over-fit to the training data, where, increase in validation loss can be imagined as the tensor of assigned pixel-wise labels $\hat{\tensor{S}}_t$ getting noisier as the model starts over-fitting with further training.

\subsection{Prediction Procedure}
\label{subsec:prediction}

Due to the model $\mathcal{U}$ being trained on tiles of size $256 \times 256\, \mathrm{pixels}$ which were cropped out of the images, the model performed optimally when the prediction was obtained using similarly sized tiles cropped out of each $\tensor{I}^{PR}$. Such a behaviour, where semantic segmentation models perform distinctly based on the size of the input image is well established \cite{zhu2023adaptive}. However, stitching them together to reconstruct the full-size images also resulted in undesirable artifacts and discontinuities in prediction values at tile edges, which necessitated certain corrective measures such as overlapping at the edges during prediction. Nevertheless, it is presumed that training on large mini-batches of full-sized images can effectively avoid such artifacts.

The entire framework was implemented using \emph{Pytorch} \cite{paszke2019pytorch} deep learning library. On a workstation equipped with AMD Ryzen-9 7950X with 128GB system memory and one Nvidia RTX4090 GPU with 24GB VRAM, training took approximately $3\, \text{-} \,4$ min per epoch (i.e approx. 5-7h for the entire training process) and prediction for $\tensor{T}_{\mathrm{PR}}$ required approximately $1 \text{-} 2 \, \mathrm{min}$. However it is to be noted that these times include certain overheads that were suboptimally implemented and performance may improve with an optimized code.

\section{Experiments}
\label{sec:experiments}

Numerical experiments were first performed in a fully unsupervised manner where the model was trained by constraining it to produce three channels at the output corresponding to three distinct phases (US$3$). In this case it was found that the model consistently failed to unambiguously detect the porous phase in the images. Hence, the unsupervised training was repeated by relaxing the constraint on the output channels to four instead (US$4$). The two unsupervised cases are presented in Sections \ref{subsec:unsup_3ch} and \ref{subsec:unsup_4ch} respectively.

\begin{figure}[hbt!]
\centering
\includegraphics[trim= 0 190 0 0, width = 1.0\textwidth]{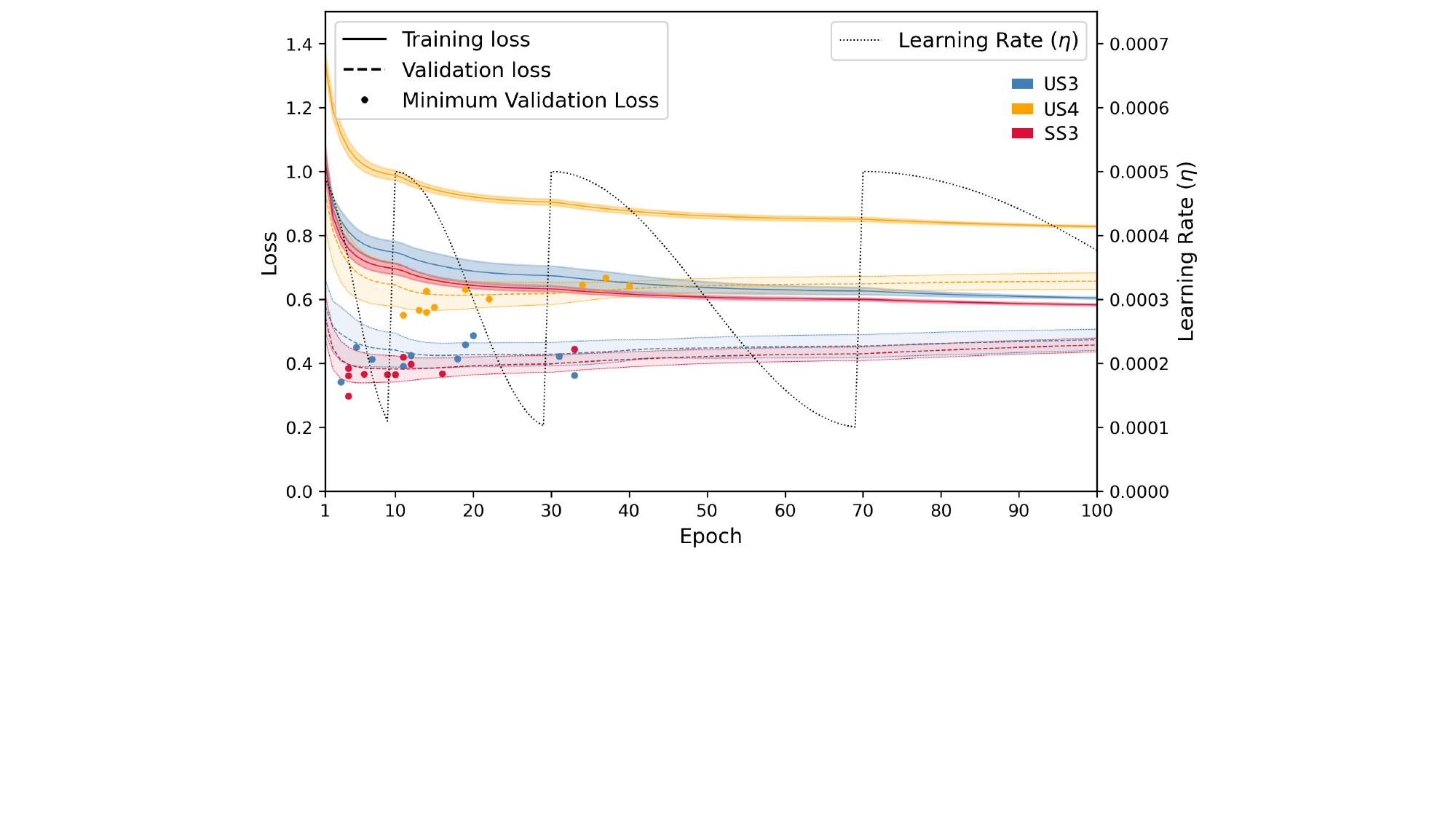}
\caption{Mean training losses and mean validation losses averaged over $10$-training runs for each case represented using solid and dashed line respectively. The standard deviation over the $10$-runs is represented using the coloured region around the mean losses. Dotted black line indicates learning rate. The minimum validation loss for each experiment is marked using a dot.}
\label{fig:all_loss}
\end{figure}

Since the porous phase was easily discernible and therefore could be segmented using simple thresholding, we further tested a semi-supervised training method. In which case, the model was constrained to produce three output channels (SS$3$). In this case, self-annotation was only performed for the aggregates and the mortar, while labels of the porous regions were pre-specified using labels obtained by thresholding. This case is presented in Section \ref{subsec:semi_sup_3ch}.

For each case, the $10$ distinct models were trained using identical parameters, the average of the mean training and validation losses for each is presented in \autoref{fig:all_loss}. It was observed that although the evolution of losses followed similar pattern, there was considerable discrepancy in the epoch at which the minimum validation loss was observed. Nevertheless, the prediction outcomes were qualitatively similar and replicable among distinct training runs in all the cases. Since in practice, it can only be expected to run the training once, in the subsequent sections, we therefore only present an example of results that can be expected without any reference to a specific training run or to the epoch at which minimum mean loss on the validation dataset was observed.

For this, some changes in notation is introduced, whereby, while referring exclusively to the results in each of the cases, we omit any reference to epoch $e$ and iteration $t$ for reasons stated above and only refer to the cases using a subscript in each symbol. Otherwise, the symbols retain identical meaning as in Section \ref{sec:unsupervised_segmentation}.

\FloatBarrier
\subsection{Unsupervised Training: Given 3-phases (US3)}
\label{subsec:unsup_3ch}

\begin{figure}[h!]
\centering
\includegraphics[trim = 0 100 0 0, width=0.8\textwidth]{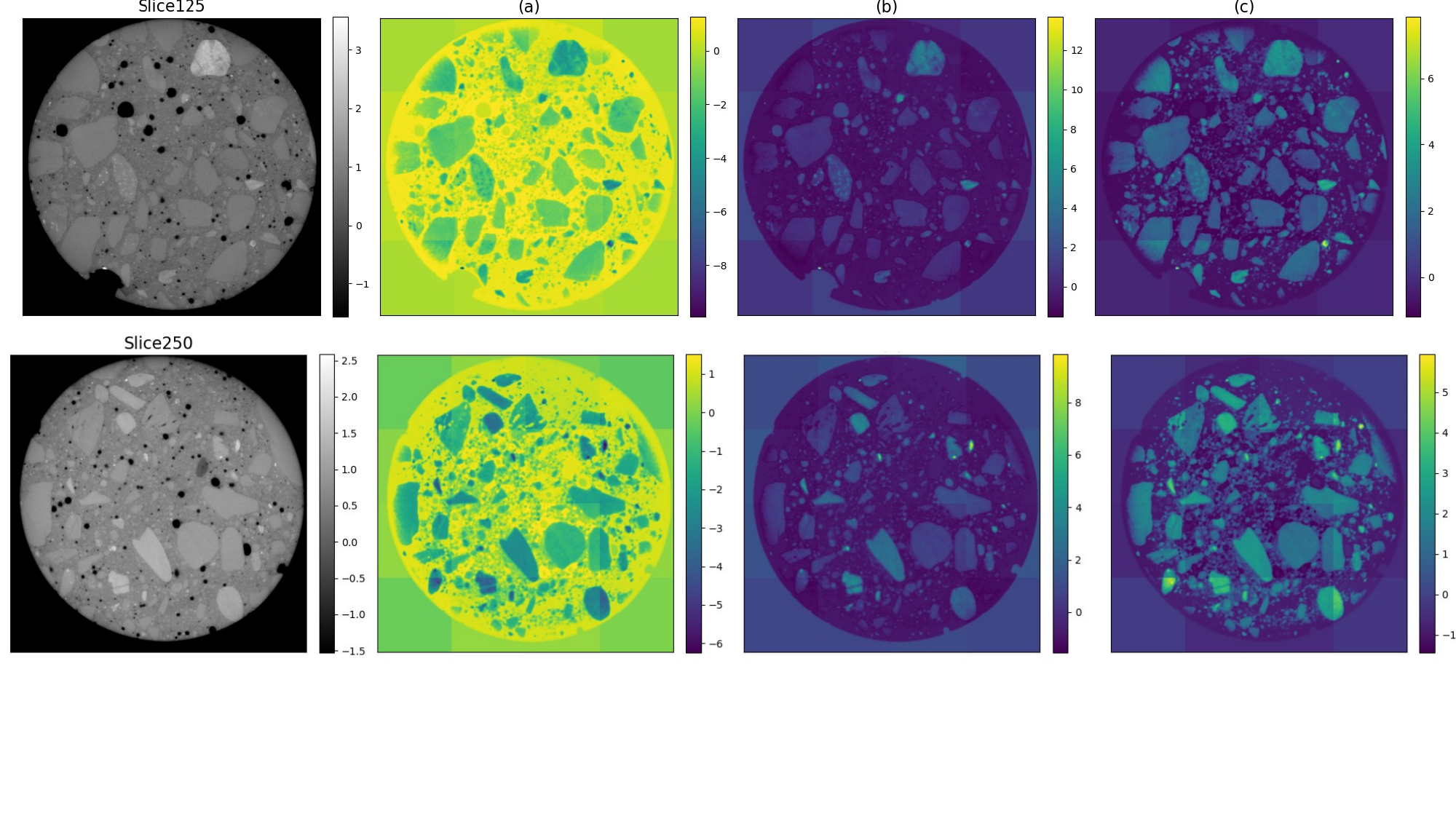}
\caption{$\mathrm{US3}$: Example of $3$-channel prediction results using model trained in an unsupervised manner for the given input slices (left). Column (a) and (c) can be associated with the mortar and aggregate phases respectively. Channel (b) is ambiguous and displays similar scores for porosity and aggregates.}
\label{fig:3_channel_output}
\end{figure}

\begin{figure}[ht!]
\centering
\vspace*{0.25cm}
\begin{overpic}[trim = 0 200 0 0, clip, width=0.8\textwidth]{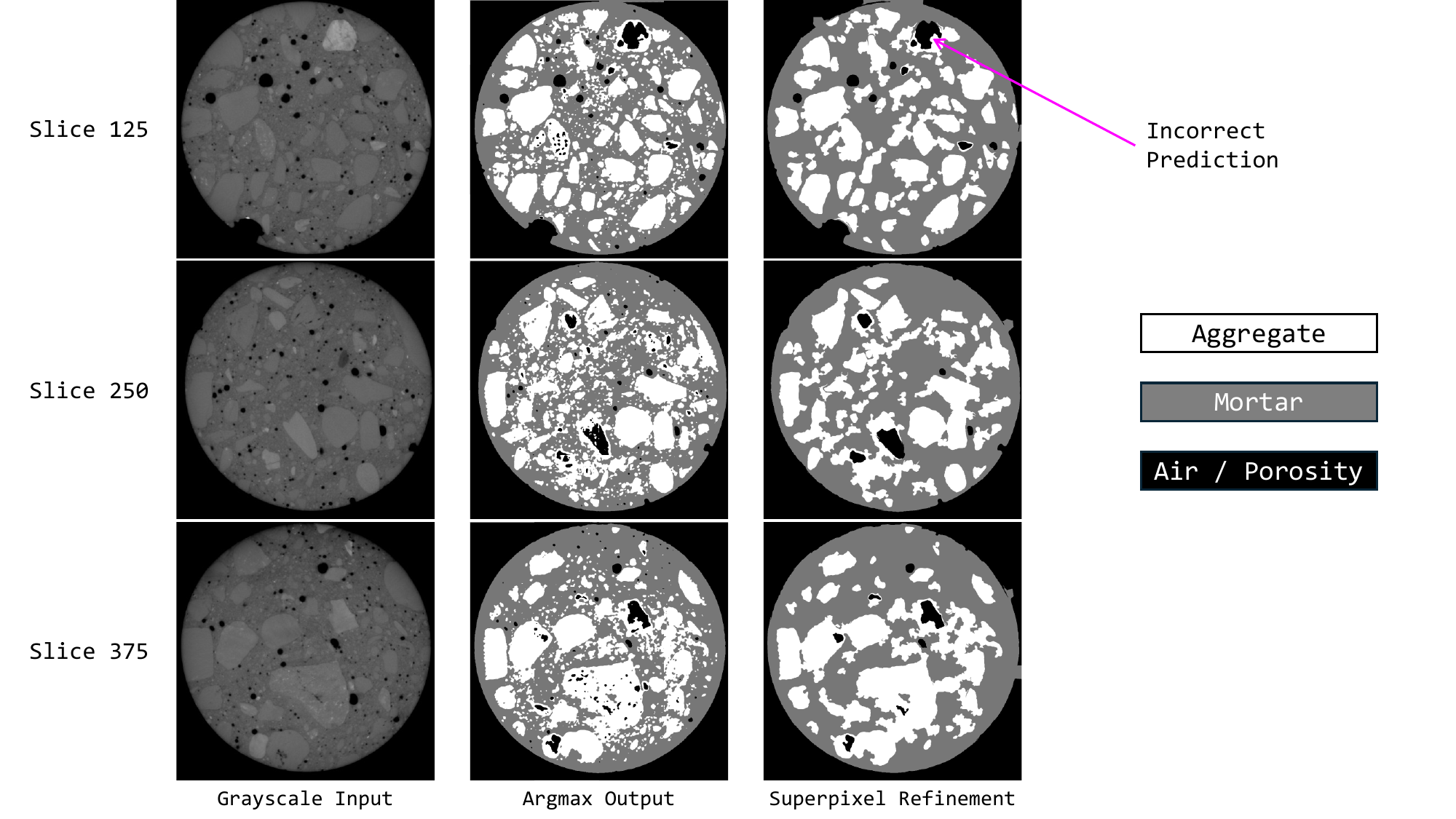}
\put(19,36){\tiny Input}
\put(32.3,36){\tiny $\arg\max$ classification}
\put(52.7,36){\tiny superpixel refinement}
\end{overpic}
\caption{$\mathrm{US3}$: $\arg\max$ classification of $\bar{\tensor{Y}}^{\mathrm{PR}}_\mathrm{US3}$ and the superpixel refinement $\tilde{\tensor{L}}^\mathrm{PR}_\mathrm{US3}$ using parameters equivalent to that used during training.}
\label{fig:argmax_3ch_unsup}
\end{figure}

The model was first constrained to produce output comprising $3$-feature channels, one for each material phase under consideration, and the self-annotation procedure was expected to generate labels for each of the $3$ material phases. In the normalized prediction result $\bar{\tensor{Y}}^{\mathrm{PR}}_\mathrm{US3}$ comprising $3$ feature channels, examples of which for two slices are presented in \autoref{fig:3_channel_output}, it was observed that the model is able to assign one channel each to the aggregates and the mortar phase. However, it did not assign the third channel exclusively to the porous phase and instead, generated similar prediction values for both the aggregates and mortar phases, leading to ambiguity in its semantic meaning. This is particularly pronounced in case of bright aggregates, for which the model generated especially high values in this channel. This resulted in incorrect classification of certain aggregates in the pixel-wise assigned labels $\hat{\tensor{S}}^\mathrm{PR}_\mathrm{US3}$ and the corresponding superpixel refinements $\tilde{\tensor{L}}^\mathrm{PR}_{\mathrm{US3}}$ examples of which are presented in \autoref{fig:argmax_3ch_unsup} for the same slices.

\subsection{Unsupervised Training:  Given 4-phases (US4)}
\label{subsec:unsup_4ch}

\begin{figure}[h!]
\centering
\includegraphics[trim = 0 100 0 0, clip, width=0.8\textwidth]{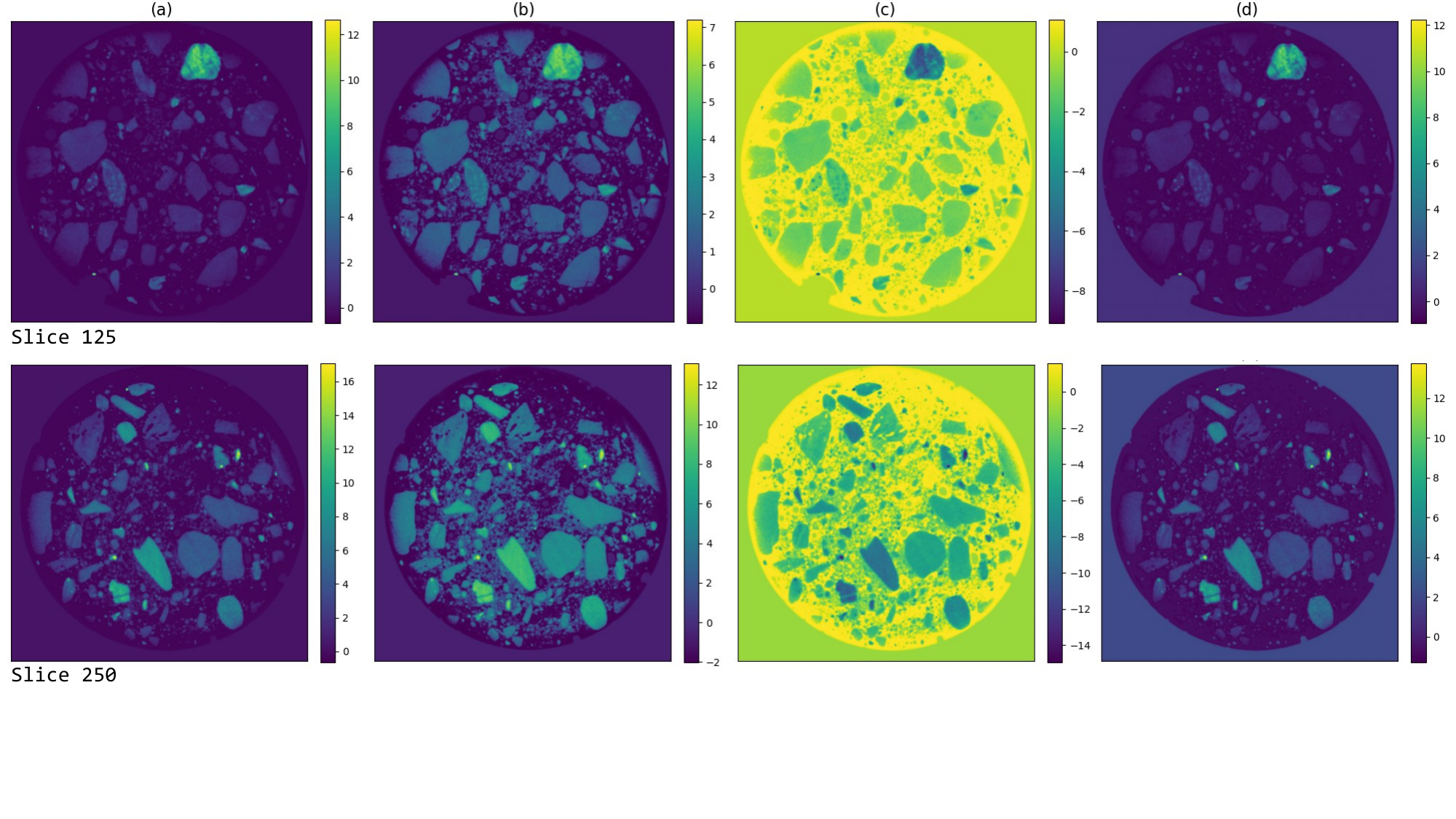}
\caption{$\mathrm{US4}$: Example of $4$-channel prediction results using model trained in the unsupervised scenario for input slices presented in \autoref{fig:3_channel_output} (left). Column (a) and (b) are associated with the aggregate phase, (c) is associated with the mortar phase. Column (d) is ambiguous with similar scores for porosity and aggregates.}
\label{fig:4_channel_output_us}
\end{figure}

\begin{figure}[h!]
\centering
\vspace*{0.25cm}
\begin{overpic}[trim = 0 200 0 0, clip, width=0.8\textwidth]{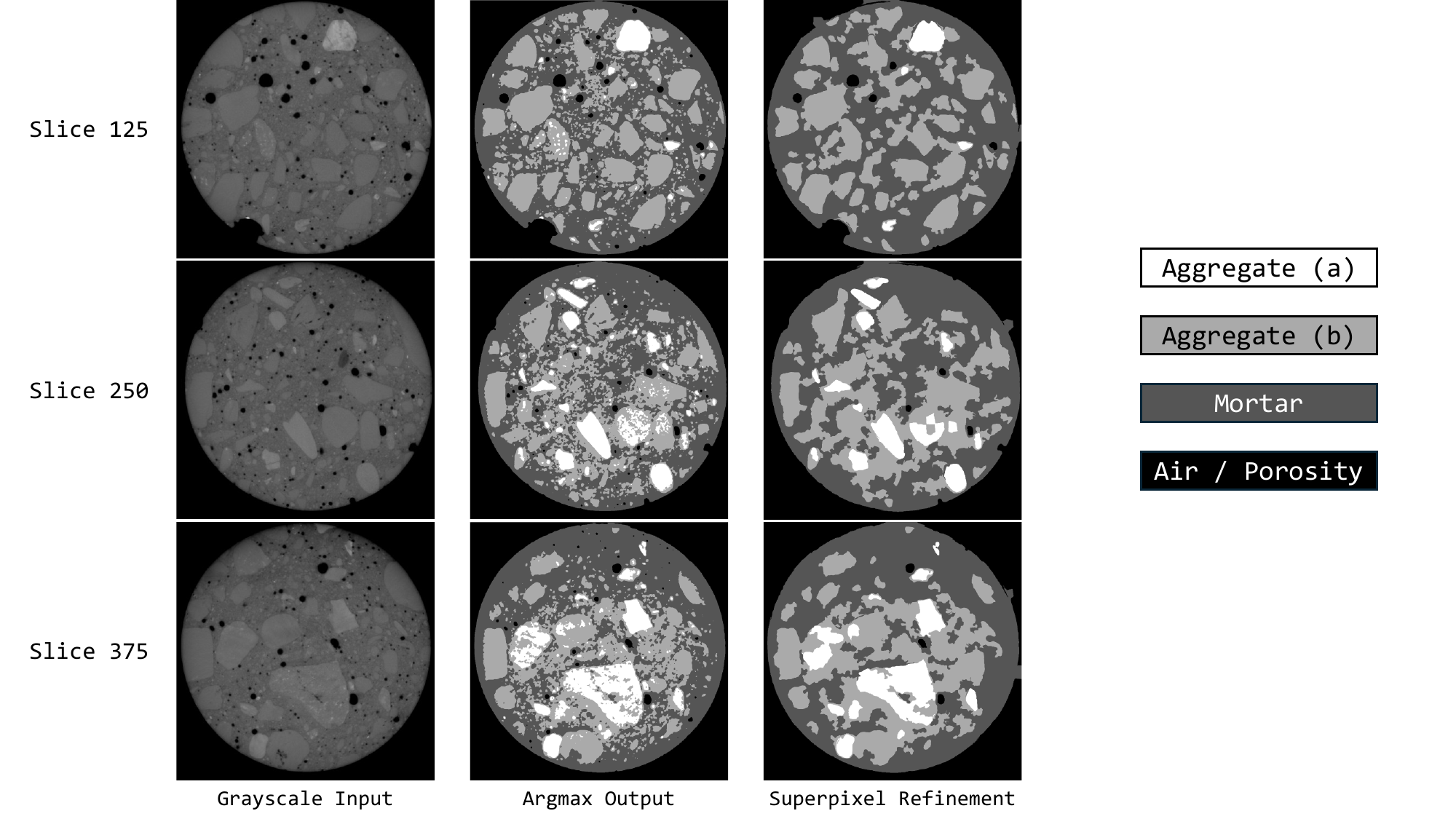}
\put(19,36){\tiny Input}
\put(32.3,36){\tiny $\arg\max$ classification}
\put(52.7,36){\tiny superpixel refinement}
\end{overpic}
\caption{$\mathrm{US4}$: $\arg\max$ classification of $\bar{\tensor{Y}}^{\mathrm{PR}}_\mathrm{US4}$ and the superpixel refinement $\tilde{\tensor{L}}^\mathrm{PR}_\mathrm{US4}$ using parameters equivalent to that used during training.}
\label{fig:argmax_4ch_unsup}
\end{figure}

The constraint on the number of output features was relaxed to $4$ to investigate if it had any ameliorative effect on the ambiguous classification of porosity and aggregates. The remainder of the training and prediction procedure is unchanged. However, now the self-annotation procedure was expected to generate $4$-distinct labels, which is more than the number of material phases.
Example of $\bar{\tensor{Y}}^{\mathrm{PR}}_\mathrm{US4}$ for identical slices presented in \autoref{fig:3_channel_output} (left) are presented in \autoref{fig:4_channel_output_us}. The model is able to unambiguously assign channels in $\bar{\tensor{Y}}^{\mathrm{PR}}_\mathrm{US4}$ to the aggregate and the mortar phases. Moreover, the model also assigns an additional channel to the aggregate phase, but it remains unable to unambiguously assign the remaining feature channel exclusively to the porous phase and as before generates similar scores for the aggregates and the porosity.

The resulting $\hat{\tensor{S}}^\mathrm{PR}_\mathrm{US4}$, for $\arg\max$ over all the $4$-channels is presented in \autoref{fig:argmax_4ch_unsup} alongside its superpixel-refinement $\tilde{\tensor{L}}^\mathrm{PR}_\mathrm{US4}$ for two slices. When $\hat{\tensor{S}}^\mathrm{PR}_\mathrm{US4}$ was generated by suppressing the aggregate channel (a), incorrect classification was observed for some particularly bright aggregates as the porous phase, similar to the previous case. In that respect, aggregate channel (a) seems to have a corrective effect and counteracts the incorrect classification of the ambiguous channel. Nevertheless, aggregate channel (b) continues to dominate the remaining aggregates and also results in incorrect classification of mortars as aggregates, particularly evident in regions with high concentration of small aggregates.

\subsection{Semi-supervised Training:  Given 3-phases (SS3)}
\label{subsec:semi_sup_3ch}

\begin{figure}[h!]
\centering
\includegraphics[trim = 0 100 0 0, width=0.8\textwidth]{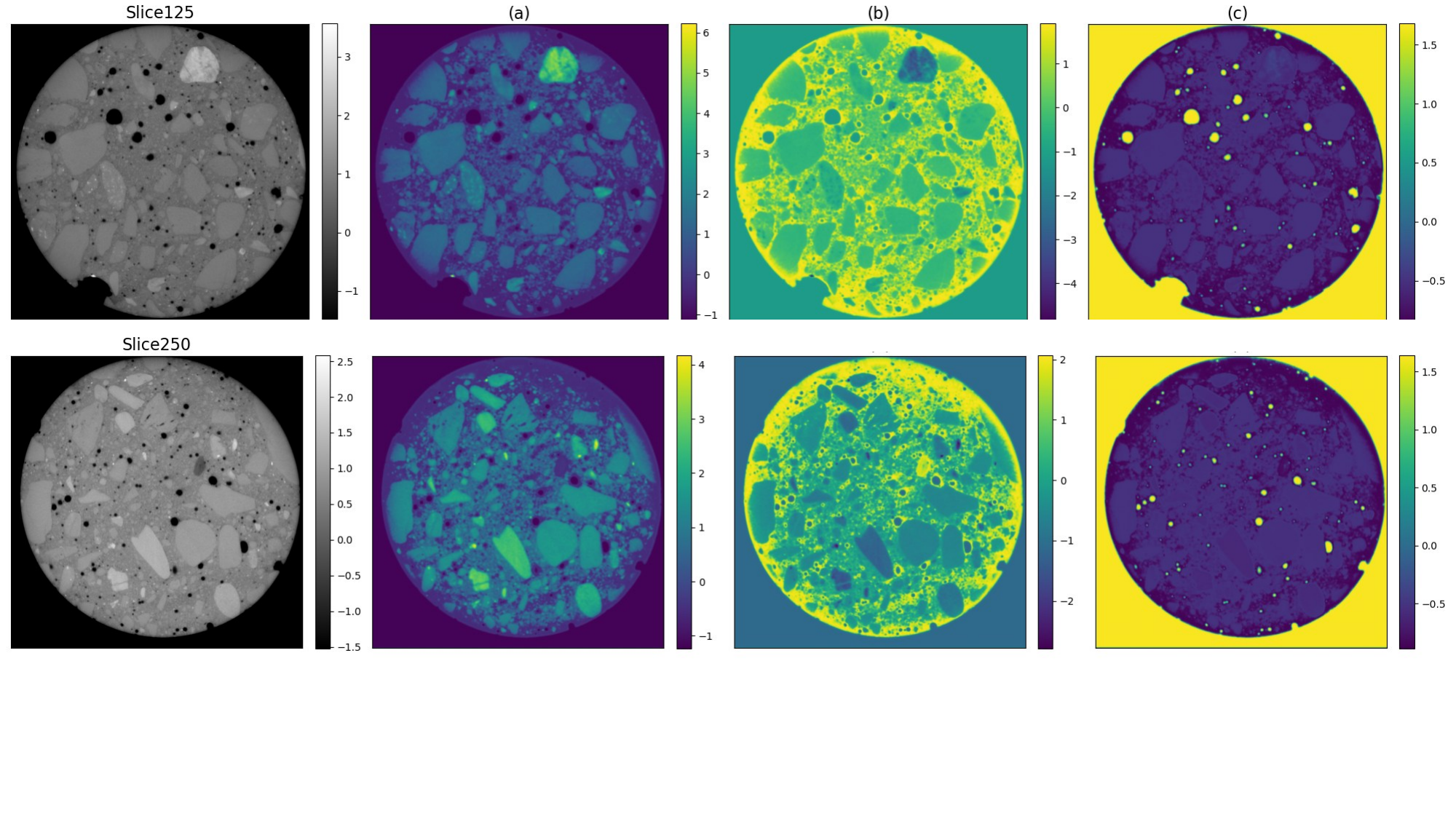}
\caption{$\mathrm{SS3}$: Example of $3$-channel prediction results using the model trained in the semi-supervised manner for the given input slices (left). Column (a), (b) and (c) can unambiguously be associated with the aggregate, mortar and porosity respectively.}
\label{fig:3_channel_output_semi_sup}

\centering
\vspace*{0.25cm}
\begin{overpic}[trim = 0 200 0 0, clip, width=0.8\textwidth]{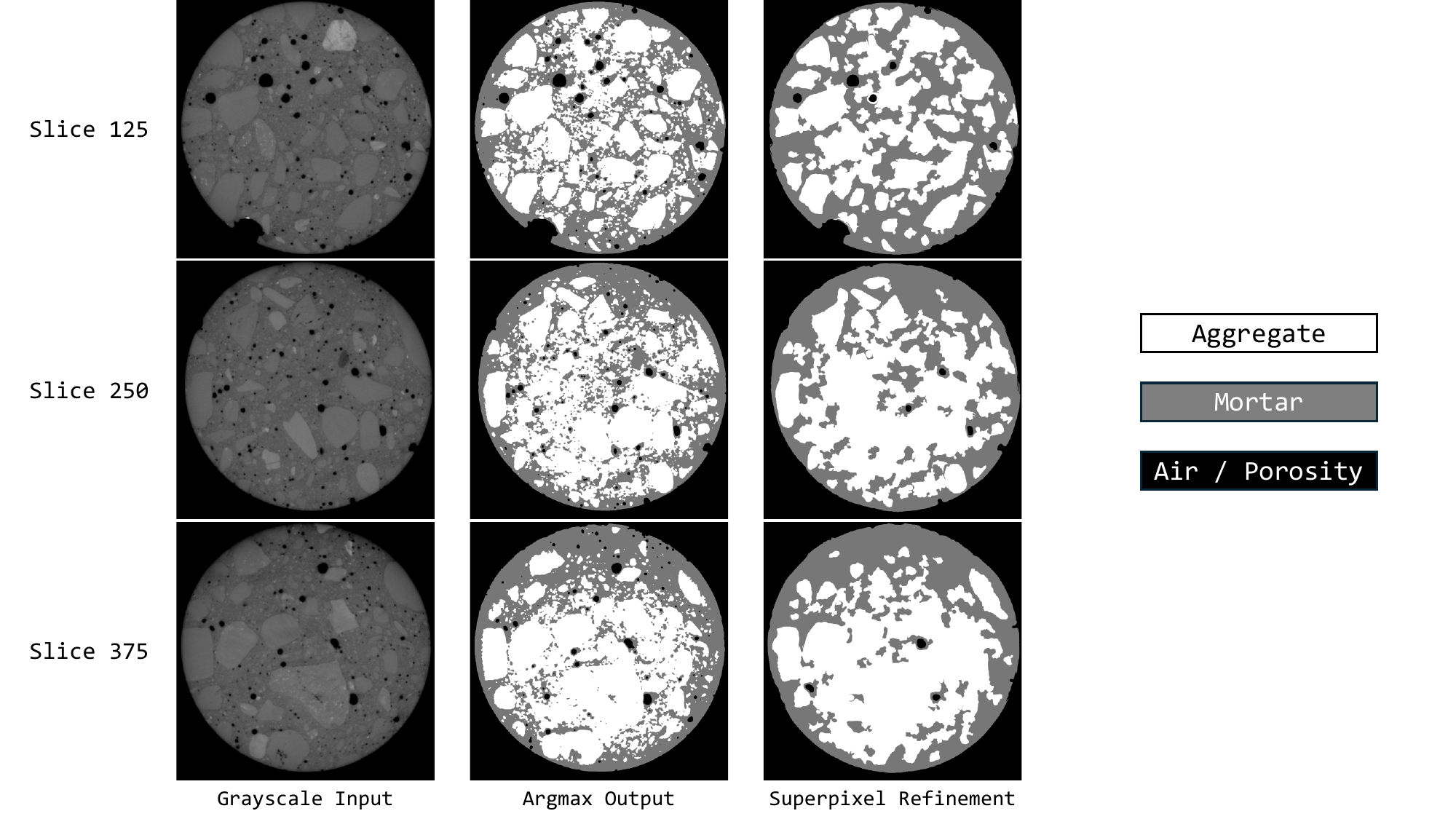}
\put(19,36){\tiny Input}
\put(32.3,36){\tiny $\arg\max$ classification}
\put(52.7,36){\tiny superpixel refinement}
\end{overpic}
\caption{$\mathrm{SS3}$: $\arg\max$ classification of $\bar{\tensor{Y}}^{\mathrm{PR}}_{\mathrm{SS3}}$ and the superpixel refinement using parameters equivalent to that used during training. }
\label{fig:argmax_3ch_semisup}
\end{figure}

Due to the limitations associated with the first two unsupervised cases involving ambiguity in identification of porosity, and considering that the porous phase in the XCT data is easy to resolve and segment using thresholding, experiments were also carried out in a semi-supervised manner whereby, the self-annotation procedure was modified to only label regions corresponding to the aggregate and the mortar phases while the regions occupied by the porous phase was annotated using the thresholded data. Given how well the intensities of the porous and non-porous regions are separated in the original XCT images, there are multiple thresholds at which almost identical results can be obtained, hence, specific details pertaining to thresholding are not presented here.

For the semi-supervised training, the model was constrained to produce $3$ feature channels at the output. At each iteration $t$ and for each image, the self-annotation procedure was first performed over two of the three output channels of $\bar{\tensor{Y}}_t$, which were pre-assigned to the aggregate and the mortar phases. This led to an intermediate binary dynamic label for the aggregates and mortar phases and included incorrectly labelled regions supposed to be associated with the porous phase. The label for the porous regions was subsequently corrected by assigning the corresponding label to those region to attain the final $3$-class label $\tilde{\tensor{L}}_t$ consisting of both automatically labelled regions and regions with predefined labels. No other changes to the remainder of the training and prediction procedure were made. Hence, the superpixel refinements in this case is not identical to the tensor $\tilde{\tensor{L}}_{t}$ in Section \ref{sec:unsupervised_segmentation} due to difference in the annotation procedure.

Normalized model prediction $\bar{\tensor{Y}}^{\mathrm{PR}}_\mathrm{SS3}$ was obtained comprising $3$ feature channels, examples of which for two slices are presented in \autoref{fig:3_channel_output_semi_sup}. From the images, it is evident that the model is able to resolve all the three distinct phases and hence assign a channel to each of the individual phases. In \autoref{fig:argmax_3ch_semisup}, $\hat{\tensor{S}}^\mathrm{PR}_\mathrm{SS3}$ is presented for two slices alongside their superpixel refinement.

\FloatBarrier
\section{Post-processing}
\label{sec:postprocessisng}

As demonstrated in the previous section, both the semi-supervised and the unsupervised methods were able to train the model such that it was able to distinguish between aggregates and mortar, without requiring labelled data for these phases. Nevertheless, generation of final classification label remains a challenge since the outputs obtained after $\arg\max$ classification using (\ref{eqn:argmax}) of $\bar{\tensor{Y}}_t$ is not necessarily optimal, hence some degree of post-processing is required.

\begin{figure}[ht!]
\centering
\includegraphics[trim = 0 50 0 0, clip, width=0.8\textwidth]{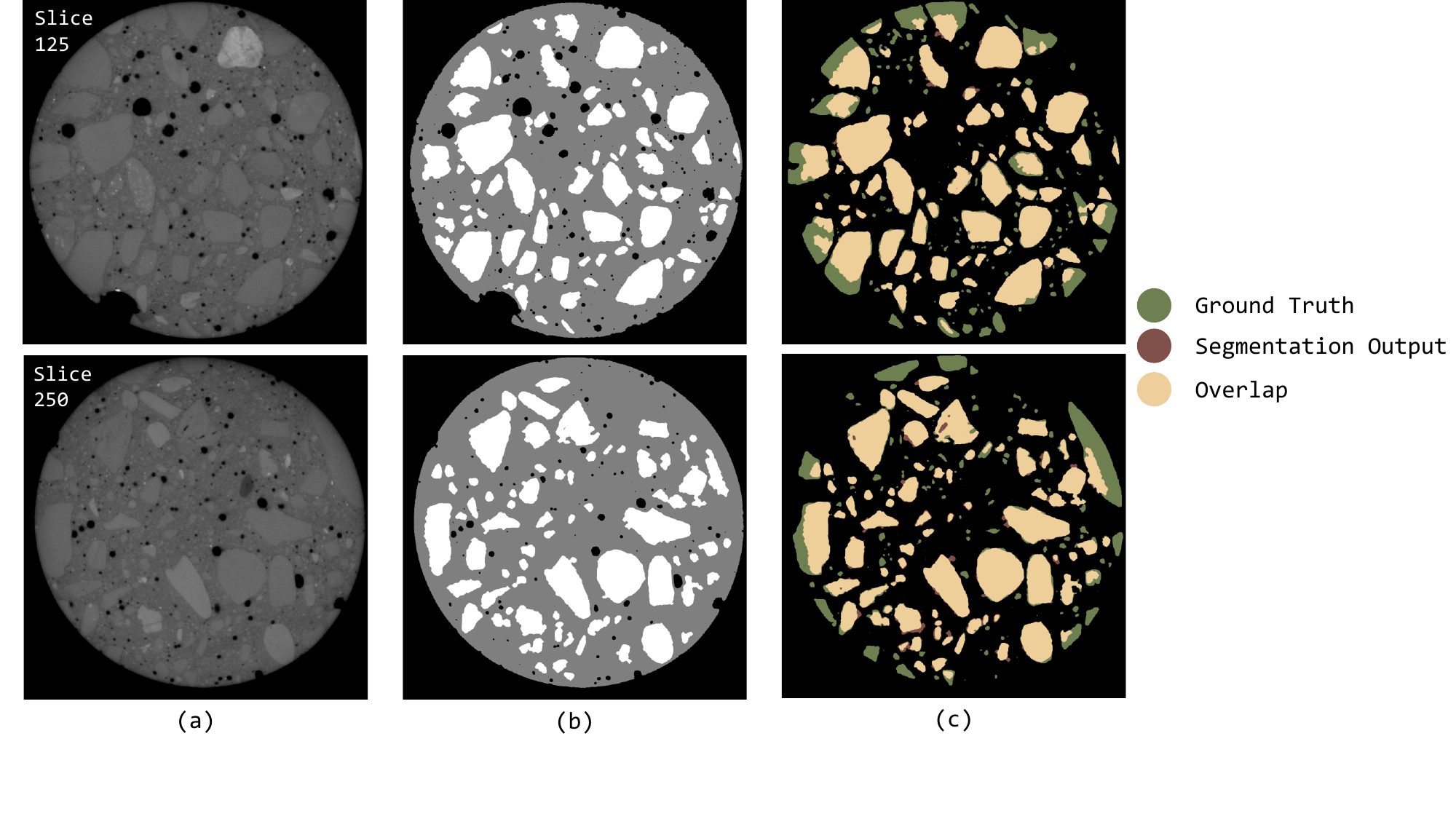}
\caption{Original XCT slices from $\tensor{T}^{\mathrm{PR}}$ (a). Example of 3-phase segmentation (b) using a thresholding-based post-processing approach. Aggregates identified in white, mortar in gray and porosity in black. Comparison of Aggregates obtained following the process against ground truth labels for aggregates (c).}
\label{fig:example_segmentations_1}
\end{figure}

For post-processing, a manual thresholding-based approach is employed for obtaining the final classification labels. In this case, each channel of the normalized model output is treated individually. Since in a $3$-phase scenario, a single phase is complementary to the remaining two phases, it is possible to use this approach on two of the three phases while third phase can be assigned the regions in an image left unoccupied by the other two phases. An example of this approach using $\bar{\tensor{Y}}^{\mathrm{PR}}_{\mathrm{SS3}}$ for generation of the final labels is presented in \autoref{fig:example_segmentations_1} where the channels corresponding to the porosity and aggregates were thresholded to generate a binary image for each phase and subsequently combined to produce three-phase images. Although, the threshold for each image is unique, for an entire sample, one may generate thresholds for a small set of slices at certain intervals of an XCT sample and subsequently use an interpolation algorithms, for example PCHIP Interpolator which is $\mathrm{C}1$-smooth \cite{fritsch1984method}, to generate thresholds for the intermediate slices.

It is however to be noted that several sophisticated post-processing techniques for semantic segmentation exist such as the use of conditional random fields \cite{kamnitsas2017efficient}, Markov random fields \cite{liu2017deep}, combination of global and local context of image regions \cite{zhu2023adaptive} etc.. Nevertheless, the post-processing technique discussed in this work facilitates direct assessment of the current segmentation technique's ability to distinguish between aggregates and mortar in concrete and therefore helps in establishing a baseline for subsequent improvements.

\FloatBarrier
\section{Results and Evaluation}
\label{sec:results_and_eval}

\begin{figure}[hb!]
\centering
\includegraphics[trim = 0 50 0 0, clip, width=0.8\textwidth]{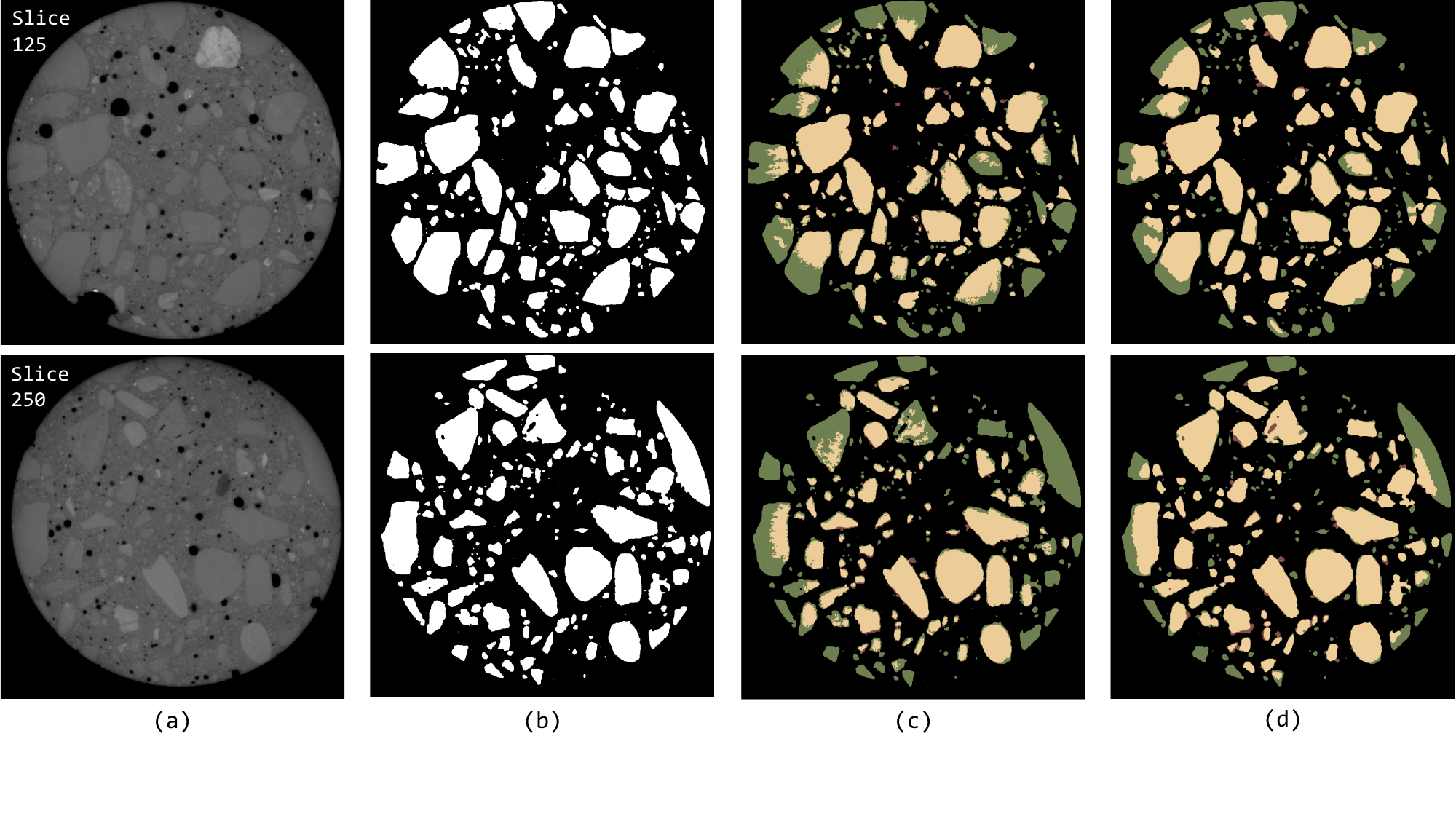}
\caption{Comparison of binarized results of aggregates. (a) Input slices, (b) Ground Truth, (c) Aggregates obtained by direct manual thresholding of $\tensor{T}^{\mathrm{PR}}$ vs. Ground Truth, (d) Aggregates obtained by thresholding the aggregate-specific channel of $\bar{\tensor{Y}}^{\mathrm{PR}}_\mathrm{SS3}$ vs. Ground Truth.}
\label{fig:thresholding_vs_segmentation}
\end{figure}

For evaluating the final results $\tensor{T}_{\mathrm{PR}}$ was used and 16-slices from it, taken at fixed intervals, were manually annotated and are henceforth referred to as the ground truth. The evaluation is only focused on the aggregates since mortar is complementary to it. $\bar{\tensor{Y}}^{\mathrm{PR}}_\mathrm{SS3}$ was used since it is able to unambiguously assign a channel each to the given phases. The thresholding-based approach for post-processing was used on the channel corresponding to the aggregate phase to obtain binary images consisting of only the aggregates. The resulting binary images of aggregates were further processed using a series of morphological operations to reduce noise and remove artifacts. Small holes up to $512$px in area were first filled, followed by binary erosion to eliminate narrow connections between large objects. Subsequently, objects smaller than $128$px in area were removed, and a final binary dilation was applied to restore areas lost during erosion. Furthermore, objects smaller than $256$px were also removed from the manually annotated (ground truth) images since they were too small to be considered part of the aggregate phase. For additional comparison, we also use examples of the aggregates obtained via direct manual thresholding of the XCT slices of $\tensor{T}^{\mathrm{PR}}$, which were processed by removing objects smaller than $128$px and removing holes smaller than $256$px. In \autoref{fig:thresholding_vs_segmentation}, an example of the results obtained by thresholding $\bar{\tensor{Y}}^{\mathrm{PR}}_\mathrm{SS3}$ is contrasted with the ground truth and the aggregates obtained by the direct thresholding of $\tensor{T}^{\mathrm{PR}}$.

\FloatBarrier
\subsection{Qualitative Evaluation}

\begin{figure}[htb!]
\centering
\includegraphics[trim = 0 60 0 20, clip, width=0.8\textwidth]{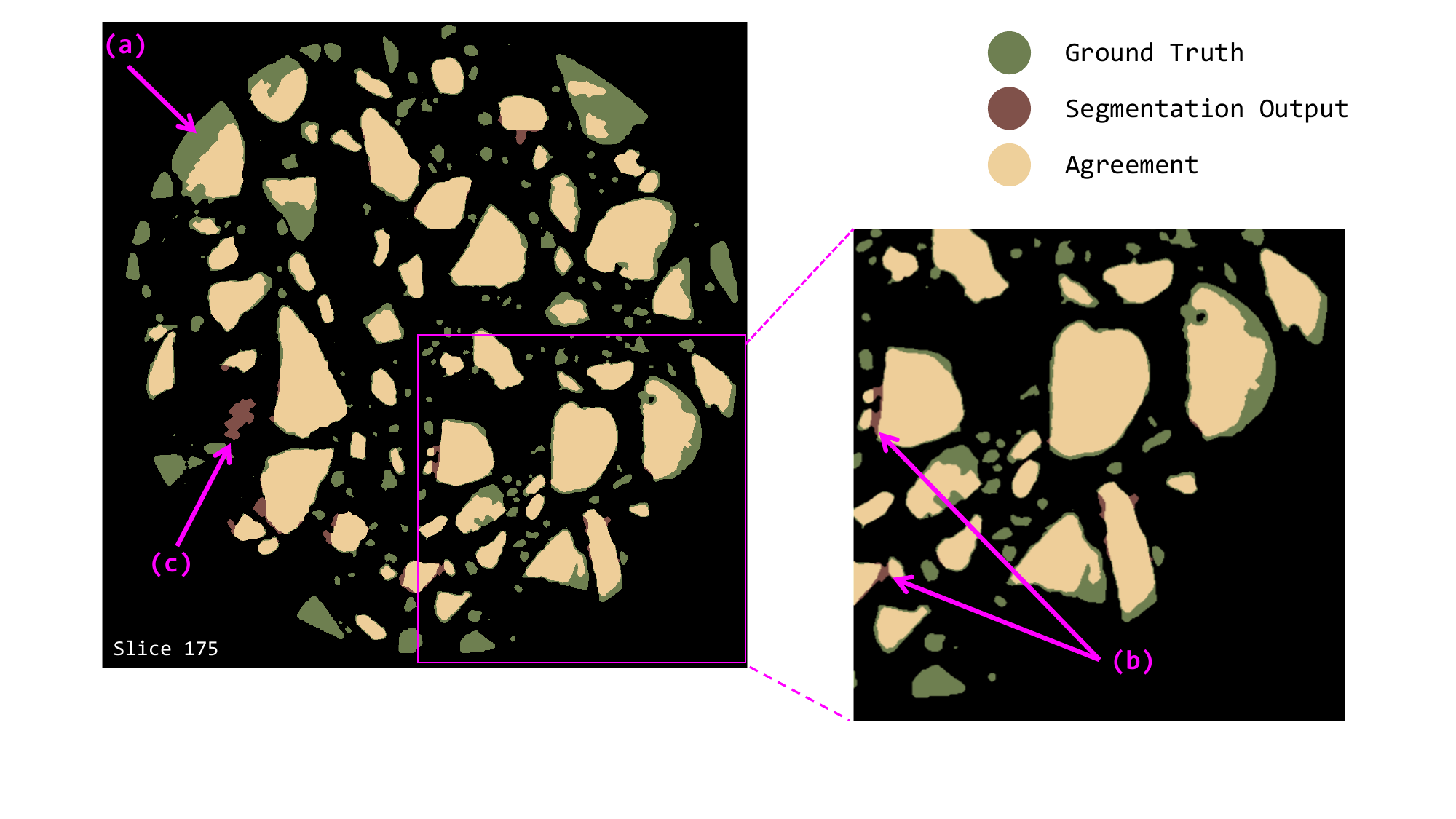}
\caption{(a) Model fails to identify aggregates at the periphery of the cylindrical casting.  (b) In case of aggregates located in close proximity, the model fails to effectively demarcate the boundaries between such aggregates. (c) A number of small aggregates clumped together to form incorrectly labelled aggregates.} \label{fig:qualitative_segmentation_analysis}
\end{figure}

From \autoref{fig:thresholding_vs_segmentation}, it is evident that the segmentation model produces better results when compared to direct thresholding of XCT samples. However, on a qualitative standpoint a few limitations were also observed as shown in \autoref{fig:qualitative_segmentation_analysis} for a single slice against the ground truth. It was observed that the model performs well in identifying most of the aggregates, however, it consistently fails to detect aggregates near the periphery of the cylinder across the entire sample, as indicated by (a) in \autoref{fig:qualitative_segmentation_analysis}. We assume that the sharp drop in pixel values between the solid phase and the porous region at the periphery of the cylinder could be the primary source of this error considering (\ref{eqn:convolution}) where a weighted summation of pixels are computed. However, verification of this requires further investigation. Moreover, throughout the sample, in cases where aggregates are in close proximity, the model fails to identify narrow gaps of mortar between aggregates as shown by (b) in \autoref{fig:qualitative_segmentation_analysis}. This, as a consequent has the effect that in regions where a number of small aggregates are located in close proximity, which should generally be assigned to the mortar phase based on their size, the model essentially detects these aggregates and joins them to form large regions that are incorrectly classified as aggregate indicated by (c) in \autoref{fig:qualitative_segmentation_analysis}.

\subsection{Quantitative Evaluation}

\begin{figure}[h!]
\centering
\includegraphics[trim = 0 210 0 0, clip, width=0.9\textwidth]{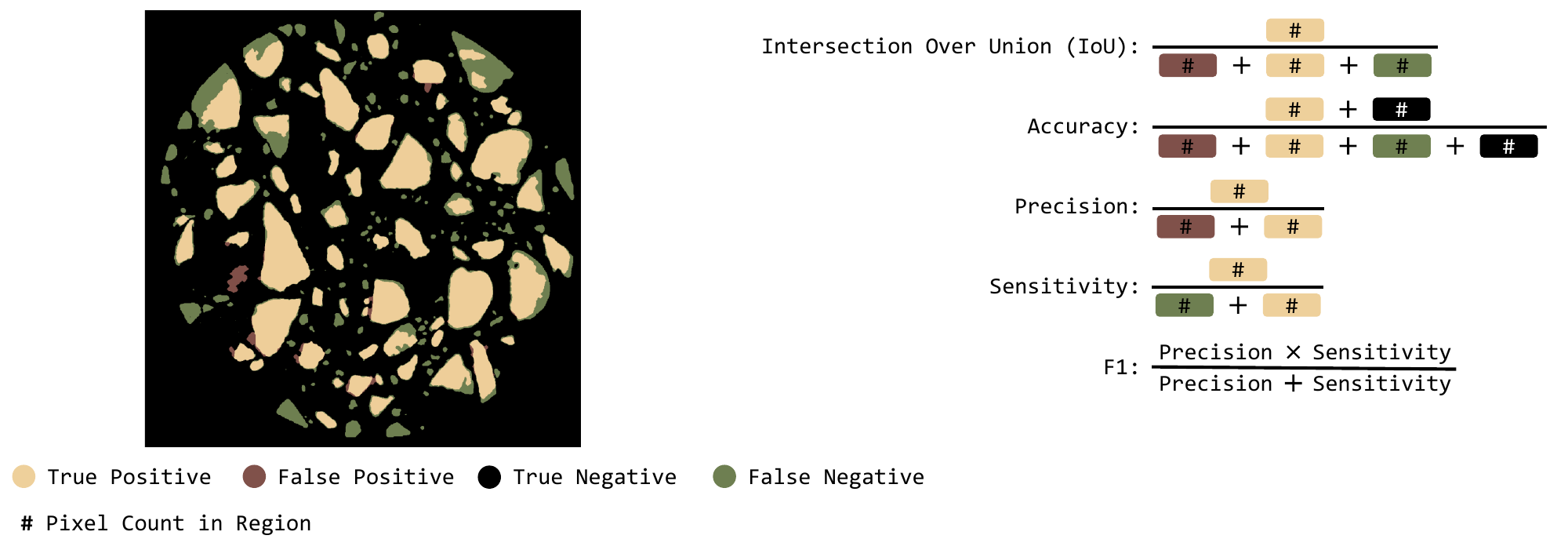}
\caption{Interpretation of colours and definition of performance metrics.}
\label{fig:quantitative_evaluation_metrics}
\end{figure}

\begin{figure}[h!]
\centering
\includegraphics[trim = 0 0 0 0, clip, width=0.49\textwidth]{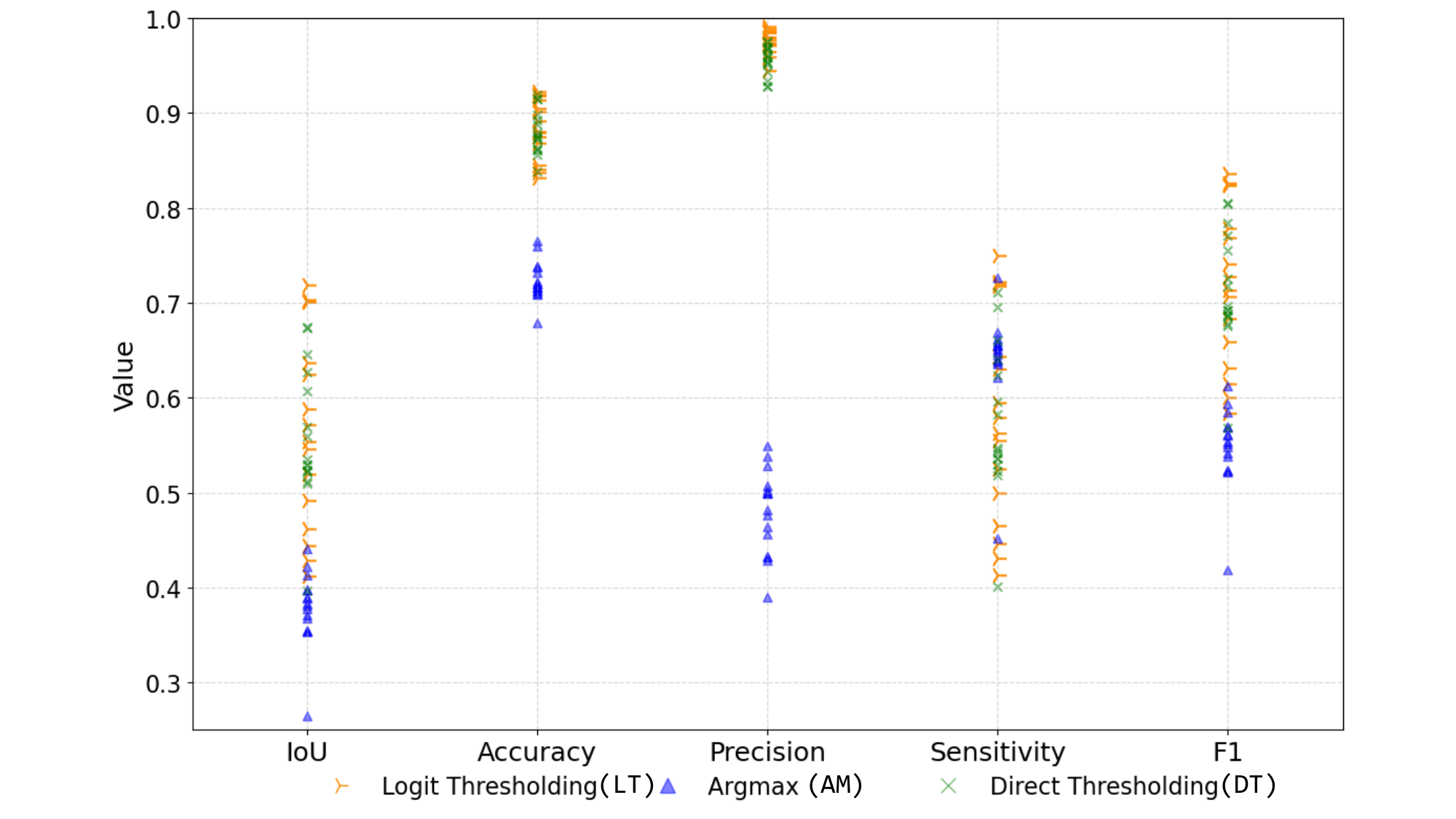}
\includegraphics[trim = 0 0 0 0, clip, width=0.49\textwidth]{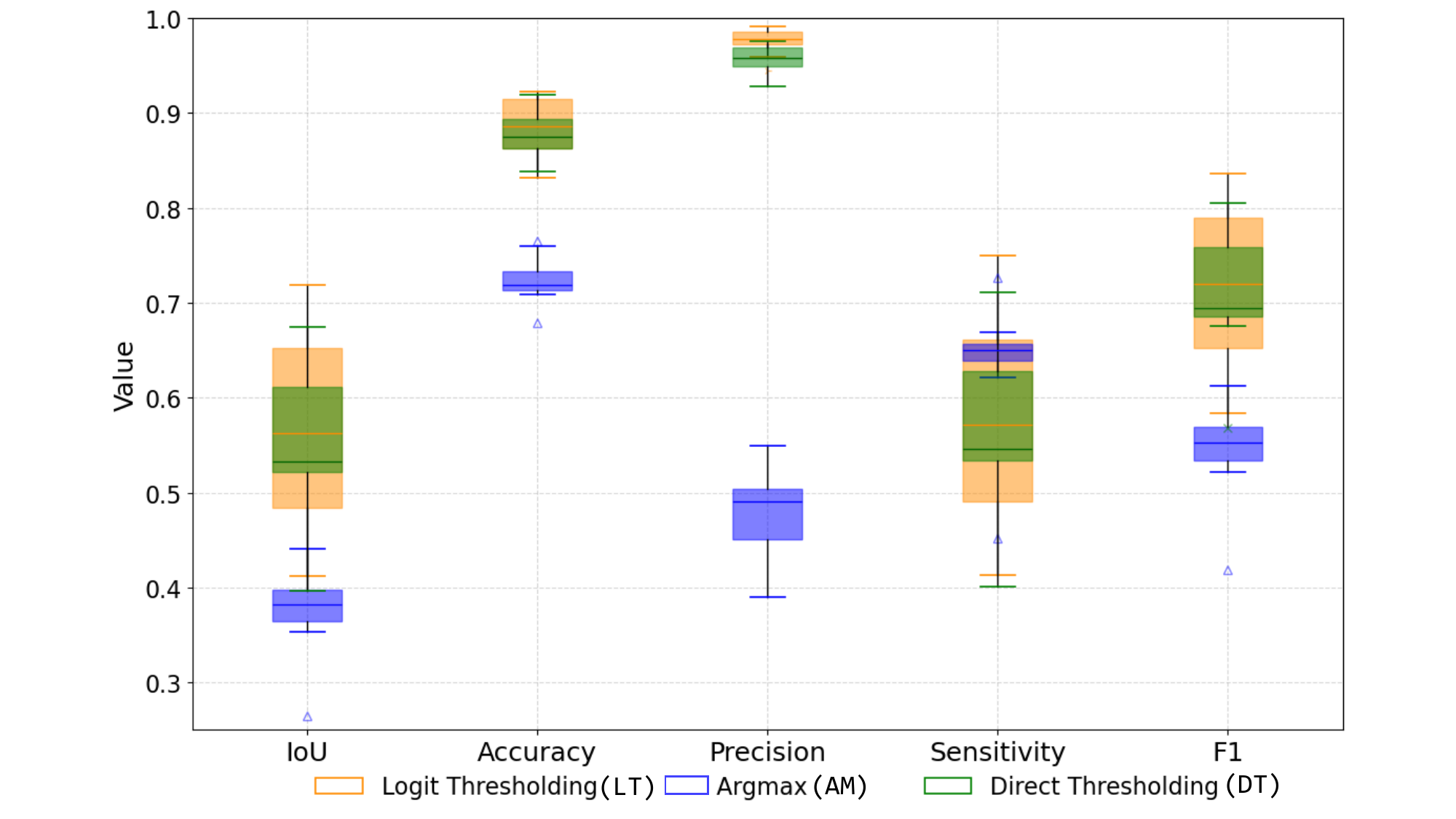}
\caption{Quantitative evaluation of segmentation quality focusing on the aggregates compared to the ground truth of $16$ slices using both post-processing methods. Logit thresholding (LT): thresholding applied to the aggregate-specific
channel of $\bar{\tensor{Y}}^{\mathrm{PR}}_\mathrm{SS3}$, Direct Thresholding (DT): thresholding applied to $\tensor{T}^{\mathrm{PR}}$ and $\arg\max$ (AM): aggregates extracted after  $\arg\max$ is applied to $\bar{\tensor{Y}}^{\mathrm{PR}}_\mathrm{SS3}$.}
 \label{fig:quantitative_evaluation}
\end{figure}

Quantitative evaluation was carried out using the metrics shown in \autoref{fig:quantitative_evaluation_metrics} for each of the $16$ labelled slices of $\tensor{T}_{\mathrm{PR}}$, considering only the aggregate phase. The results in \autoref{fig:quantitative_evaluation} and \autoref{tab:quantitative_metric_table} compare the following three approaches: thresholding applied to the aggregate-specific channel of $\bar{\tensor{Y}}^{\mathrm{PR}}_\mathrm{SS3}$ (LT), thresholding applied to $\tensor{T}^{\mathrm{PR}}$ (DT), and $\arg\max$ classification (AM) of $\bar{\tensor{Y}}^{\mathrm{PR}}_\mathrm{SS3}$. 

\begin{table}[htbp]
\centering
\scriptsize
\setlength{\tabcolsep}{3.5pt} 
\caption{Statistical summary of quantitative evaluation metrics for aggregate-phase segmentation across 16 labeled slices of $\tensor{T}_{\mathrm{PR}}$, comparing logit thresholding, direct thresholding, and $\arg\max$ classification. Logit thresholding achieves the strongest overall performance across most metrics, particularly in IoU, accuracy, precision, and F1-score.}
\label{tab:quantitative_metric_table}
\begin{tabular}{|l|l c c c c c c c c c c c c|}
\hline
{Metric} & {Model} & {Mean} & {Median} & {Min.} & {Max.} & {Range} & {Var.} & {Std. Dev.} & {$Q_1$ (25\%)} & {$Q_3$ (75\%)} & {IqR} & {Skew.} & {Kurt.} \\
\hline
\multirow{3}{*}{\textbf{IoU}} 
& LT  & 0.57 & 0.56 & 0.41 & 0.72 & 0.31 & 0.01 & 0.10 & 0.48 & 0.65 & 0.17 & 0.03 & -1.29    \\
& AM  &  0.38 & 0.38 & 0.26 & 0.44 & 0.18 & 0.002 & 0.04 & 0.36 & 0.40 & 0.03 & -1.30 & 2.75  \\
& DT   & 0.56 & 0.53 & 0.40 & 0.67 & 0.28 & 0.005 & 0.07 & 0.52 & 0.61 & 0.09 & -0.10 & 0.01  \\
\hline
\multirow{3}{*}{\textbf{Accuracy}} 
& LT  & 0.884 & 0.89 & 0.83 & 0.922 & 0.091 & 0.001 & 0.03 & 0.87 & 0.91 & 0.05 & -0.37 & -1.26   \\
& AM & 0.72 & 0.72 & 0.68 & 0.77 & 0.086 & 0.0004 & 0.021 & 0.71 & 0.73 & 0.02 & 0.29 & 0.55      \\
& DT   & 0.880 & 0.88 & 0.84 & 0.918 & 0.081 & 0.0005 & 0.023 & 0.86 & 0.89 & 0.03 & 0.24 & -0.70 \\
\hline
\multirow{3}{*}{\textbf{Precision}} 
& LT  & 0.98 & 0.98 & 0.94 & 0.99 & 0.05 & 0.00015 & 0.01 & 0.97 & 0.99 & 0.01 & -1.05 & 0.71   \\
& AM & 0.48 & 0.49 & 0.39 & 0.55 & 0.16 & 0.002 & 0.04 & 0.45 & 0.50 & 0.05 & -0.34 & -0.62     \\
& DT   & 0.96 & 0.96 & 0.93 & 0.98 & 0.05 & 0.00025 & 0.02 & 0.95 & 0.97 & 0.02 & -0.61 & -0.76 \\
\hline
\multirow{3}{*}{\textbf{Sensitivity}} 
& LT  & 0.58 & 0.57 & 0.41 & 0.75 & 0.34 & 0.01 & 0.11 & 0.49 & 0.662 & 0.17 & 0.09 & -1.23   \\
& AM & 0.64 & 0.65 & 0.45 & 0.73 & 0.27 & 0.003 & 0.06 & 0.64 & 0.657 & 0.02 & -2.47 & 7.11   \\
& DT   & 0.57 & 0.55 & 0.40 & 0.71 & 0.31 & 0.006 & 0.08 & 0.53 & 0.63 & 0.09 & -0.05 & 0.02  \\
\hline
\multirow{3}{*}{\textbf{F1-Score}} 
& LT  & 0.72 & 0.72 & 0.58 & 0.84 & 0.25 & 0.01 & 0.09 & 0.65 & 0.79 & 0.14 & -0.11 & -1.26    \\
& AM & 0.55 & 0.55 & 0.42 & 0.61 & 0.19 & 0.002 & 0.043 & 0.53 & 0.57 & 0.035 & -1.56 & 3.41   \\
& DT & 0.71 & 0.69 & 0.57 & 0.81 & 0.24 & 0.004 & 0.06 & 0.69 & 0.76 & 0.07 & -0.39 & 0.49     \\
\hline
\end{tabular}
\end{table}

The results indicate that LT provides the most favourable balance between overlap quality and class-wise correctness. It achieves the highest mean IoU ($0.57$), Accuracy ($0.884$), Precision ($0.98$), and F1-score ($0.72$), suggesting that it produces segmentations that are not only spatially closer to the reference masks, but also contain fewer false-positive predictions. By contrast, AM attains the highest mean Sensitivity ($0.64$), which indicates a tendency to capture more of the target region, but this comes with a marked reduction in Precision ($0.48$), IoU ($0.38$), and F1-score ($0.55$), which is consistent with a tendency toward more liberal positive predictions leading to over-segmentation. Relative to DT, LT yields small but consistent improvements across the mean metrics, indicating that its performance is more favourable on average across the evaluated slices. In addition, LT achieves the best observed peak performance, with a maximum IoU of $0.72$ and F1-score of $0.84$, showing that it can produce particularly accurate segmentations in the most favourable cases. This suggests that LT not only improves typical performance, but also has a higher best-case segmentation potential than DT. The spread statistics show that LT is somewhat more variable than DT for several metrics, as reflected by larger ranges, variances, standard deviations, and interquartile ranges, particularly for IoU and Sensitivity. For example, LT has a wider IoU range ($0.31$ vs. $0.18$) and IQR ($0.17$ vs. $0.03$) than DT, and similarly a wider Sensitivity range ($0.34$ vs. $0.27$) and IQR ($0.17$ vs. $0.02$). At the same time, the close agreement between mean and median values for LT suggests that its performance is fairly balanced and not driven by a small number of extreme cases.

Taken together, these findings suggest that LT offers the most favourable trade-off for aggregate-phase semantic segmentation. It achieves the strongest overall mask agreement and class-wise correctness on average, while also attaining the best peak performance. In particular, LT provides a better balance between sensitivity and precision, reducing false-positive predictions while maintaining competitive sensitivity relative to DT and AM. Although LT exhibits a somewhat broader distribution of outcomes across slices than DT, which may be attributed to certain pathological cases at the top or bottom adjacent slices particular to $\tensor{T}^{\mathrm{PR}}$ that were not sufficiently represented in the training dataset, its overall performance indicates greater segmentation effectiveness.

\section{Conclusion}
\label{sec:conclusion}

XCT images of concrete often show poor discernibility between aggregates and mortar because these phases have similar X-ray attenuation coefficients, resulting in overlapping greyscale intensities. This low contrast creates a challenge for semantic segmentation of the images into distinct, non-overlapping regions corresponding to each phase. Although deep learning-based methods have proven capable of segmenting such images, the limited availability of labelled training datasets remains a major barrier to wider adoption of XCT for downstream applications. The self-annotation procedure presented in this study for unsupervised training of a CNN-based U-Net segmentation model demonstrates the potential to address the challenge of segmenting low-contrast XCT scans of concrete in low-data regimes. The proposed approach enables the model to identify the prescribed phases without labelled data and to transform the bimodal pixel-value distribution of the original greyscale XCT images in a way that improves discernibility. 

The results presented were obtained from out-of-distribution data, i.e., a test sample which was not used for any part of the training, unlike previous studies where the models are usually tested on unused slices of the same sample. Furthermore, no test data-specific model fine-tuning was performed. While, ideally, the $\arg\max$ classification should be the final segmentation, in this case it resulted in over-segmentation. Instead, thresholding of the logit channel corresponding to the aggregate phase was found to be more suitable, and resulted in improved performance compared to direct thresholding of the greyscale images. The resulting classification was evaluated against manually annotated ground-truth slices and consistently outperformed direct thresholding of the original XCT scans in aggregate-phase identification. Qualitative analysis showed strong overall detection of aggregates, although limitations were observed near the sample periphery and in regions where aggregates were closely spaced, leading to occasional under- and over-segmentation. Among the evaluated approaches, logit thresholding achieved the best overall performance, with the highest mean in all evaluation metrics, indicating superior overlap quality and class-wise correctness. Although logit thresholding exhibited greater variability across slices than direct thresholding, it provided the most favourable balance between sensitivity and precision, demonstrating the strongest overall segmentation effectiveness.

It is important to note that, due to the unsupervised nature of the proposed method, standard ad-hoc data augmentation techniques were deliberately excluded. Given the presence of XCT-specific attributes in the images and the dynamic labelling approach, such techniques were deemed suboptimal in this context. Rather than applying generic pipelines, the investigation of tailored augmentation strategies has been reserved for future work. Although the untrained U-Net architecture was fixed from the outset, future work could examine whether smaller models, different model architectures or pre-training can achieve comparable performance for this specific application. Furthermore, extending the approach to 3D should be straightforward from the current 2D framework.

Nevertheless, certain limitations of the method were also identified, specifically the consistent inability of the model to unambiguously distinguish between aggregates and porosity in the fully unsupervised scenario, which necessitated the usage of masked dynamic labels to limit self-annotation to mortar and aggregates only. Furthermore, per-channel normalization of the model output is an aspect that requires further studies. Although necessary to prevent the model from converging to a degenerate solution, rescaling of the model output for $\arg\max$ classification is detrimental in cases where certain semantic classes are not sufficiently represented in a given image. 

Overall, this study demonstrates that an unsupervised CNN-based U-Net framework can extract meaningful phase information from low-contrast concrete XCT scans without requiring labelled training data, while remaining robust on out-of-distribution samples. By improving the separation of aggregates and mortar in challenging greyscale volumes, the method offers a practical route for rapid pre-segmentation and downstream applications. At the same time, the observed difficulties in resolving porosity, peripheral aggregates, and densely clustered inclusions underscore the need for class-aware refinement and further methodological development, particularly in the treatment of output normalization and segmentation thresholds. Future work should therefore focus on more adaptive post-processing, tailored augmentation strategies, and extensions to 3D architectures, with the goal of establishing a more generalizable and physically faithful framework for automated concrete microstructure analysis.

\section*{Data Availability Statement}
Processed datasets, software implementation and model weights are made available via Zenodo at \url{https://doi.org/10.5281/zenodo.18684512}.

\section*{CRediT Author Statement}
KD: Conceptualization, Methodology, Software, Data Curation (Pre- and Post-processing), Investigation, Formal Analysis, Validation, Writing - Original Draft,  Visualization;  
GR: Data Curation (XCT Scanning), Resources, Writing - Reviewing \& Editing; 
JS: Conceptualization, Writing - Reviewing \& Editing, Resources,  Funding Acquisition; 
AK: Conceptualization, Methodology, Writing - Reviewing \& Editing, Supervision, Funding Acquisition, Project Administration.

\section*{Acknowledgement}

This work was developed within the bilateral SUMO project, funded by the Luxembourg National Research Fund (FNR) through the project INTER/GACR/21/16555380 and the Czech Science Foundation (GACR), project No. 22-35755K. KD additionally acknowledges financial support from the Student Grant Competition of the Czech Technical University in Prague, project No. SGS23/ 152/OHK1/3T/11. JS additionally acknowledges financial support from the European Union through the Operational Programme Jan Amos Komenský under project INODIN (CZ.02.01.01/00/23\_020/0008487).



\bibliographystyle{unsrturl}
\bibliography{main}

\clearpage

\appendix
\end{document}